\documentclass[sigconf,edbt]{acmart-edbt2022}

\usepackage{graphicx}
\usepackage{color}
\usepackage{comment}
\usepackage{subcaption}

\usepackage{listings}
\definecolor{mygreen}{RGB}{28,172,0}
\definecolor{mylilas}{RGB}{170,55,241}
\usepackage{hyperref}
\hypersetup{
    colorlinks=true,
    linkcolor=blue,
    urlcolor=blue,
}

\usepackage{titlesec} 

\def\x{\textit{\textbf{x}}}
\def\y{\textit{\textbf{y}}}

\def\imBeamAMF{\texttt{im}$/\!\sqrt{\texttt{AMF}}$\,}
\def\imBeamGF{\texttt{im}$/\!\sqrt{\texttt{GF}}$\,}

\def\heightFigDarkFace{1.56cm} 
\def\heightFigDarkFaceAdaptive{1.56cm} 

\def\shrinkSpaceBetweenImages{-0.35cm}     

\def\heightTeaser{2.13cm}
\def\sFigHaze{0.137}

\begin{document}

\title{SPICE: Simple and Practical Image Clarification and Enhancement}

\author{Alexander Belyaev}
\affiliation{%
  \institution{School of Engineering \& Physical Sciences, Heriot-Watt University}
  \country{Edinburgh, UK}}
\email{a.belyaev@hw.ac.uk}

\author{Pierre-Alain Fayolle}
\affiliation{%
  \institution{Information Systems Division \\ University of Aizu}
  \country{Aizu-Wakamatsu, Japan}
}
\email{fayolle@u-aizu.ac.jp}

\author{Michael Cohen}
\affiliation{%
  \institution{Higashi Nippon International University and University of Aizu}
  \country{Japan}
}
\email{mcohen@m.tonichi-kokusai-u.ac.jp}

\begin{abstract}
We introduce a simple and efficient method to enhance and clarify images. More specifically, we deal with low light image enhancement and clarification of hazy imagery (hazy/foggy images, images containing sand dust, and underwater images). Our method involves constructing an image filter to simulate low-light or hazy conditions and deriving approximate reverse filters to minimize distortions in the enhanced images. Experimental results show that our approach is highly competitive and often surpasses state-of-the-art techniques in handling extremely dark images and in enhancing hazy images. A key advantage of our approach lies in its simplicity: Our method is implementable with just a few lines of MATLAB code (in addition to the MATLAB scripts presented in the paper, MATLAB implementations of the proposed image enhancing schemes are available at \url{https://github.com/ag-belyaev/spice}).
\end{abstract}

\begin{teaserfigure}
\hspace*{-0.2cm}
\centering
\begin{tabular}{cccccc}
\includegraphics[height=\heightTeaser]{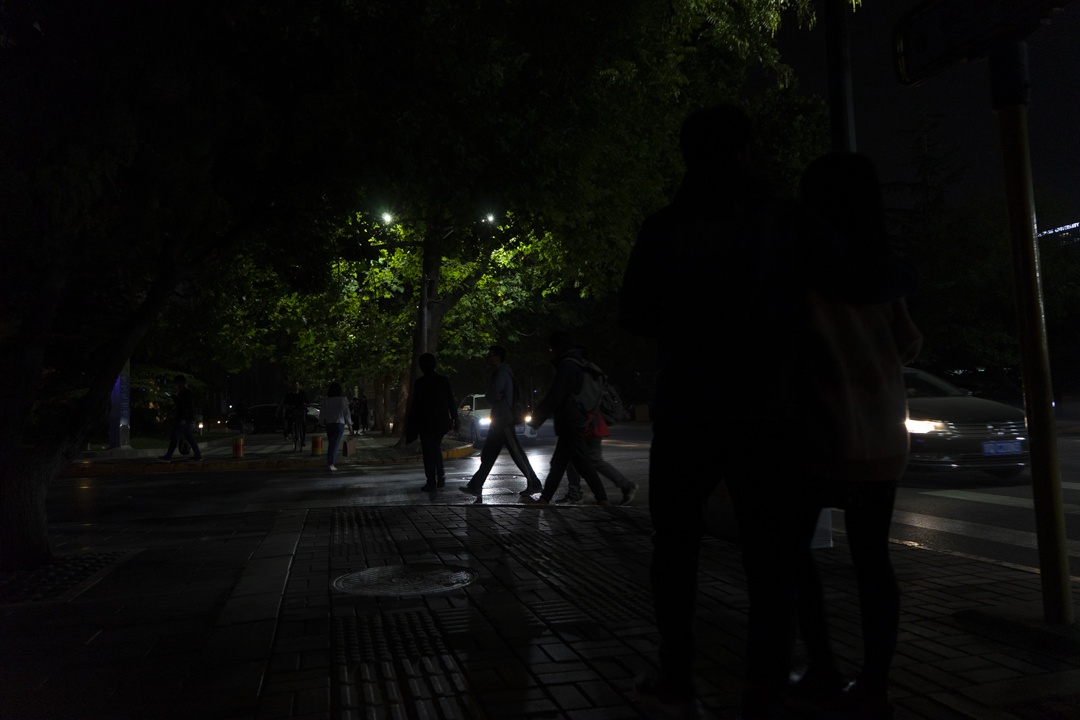}
\hspace*{\shrinkSpaceBetweenImages}&
\includegraphics[height=\heightTeaser]{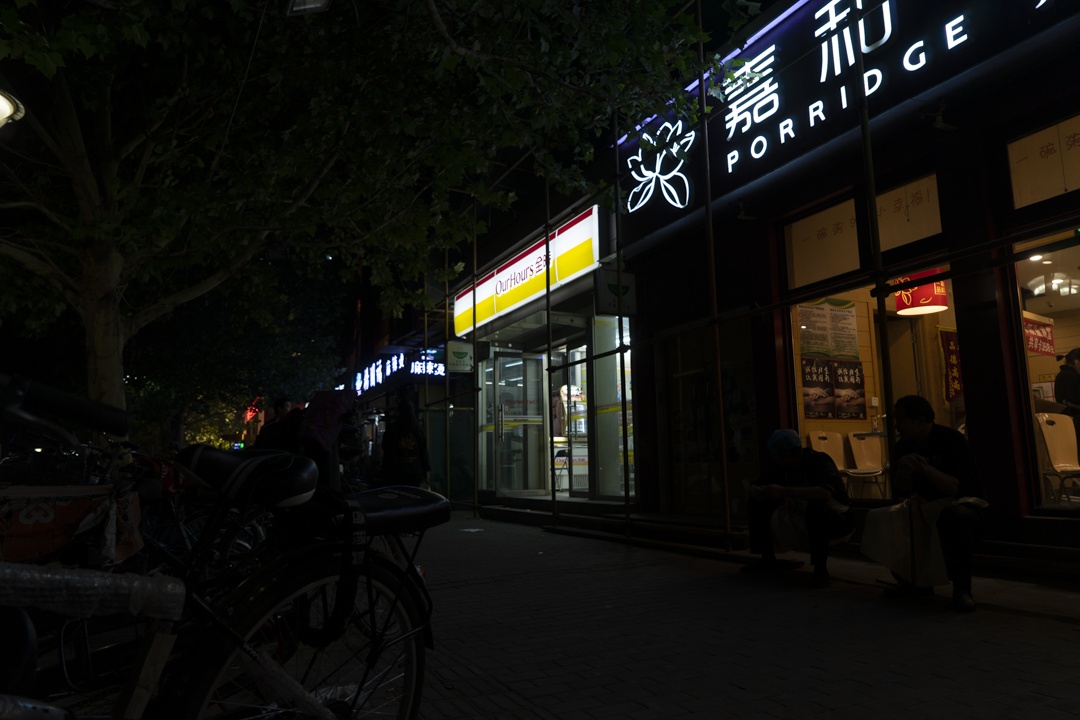}
\hspace*{\shrinkSpaceBetweenImages}&
\includegraphics[height=\heightTeaser]{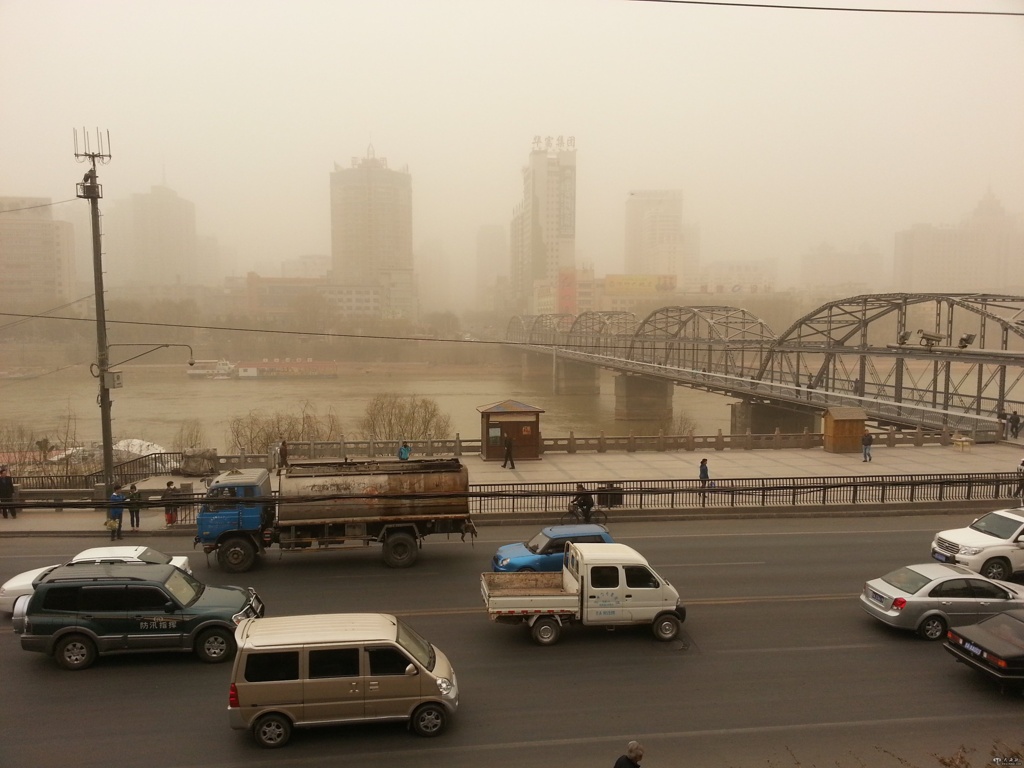}
\hspace*{\shrinkSpaceBetweenImages}&
\includegraphics[height=\heightTeaser]{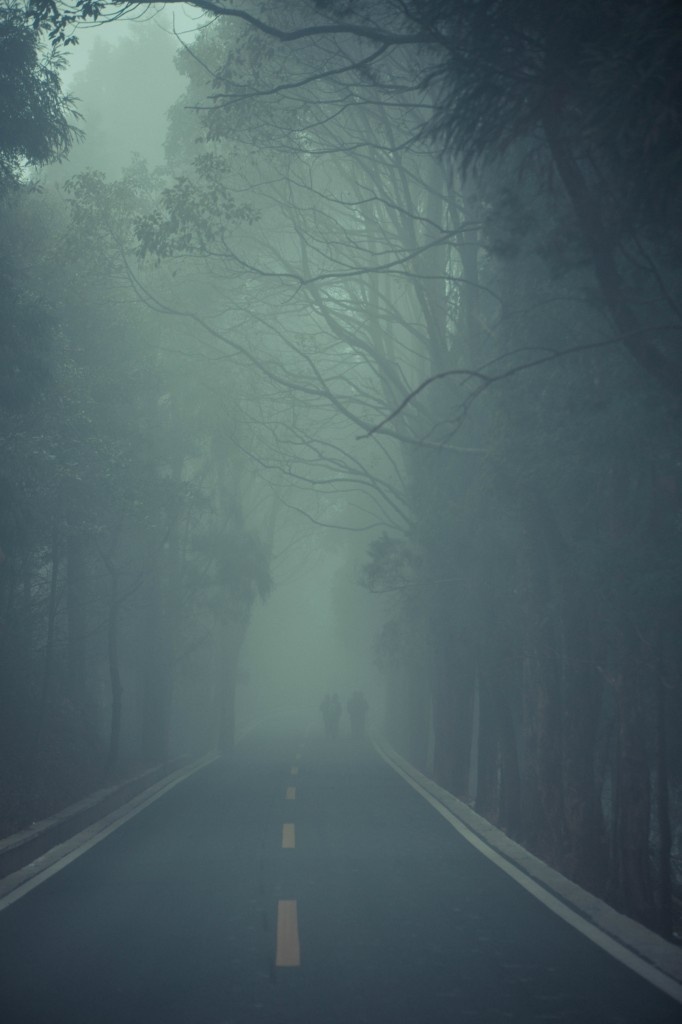}
\hspace*{\shrinkSpaceBetweenImages}&
\includegraphics[height=\heightTeaser]{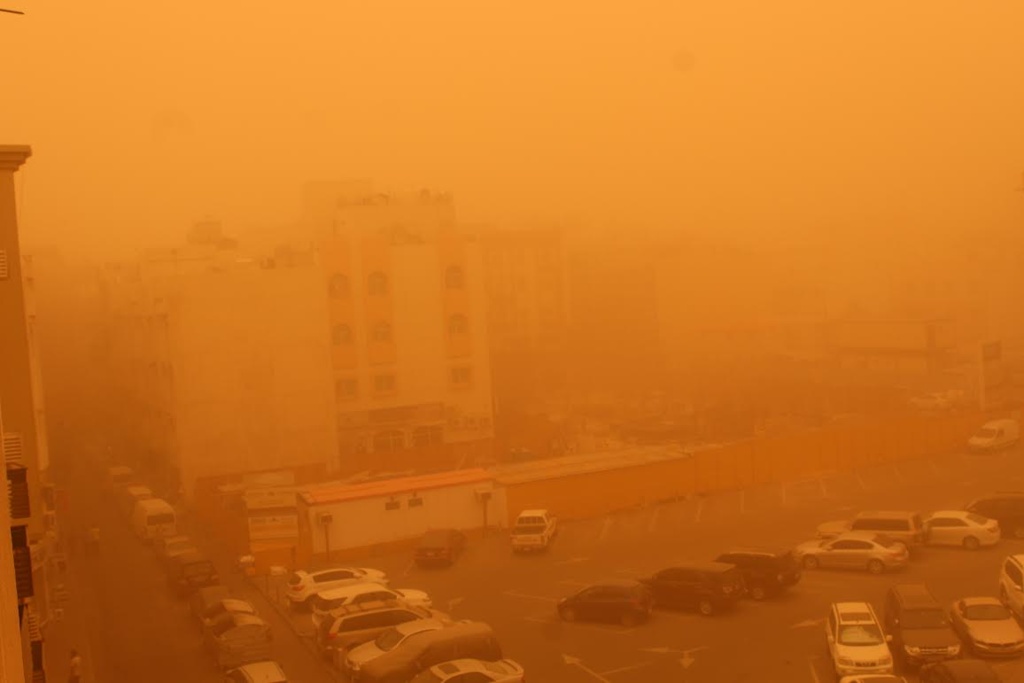}
\hspace*{\shrinkSpaceBetweenImages}&
\includegraphics[height=\heightTeaser]{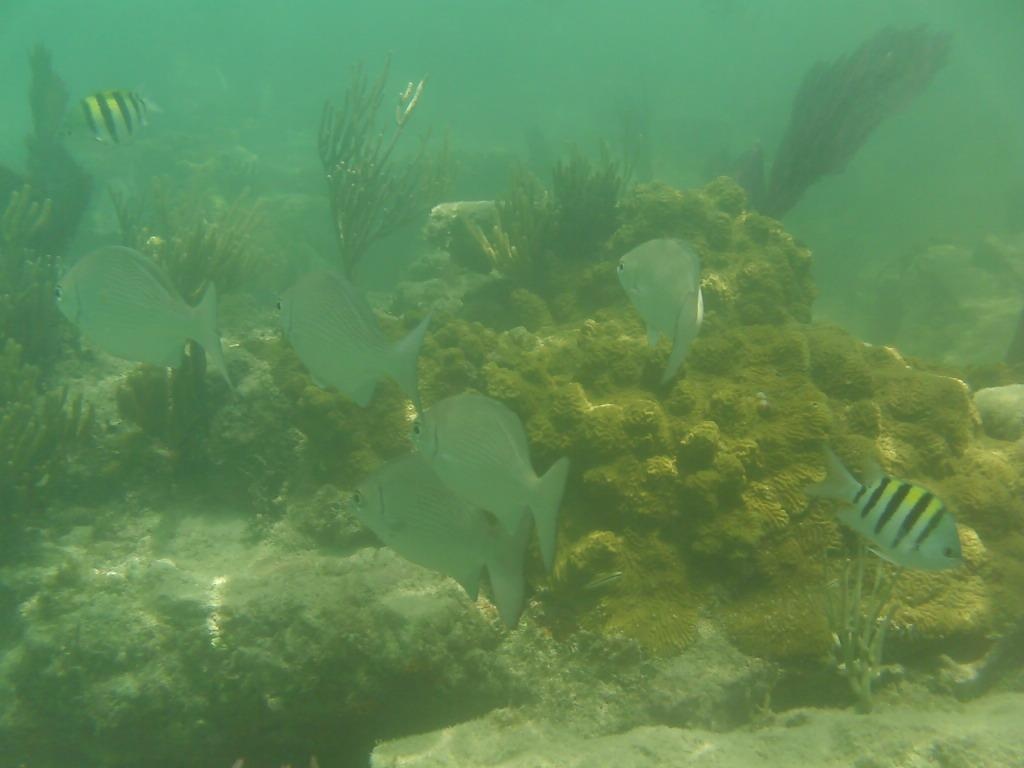}
\\
\includegraphics[height=\heightTeaser]{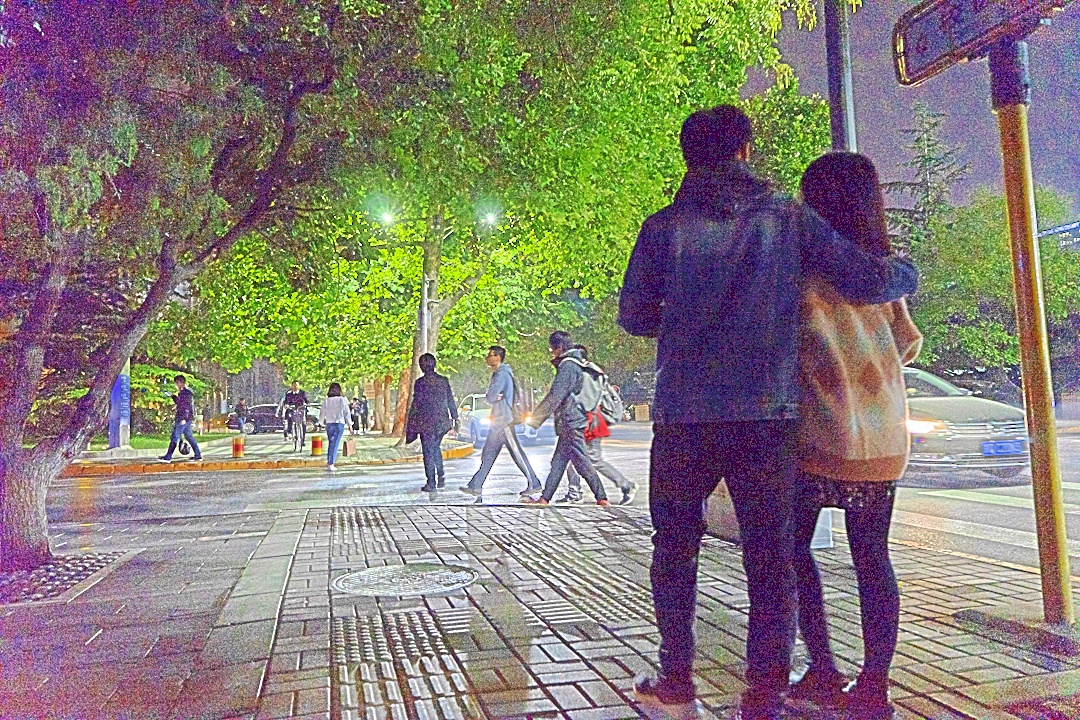}
\hspace*{\shrinkSpaceBetweenImages}&
\includegraphics[height=\heightTeaser]{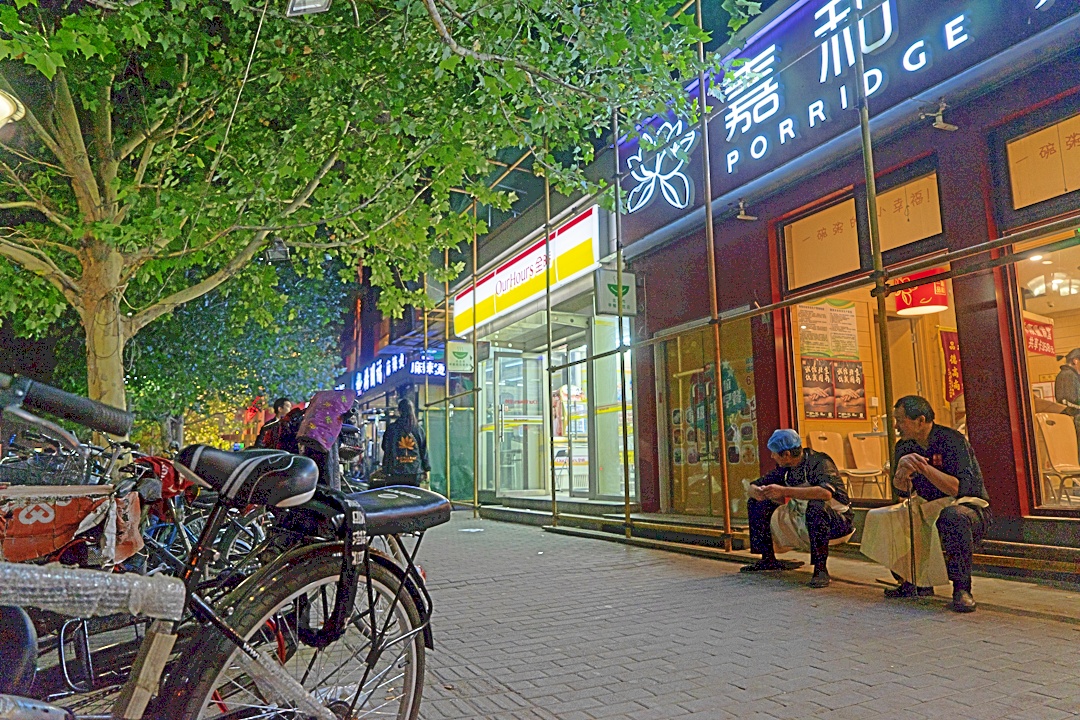}
\hspace*{\shrinkSpaceBetweenImages}&
\includegraphics[height=\heightTeaser]{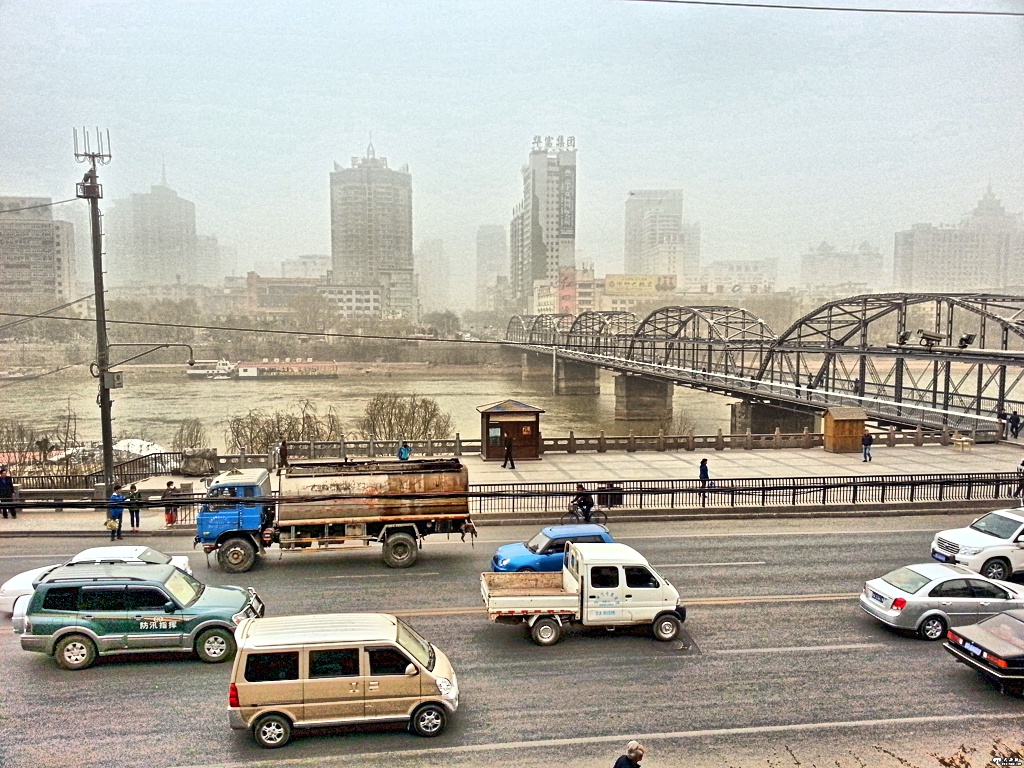}
\hspace*{\shrinkSpaceBetweenImages}&
\includegraphics[height=\heightTeaser]{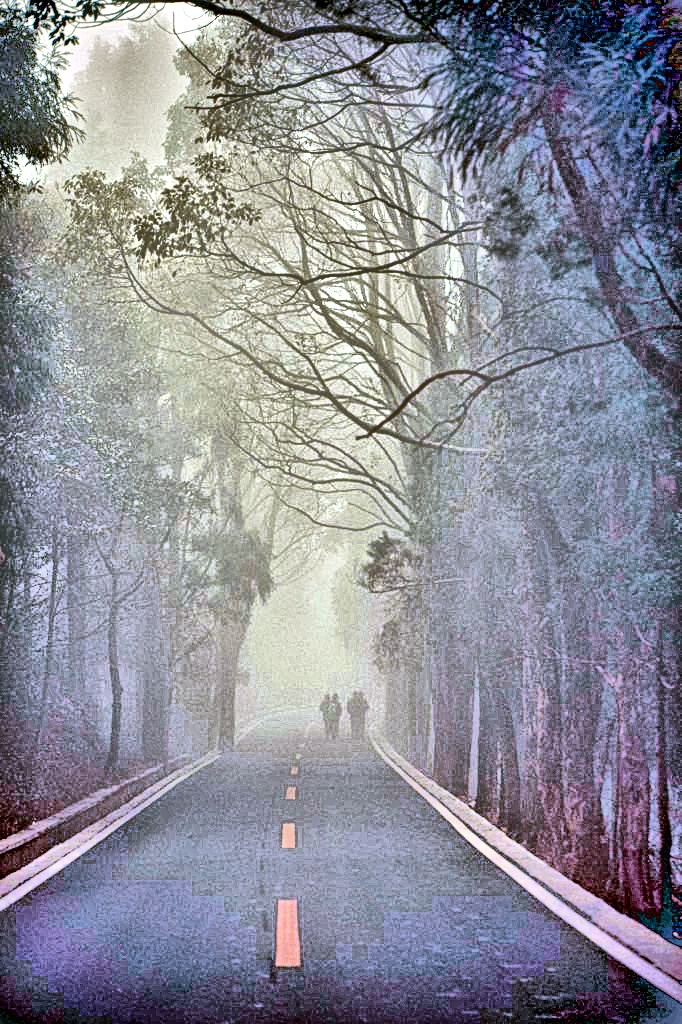}
\hspace*{\shrinkSpaceBetweenImages}&
\includegraphics[height=\heightTeaser]{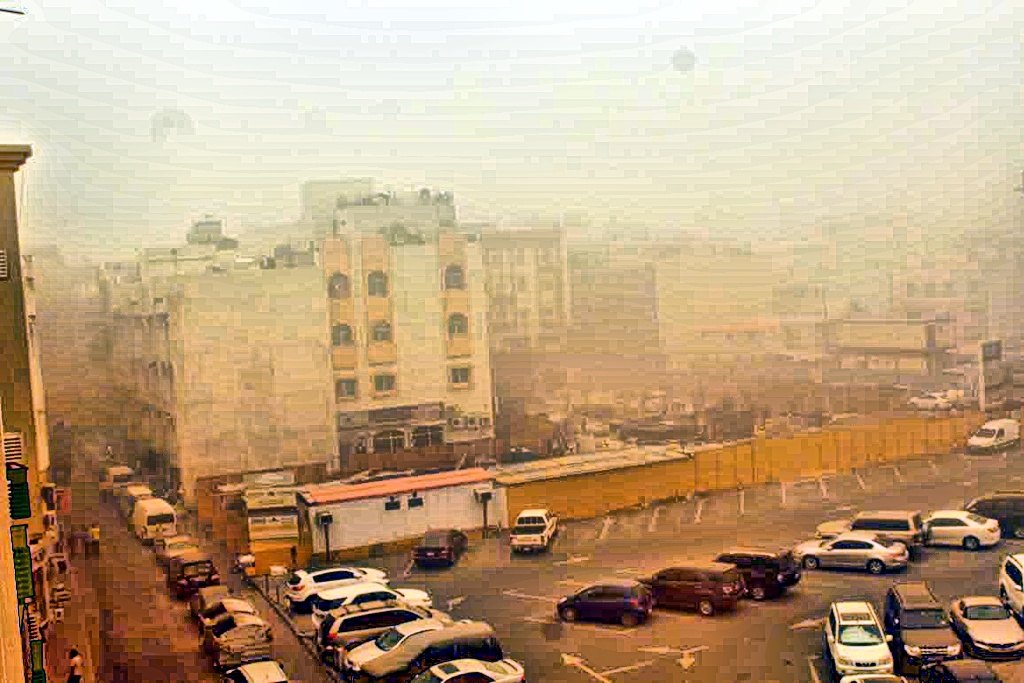}
\hspace*{\shrinkSpaceBetweenImages}&
\includegraphics[height=\heightTeaser]{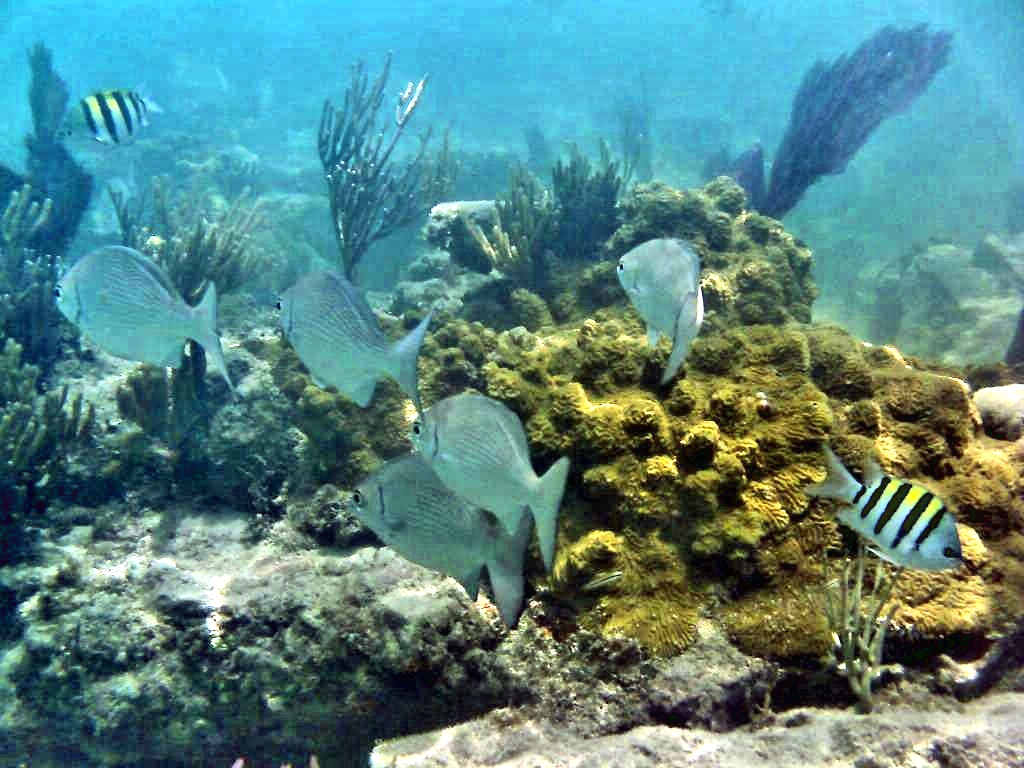}
\end{tabular}\vspace*{-0.3cm}
\caption{The first two images (Dark Face 255 and 259) demonstrate capabilities of our low light image enhancement approach. The third and fourth images (RUSH hazy 004 and 084) illustrate image dehazing capabilities of our approach. In the fifth and sixth images (RUSH sandstorm 057 and RUSH underwater 016) we demonstrate enhanced results of our approach for sandstorm and underwater images.\vspace*{0.5cm}}
\label{fig:teaser}
\end{teaserfigure}
\maketitle

\section{Introduction and Contribution}\label{sec:Intro}
Image enhancement and clarification is an active area of research. General applications include medical imaging (especially endoscopy), autonomous vehicles (navigating in low-light and hazy conditions), surveillance and security (night vision cameras), and many others.

We first focus on the problem of low-light image enhancement (LLIE), for which we provide a simple and efficient method. Then we show how a simple modification of our approach allows us to address image dehazing, i.e. the problem of clarifying hazy/foggy images (including images taken during a sandstorm or underwater).

Recent surveys on LLIE \cite{Li-etal_pami22,kim2022low,rasheed2023comprehensive} provide extensive references to existing methods and potential applications of LLIE techniques. Recent approaches leverage deep learning tools, including transformers \cite{Retinexformer_cvpr23} and diffusion models \cite{LightenDiffusion_eccv24}. Traditional methods (gamma correction, Retinex, and histogram processing) have faded in favor of modern deep learning techniques. Although deep learning has become the dominant approach for LLIE, we propose a remarkably simple alternative. Despite its elegance, our method delivers competitive performance on moderately dark images and outperforms state-of-the-art techniques when applied to extremely dark images.

Although the problems might seem unrelated, it turns out that our approach can also be adapted to enhance images with fog and haze. As is common nowadays, deep learning-based methods have been extensively investigated for clarifying turbid imagery. We refer to the recent survey \cite{Gui_survey2023} for references. However, unlike LLIE, some of the best-performing methods \cite{Guo-etal_tmcca23,Liu-etal_pami23,Ling-etal_tip23} are not deep learning-based. We show that our approach can be adapted to further improve the results of these methods.

\section{Reverse Filtering for Image Enhancement}\label{sec:RF}
Our approach is inspired by the approach known as reverse filtering \cite{Tao-etal_iccv17, Milanfar_siims18}. Reverse image filtering involves reconstructing an original image from its filtered version using only the filter itself. We introduce a reverse filtering method tailored to enhance images captured under low-light conditions. Using a haze-simulating filter instead of a darkening filter, our method can also handle clarification and enhancement of hazy images.

\subsection*{Reversing Low Light Illumination}
Considering grayscale images with intensity values between zero and one, a straightforward procedure to darken a given image is to multiply its pixel values by themselves. It causes bright regions to become only slightly dimmer, while darker areas are significantly darkened. The procedure can be easily extended as follows. Consider an image filter $f(\cdot)$ which suppresses small-scale image details. Then darkening an image $\x$ can be done by transforming $\x$ into $\x f(\x)^p$ with $0<p\leq1$. A simple reversing procedure brightens a given low-light image $\x$  by transforming it into $\x\,/\left[f(\x)^p+r\right]$, where $r$ is a small regularization parameter used to avoid division by zero.


Experiments confirm that choosing $f(\cdot)$ as the guided filter \cite{He-Sun-Tang_pami12} combined with gamma correction with the exponent in the range of $0.8\leq p\leq1.0$, produces natural and visually pleasing images. A notable advantage of the guided filter over other commonly used edge-preserving image filters, including total variation and bilateral filters, is its ability to avoid gradient reversal artifacts like solarization.

Like several other methods, our approach begins by converting an input RGB-color image to Hue-Saturation-Value (HSV) space, modifies the value component, and then transforms the modified image back to RGB.

The MATLAB function \texttt{imBeam}, implementing our LLIE method, is shown in Fig.\,\ref{fig:matlab_imBeam}.

\vspace*{-0.2cm}
\begin{figure}[!h!t!b!p]
\begin{center}
\fbox{\footnotesize
\begin{lstlisting}[language=Matlab]
f_GF = @(x)imguidedfilter(x);

function Out = imBeam(In,f)
  hsv = rgb2hsv(In); v = hsv(:,:,3);
  p = 1-0.2*(0.5+atan(100*mean(v(:))-5)/pi);
  % p = 0.5; used for image clarification
  r = 0.01; % to avoid division by zero
  v = v./(f(v).^p+r);
  % if additional sharpening is needed:
  % v = (v+adapthisteq(v))*0.5;
  hsv(:,:,3) = v; Out = hsv2rgb(hsv);
end

function out = pp(in)
  c_min = 1; c_max = 99;
  in_min = prctile(in,c_min,[1 2]);
  in_max = prctile(in,c_max,[1 2]);
  out = (in - in_min)./(in_max-in_min);
end

\end{lstlisting}}
\end{center}
\vspace*{-0.2cm}
\caption{MATLAB script for image brightening \texttt{imBeam} and simple post-processing scheme \texttt{pp}.}
\label{fig:matlab_imBeam}
\vspace*{-0.2cm}
\end{figure}

Another justification for using $v/f(v)^p$, where $f(\cdot)$ stands for a filter suppressing small-scale image details, is as follows. Filter $f(\cdot)$ decomposes the image $v$ into base and detail components:
\begin{equation}\label{eq:decomposition}
v=b+d,\quad b=f(v),\quad v/f(v)^p=b^{1-p}+d/b^p,
\end{equation}
where the first term, $b^{1-p}$, corresponds to the $\gamma$-correction of the base component with $\gamma = 1-p$ (so the closer $p$ is to $1$, the stronger image brightening is applied) and the second term, $d/b^p$, represents the enhanced image detail.

Image decomposition (\ref{eq:decomposition}) links our approach with Land's Retinex theory \cite{Land1977retinex}.

Despite its simplicity, the proposed  image brightening scheme performs well, rivaling state-of-the-art LLIE methods for extremely low-light images. Experiments indicate that for very dark images, setting the gamma correction parameter $p$ close to or equal to $1$ produces the best results, while values closer to $0.8$ deliver competitive performance for moderately low light images. These findings suggest automatically selecting the exponent based on the mean value of the processed image's value component.

\subsection*{Reducing Haze and Fog}
The ``reverse darkening'' approach adapted above for LLIE can be used to ``heal'' other types of image distortion. We note that the adaptive manifold filter (AMF) \cite{Gastal-Oliveira_sig12} with specific parameter values can be used to simulate fog and haze effects, similar to those seen in \cite[Fig.\,6d]{Gastal-Oliveira_sig12}. Based on this observation, we can use a slight modification of \texttt{imBeam} to reverse a hazing\,/\,fog effect by using an AMF filter with appropriate parameter values (AMF with $\sigma_s=20$ and $\sigma_r=0.4$ is used in our experiments). More specifically, a numerical inversion of the AMF filter can be done by multiplicative fixed point iterations
\begin{equation}\label{eq:Gold}
\x_{k+1} = \x_k\cdot\left[\y\,/{\texttt{AMF}(\x_k)}\right],
\end{equation}
where pixel-wise multiplications and divisions are assumed and $y$ is the observed hazy image, also used as the iteration starting image. The iterative process (\ref{eq:Gold}) is a simple modification of Gold's method \cite{gold1964iterative} developed for image deblurring purposes. One iteration of (\ref{eq:Gold}) followed by gamma correction with $\gamma=1/2$ yields
\begin{equation}\label{eq:Gold1sqrt}
\x=\sqrt{\y\,^2/{\texttt{AMF}(\y)}}=\y\,/\sqrt{\texttt{AMF}(\y)}
\end{equation}
which produces a reasonably good suppression of haze\,/\,fog image distortions. In practice, we apply (\ref{eq:Gold1sqrt}) only to the V-channel in the HSV-color space. In addition to using an AMF filter, we also use Contrast Limited Adaptive Histogram Equalization (CLAHE) \cite{Zuiderveld1994clahe} to further sharpen details (see commented lines in Fig.\,\ref{fig:matlab_imBeam}) and apply a simple post-processing scheme consisting of percentile-based image stretching (see again Fig.\,\ref{fig:matlab_imBeam} for the post-processing scheme used).

Images from hazy, underwater, or sand-dust conditions often suffer from color distortions (underwater images tending to appear greenish, and sand-dust images tending to have an orange cast). The post-processing scheme reduces color distortions by stretching linearly the image's colors. Similar percentile-based image stretching schemes are used in \cite{LEP_tip12,Liu-etal_pami23}.

We call the resulting image filter by \imBeamAMF. Let us denote by \imBeamGF the same image filtering scheme but with the Guided Filter (GF) \cite{He-Sun-Tang_pami12} instead of AMF, so $\y\,/\sqrt{\texttt{GF}(\y)}$ is used instead of (\ref{eq:Gold1sqrt}). For the image dehazing task, we combine \imBeamAMF with \imBeamGF, as the latter works as an enhancing filter, sharpening a given image and brightening its dark areas.

\section{Experimental results}
\label{sec:experiments}

\begin{figure*}[!h!t]
\hspace*{-0.2cm}\centering
\begin{tabular}{cccccccc}
& {\scriptsize Initial image}
& {\scriptsize SCI} & {\scriptsize LightenDiffusion} & {\scriptsize RetinexFormer}
& {\scriptsize CoLIE} & {\scriptsize STAR} & {\scriptsize \texttt{imBeam}}
\\
\rotatebox{90}{\scriptsize Dark Face 3797}
\hspace*{-0.5cm} &
\includegraphics[height=\heightFigDarkFace]{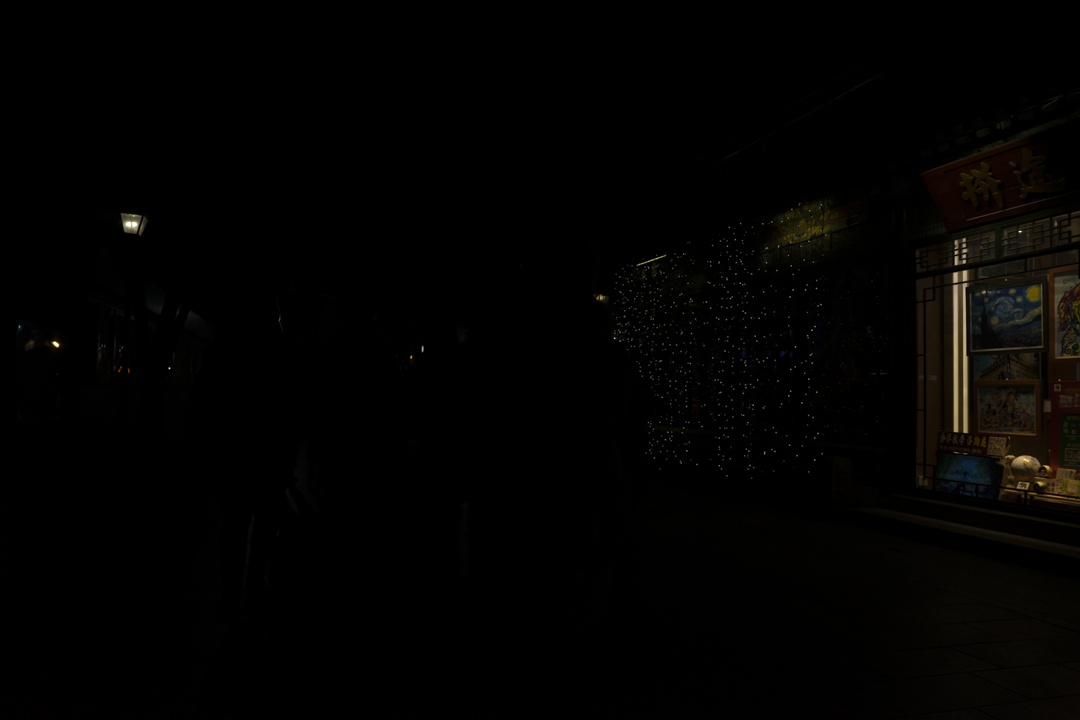}
\hspace*{\shrinkSpaceBetweenImages} &
\includegraphics[height=\heightFigDarkFace]{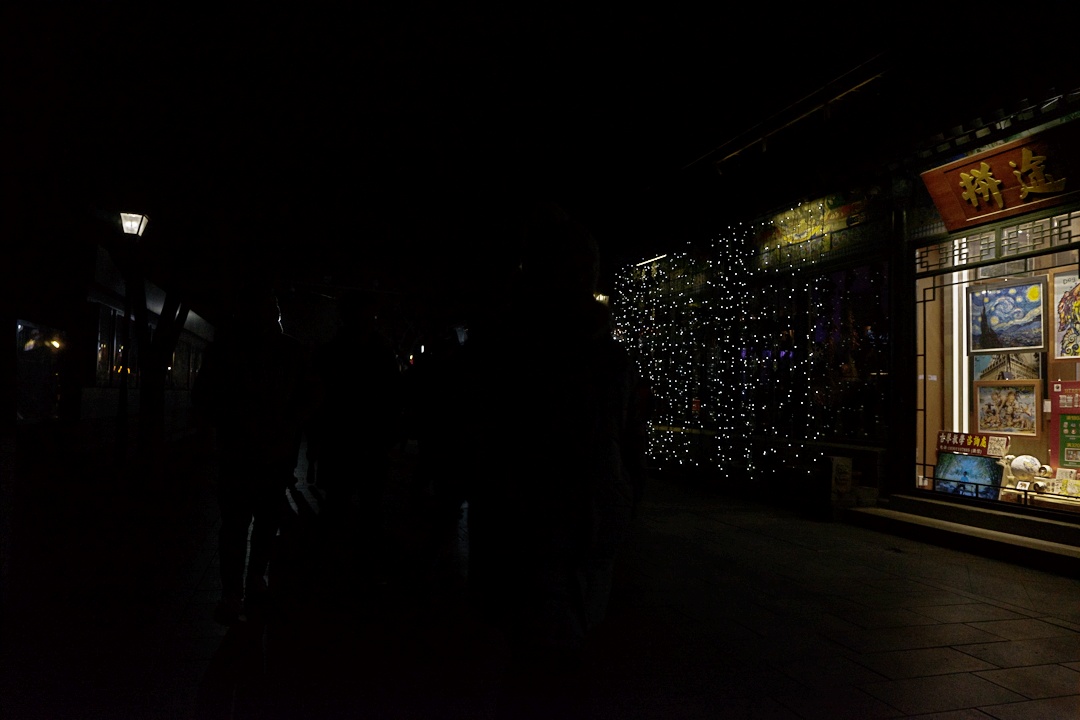}
\hspace*{\shrinkSpaceBetweenImages} &
\includegraphics[height=\heightFigDarkFace]{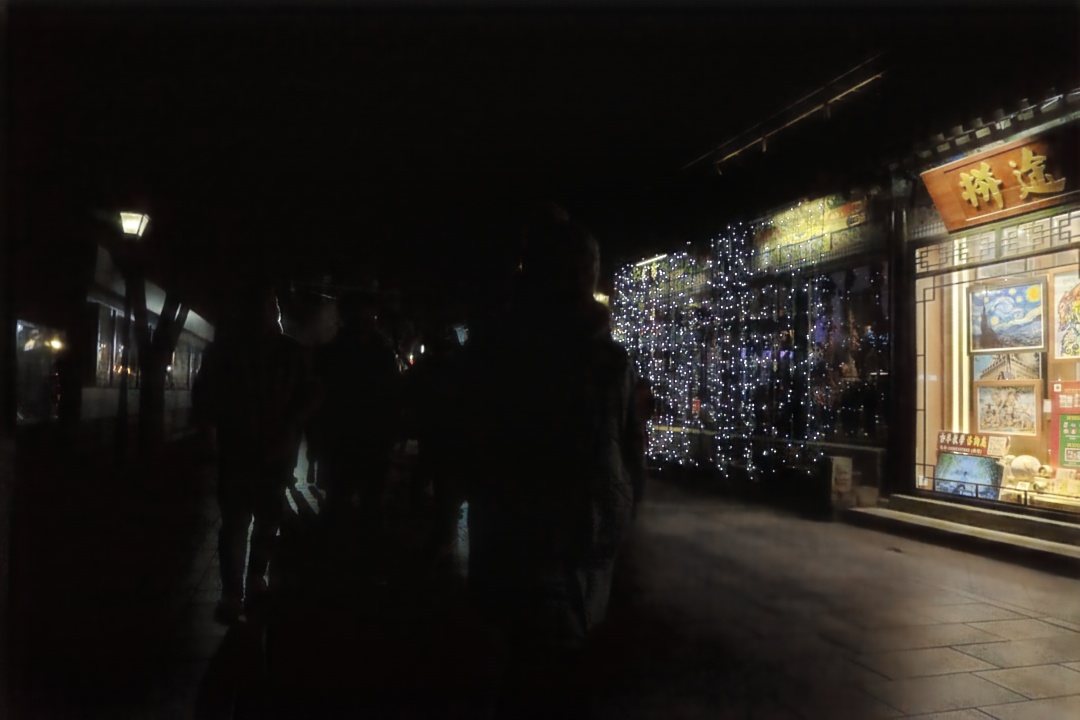}
\hspace*{\shrinkSpaceBetweenImages} &
\includegraphics[height=\heightFigDarkFace]{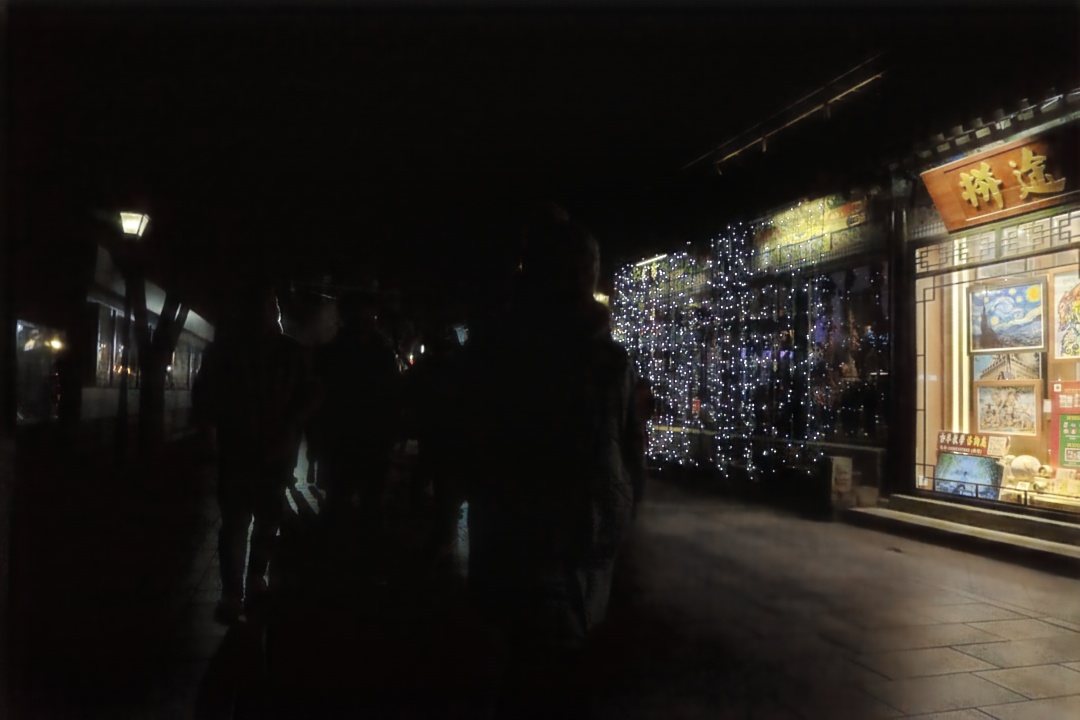}
\hspace*{\shrinkSpaceBetweenImages} &
\includegraphics[height=\heightFigDarkFace]{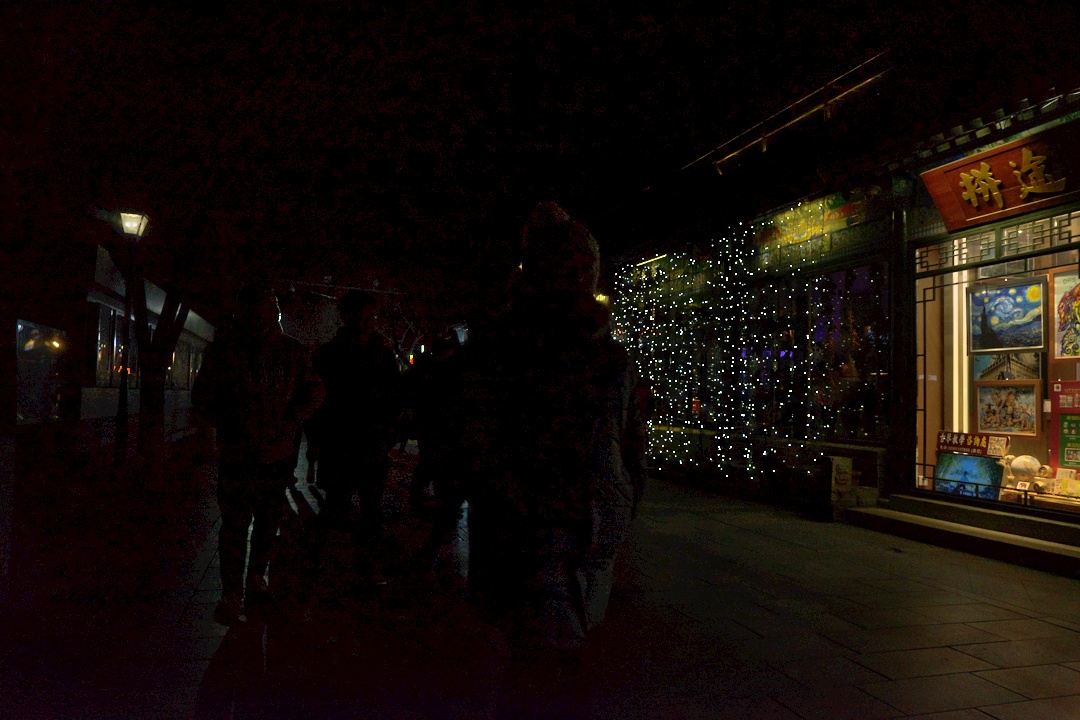}
\hspace*{\shrinkSpaceBetweenImages} &
\includegraphics[height=\heightFigDarkFace]{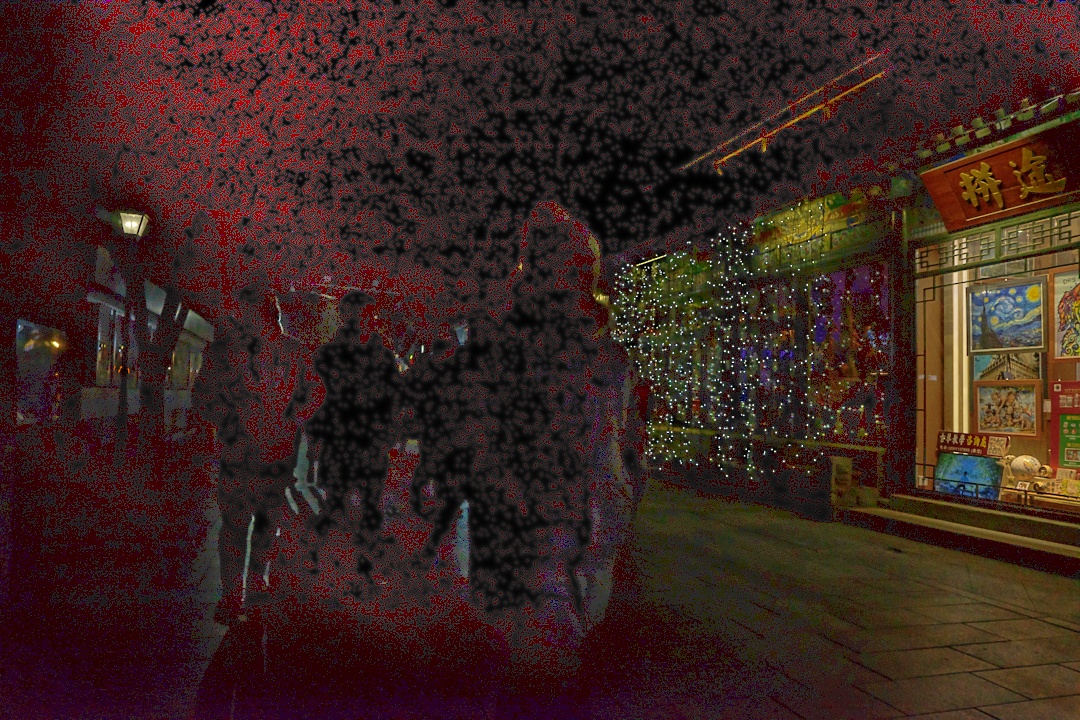}
\hspace*{\shrinkSpaceBetweenImages} &
\includegraphics[height=\heightFigDarkFace]{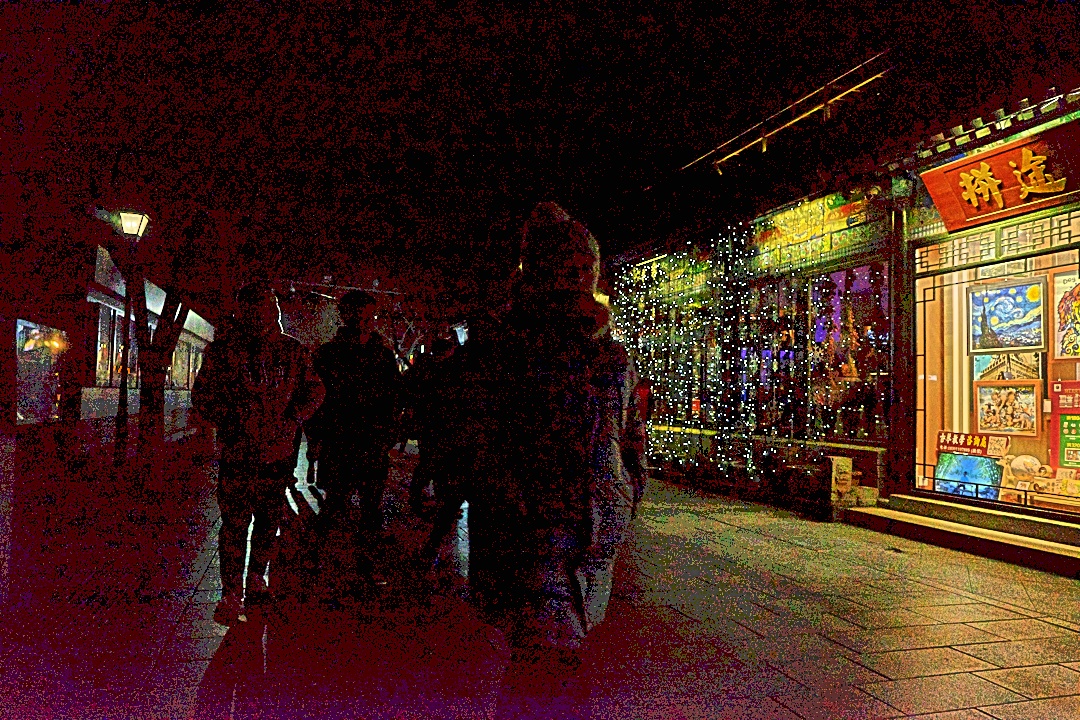}
\\
\rotatebox{90}{\scriptsize Dark Face 5000}
\hspace*{-0.5cm} &
\includegraphics[height=\heightFigDarkFace]{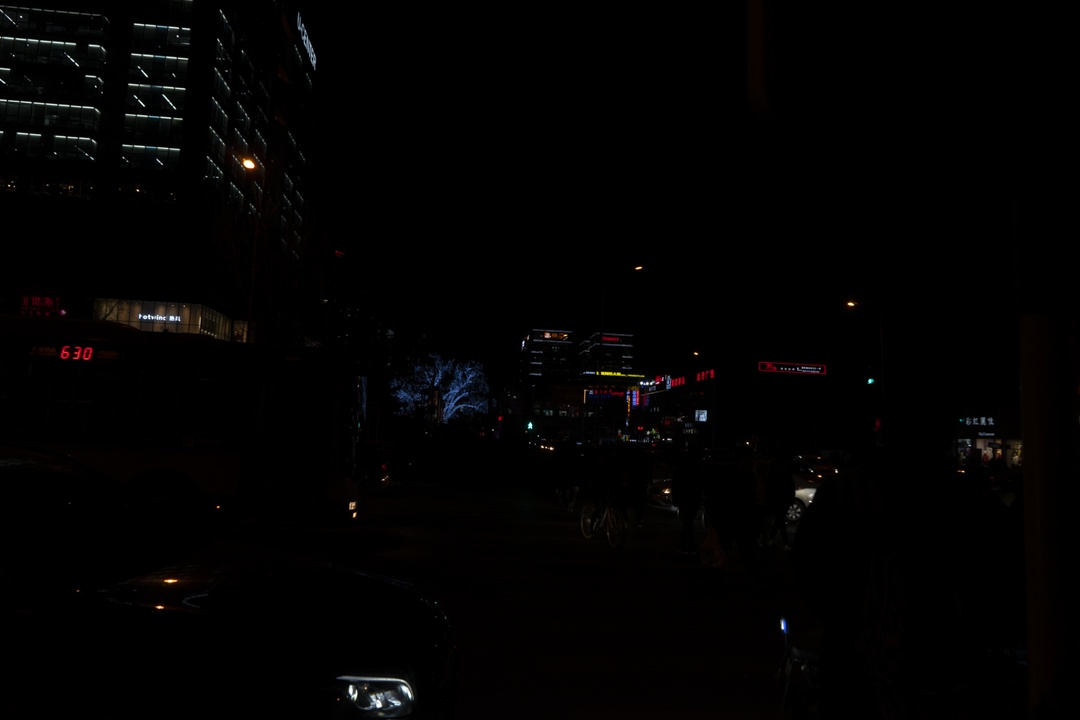}
\hspace*{\shrinkSpaceBetweenImages} &
\includegraphics[height=\heightFigDarkFace]{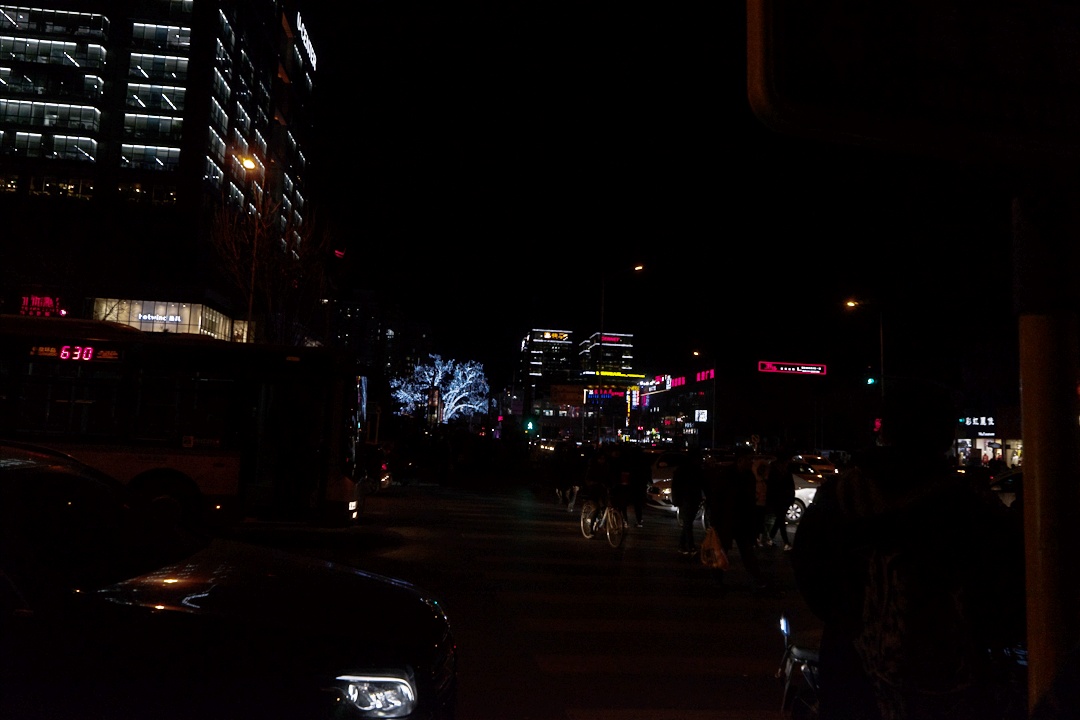}
\hspace*{\shrinkSpaceBetweenImages} &
\includegraphics[height=\heightFigDarkFace]{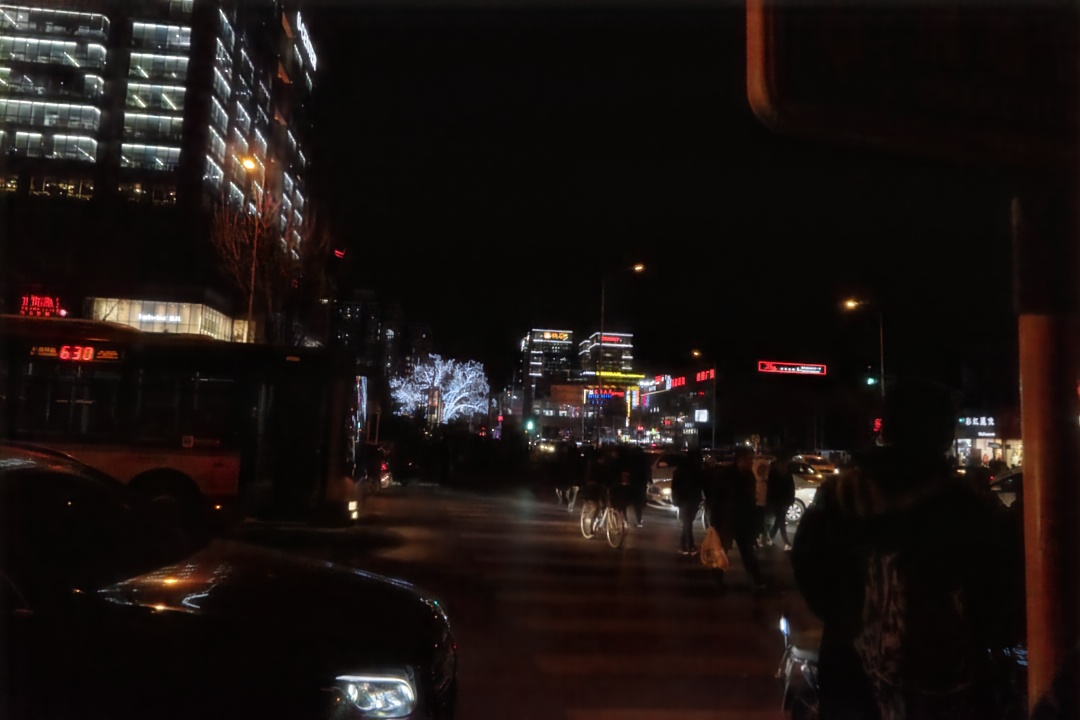}
\hspace*{\shrinkSpaceBetweenImages} &
\includegraphics[height=\heightFigDarkFace]{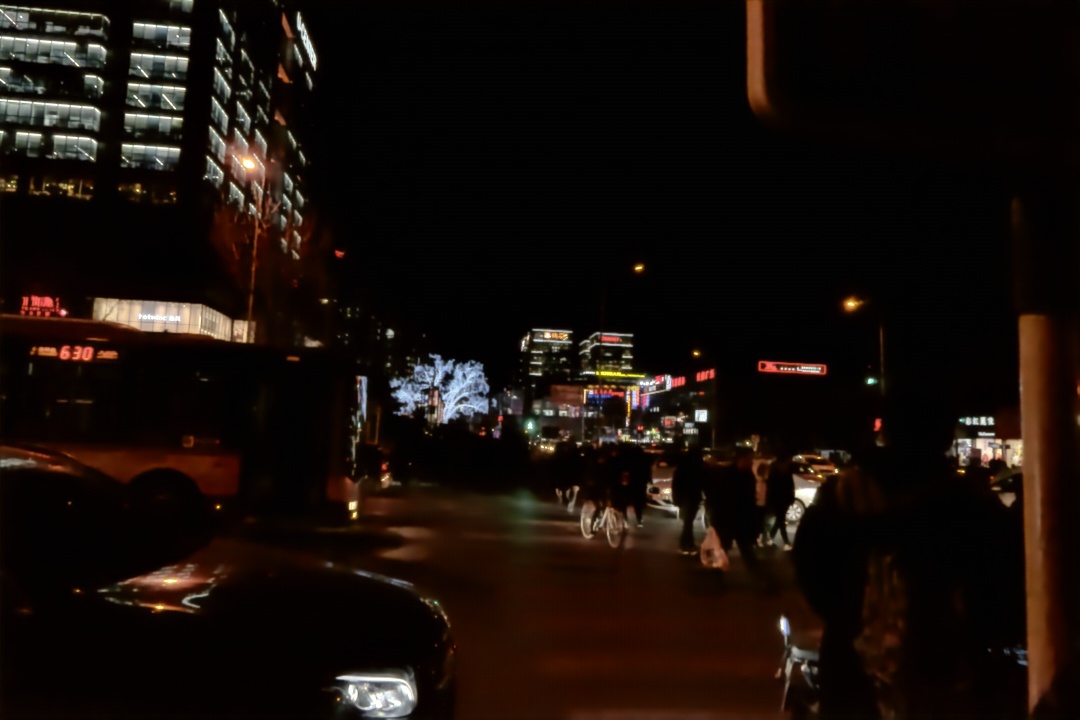}
\hspace*{\shrinkSpaceBetweenImages} &
\includegraphics[height=\heightFigDarkFace]{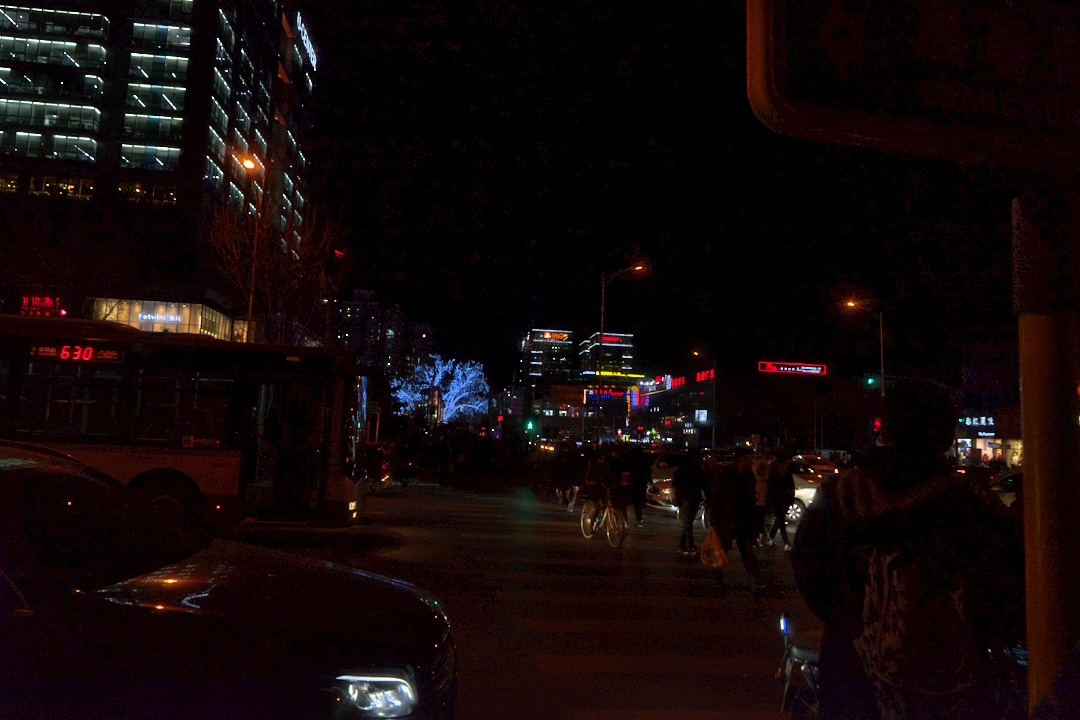}
\hspace*{\shrinkSpaceBetweenImages} &
\includegraphics[height=\heightFigDarkFace]{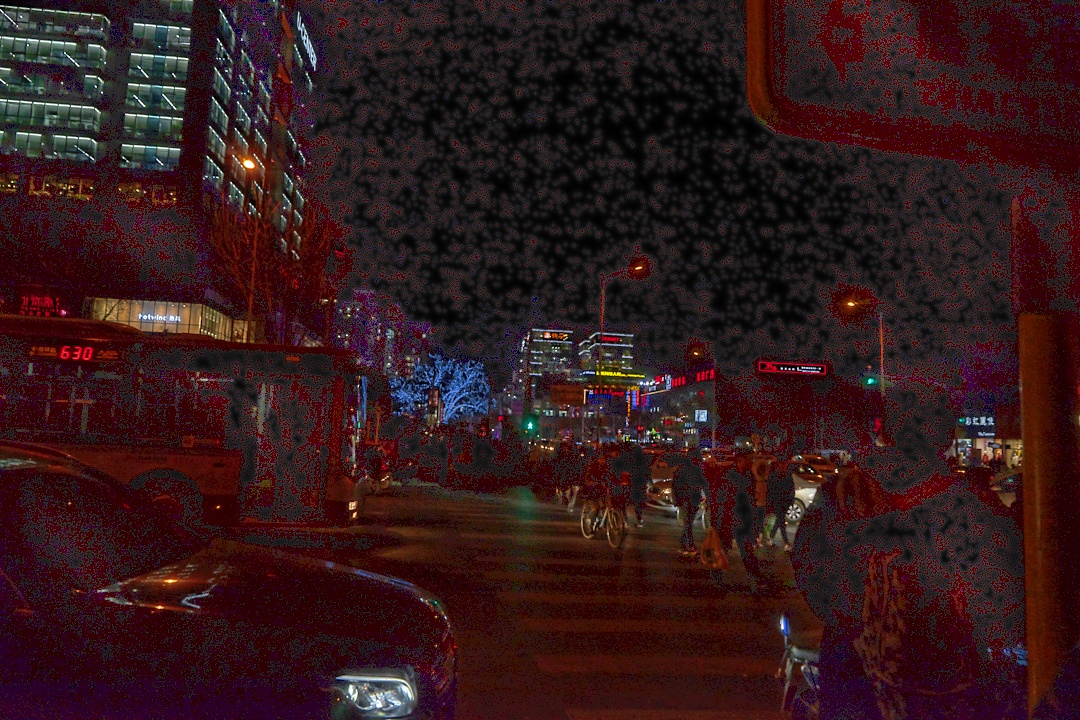}
\hspace*{\shrinkSpaceBetweenImages} &
\includegraphics[height=\heightFigDarkFace]{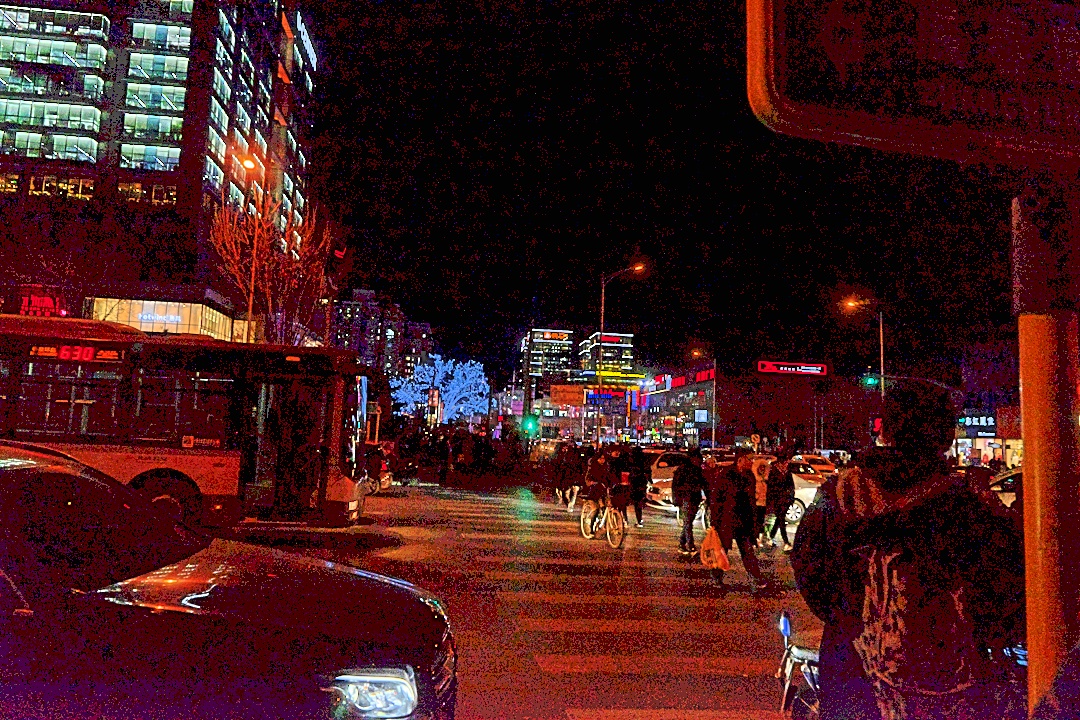}
\\
\rotatebox{90}{\scriptsize Dark Face 540}
\hspace*{-0.5cm} &
\includegraphics[height=\heightFigDarkFace]{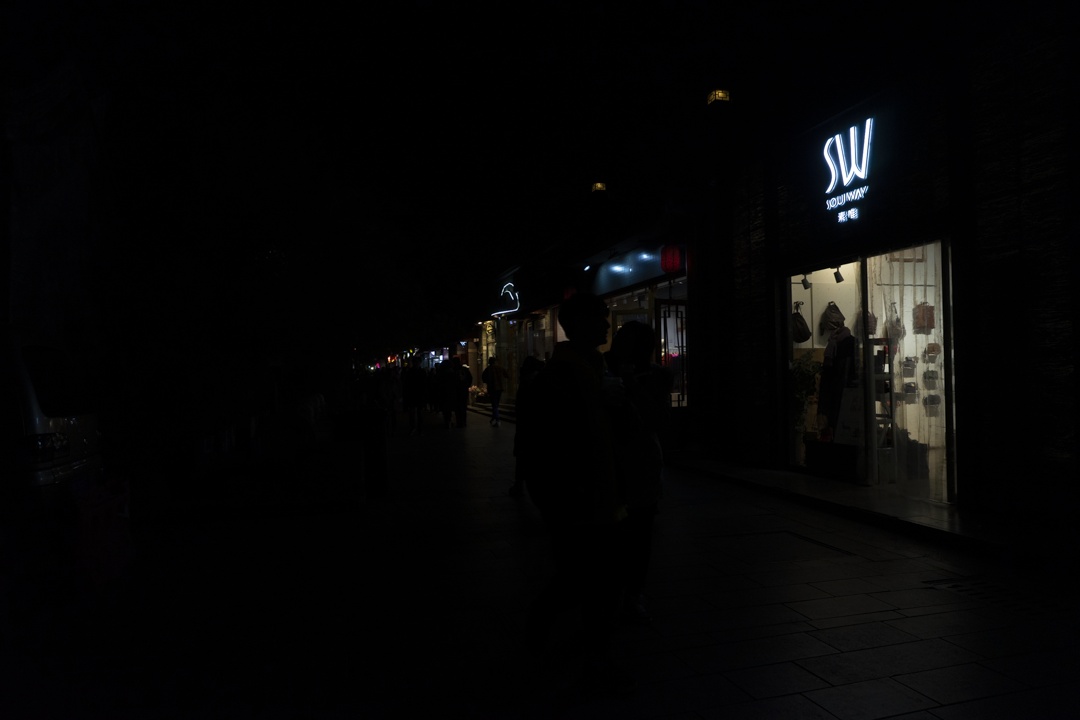}
\hspace*{\shrinkSpaceBetweenImages} &
\includegraphics[height=\heightFigDarkFace]{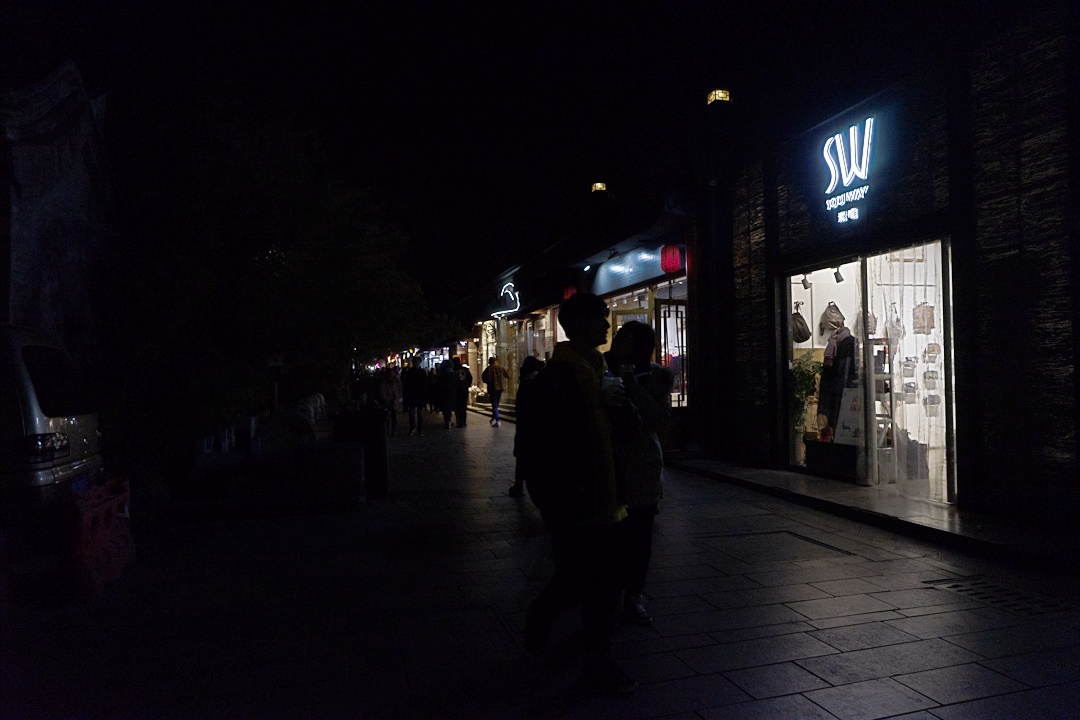}
\hspace*{\shrinkSpaceBetweenImages} &
\includegraphics[height=\heightFigDarkFace]{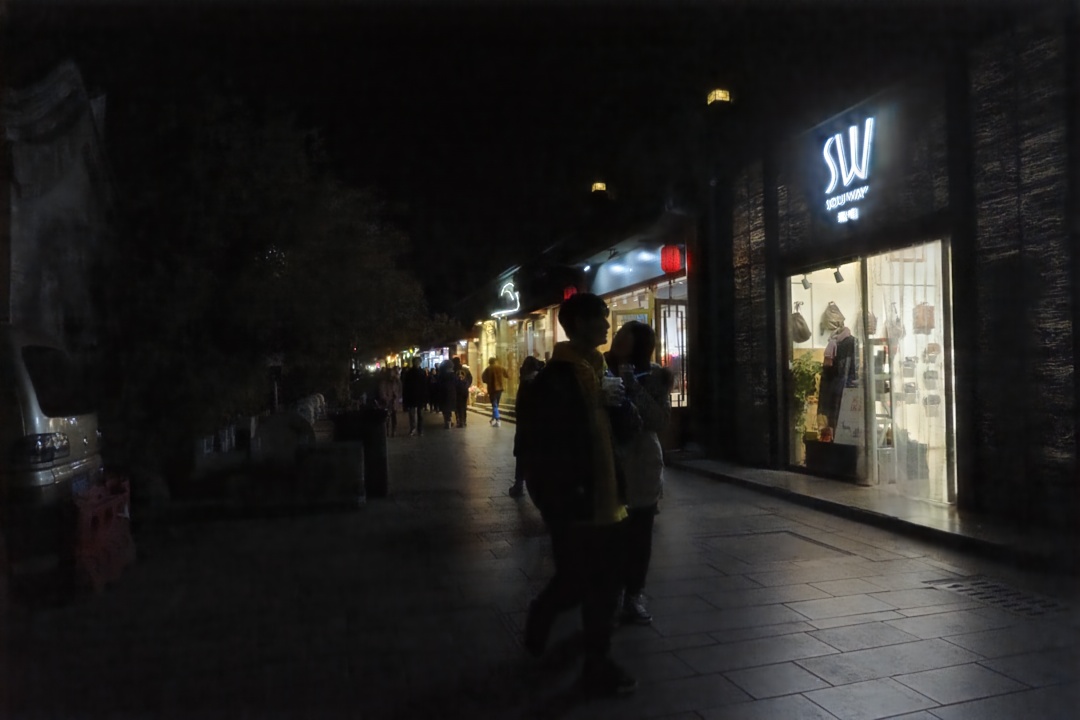}
\hspace*{\shrinkSpaceBetweenImages} &
\includegraphics[height=\heightFigDarkFace]{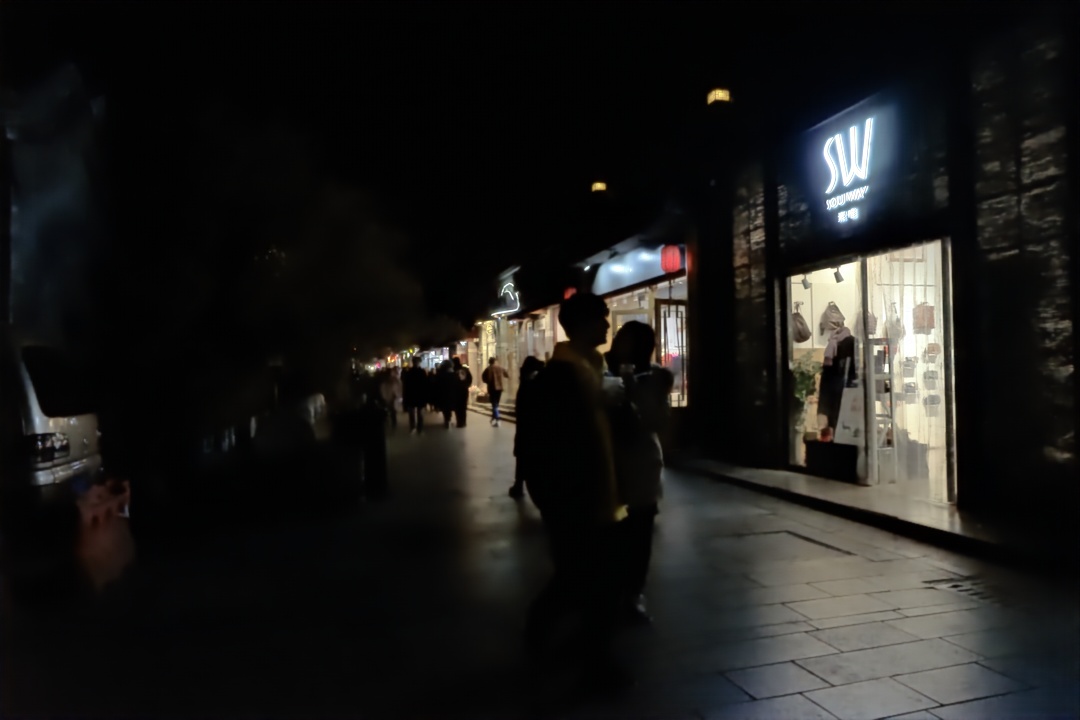}
\hspace*{\shrinkSpaceBetweenImages} &
\includegraphics[height=\heightFigDarkFace]{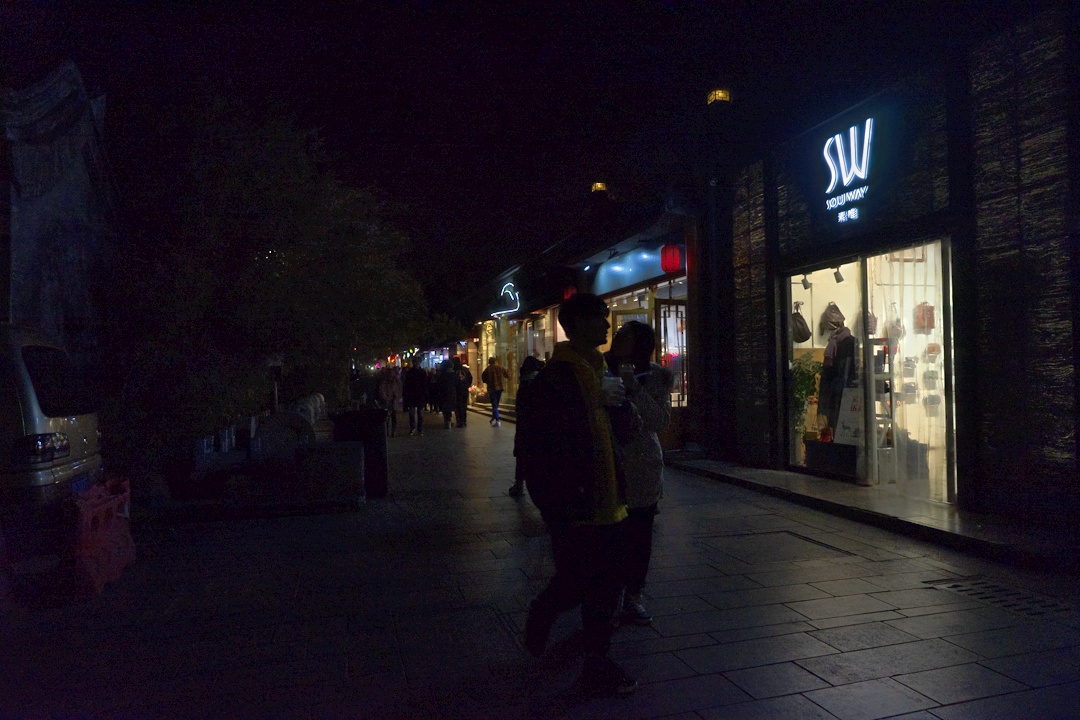}
\hspace*{\shrinkSpaceBetweenImages} &
\includegraphics[height=\heightFigDarkFace]{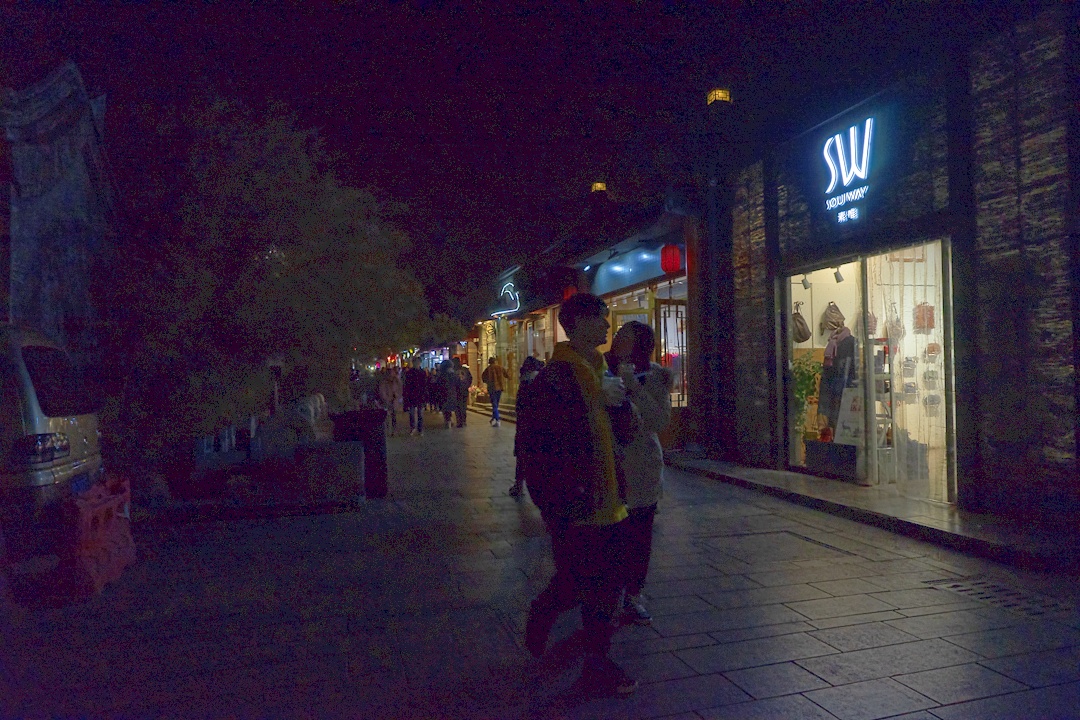}
\hspace*{\shrinkSpaceBetweenImages} &
\includegraphics[height=\heightFigDarkFace]{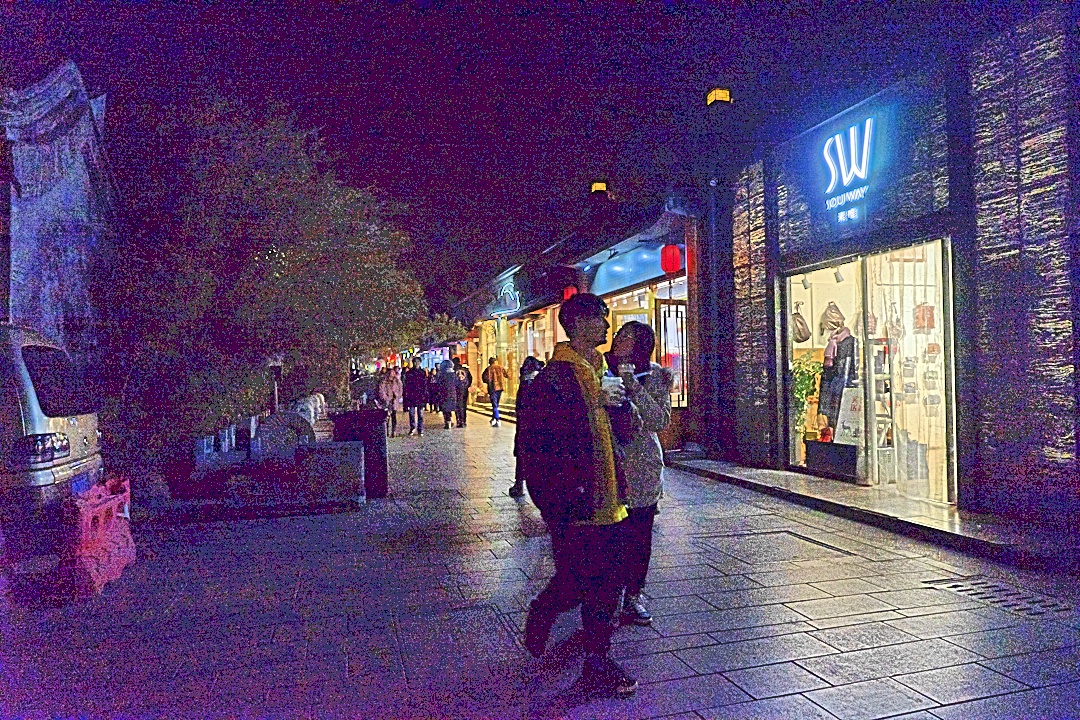}
\end{tabular}\vspace*{-0.3cm}
\caption{Visual comparison of the proposed method with several popular LLIE schemes. While these schemes show excellent performance for not-so-dark images, they are not able to match the performance of \texttt{imBeam} for extremely dark images.}
\label{fig:LLcomparison}
\vspace*{-0.2cm}
\end{figure*}

\subsection*{Low Light Image Enhancement}
\begin{figure*}[h!t!]
\hspace*{-0.2cm}\centering
\begin{tabular}{cccccccc}
& {\scriptsize Dark Face 8} & {\scriptsize Dark Face 9} & {\scriptsize Dark Face 11}
& {\scriptsize Dark Face 101} & {\scriptsize Dark Face 1353} & {\scriptsize Dark Face 1462} & {\scriptsize Dark Face 5720}
\\
\rotatebox{90}{\scriptsize original photo}
\hspace*{-0.5cm} &
\includegraphics[height=\heightFigDarkFaceAdaptive]{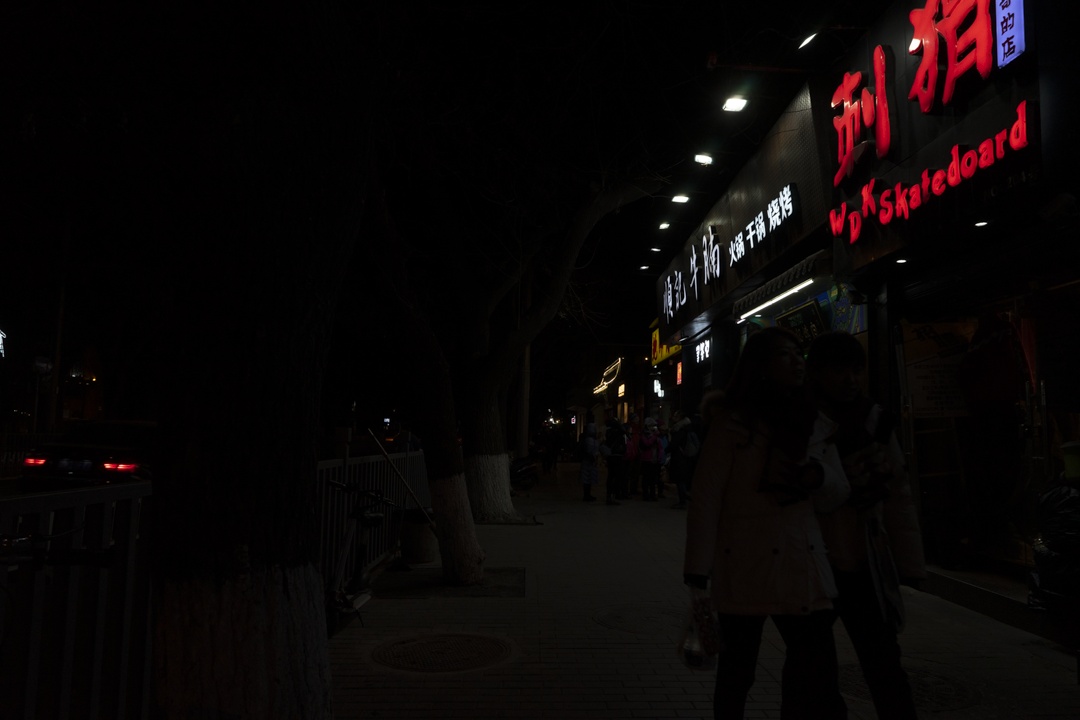}
\hspace*{\shrinkSpaceBetweenImages} &
\includegraphics[height=\heightFigDarkFaceAdaptive]{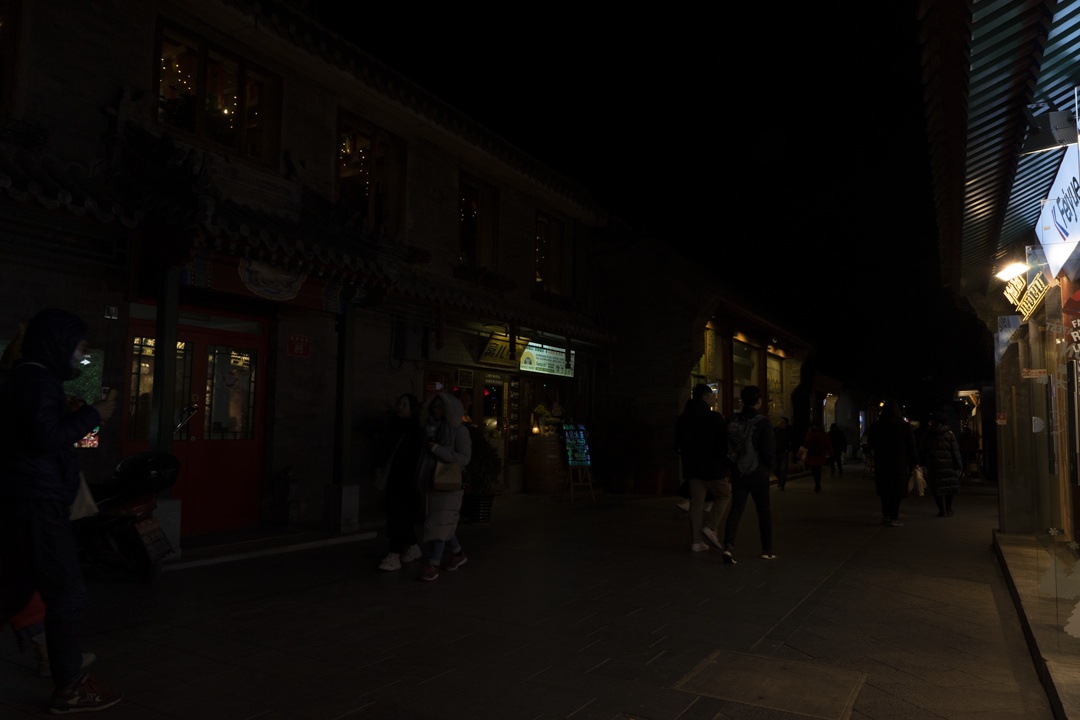}
\hspace*{\shrinkSpaceBetweenImages} &
\includegraphics[height=\heightFigDarkFaceAdaptive]{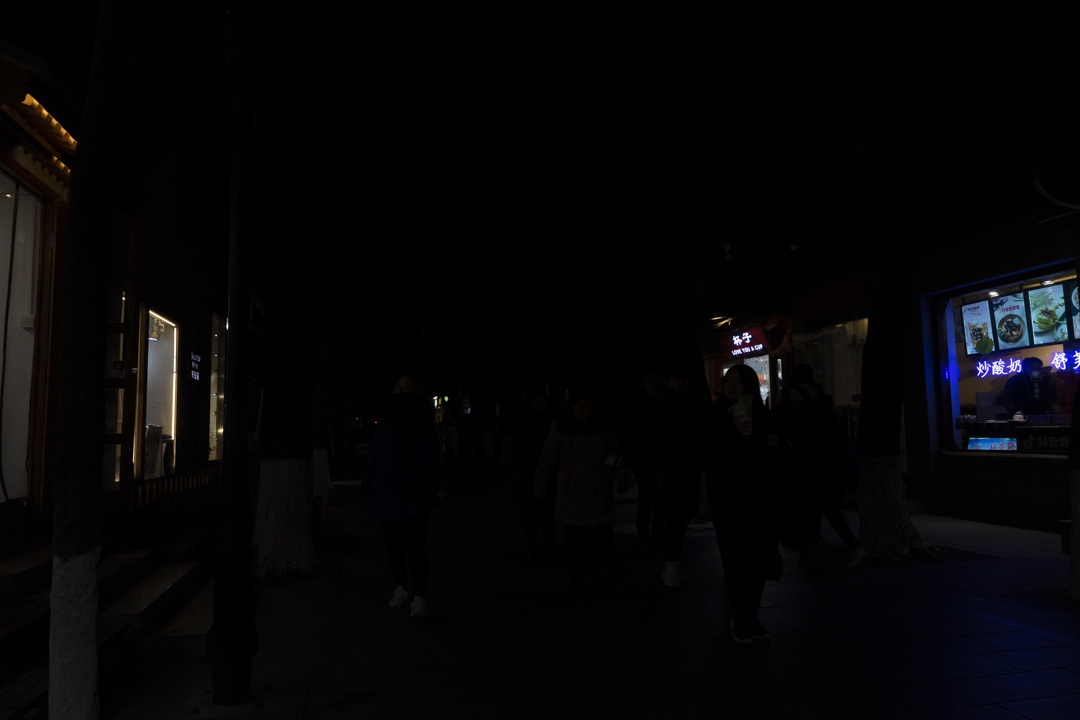}
\hspace*{\shrinkSpaceBetweenImages} &
\includegraphics[height=\heightFigDarkFaceAdaptive]{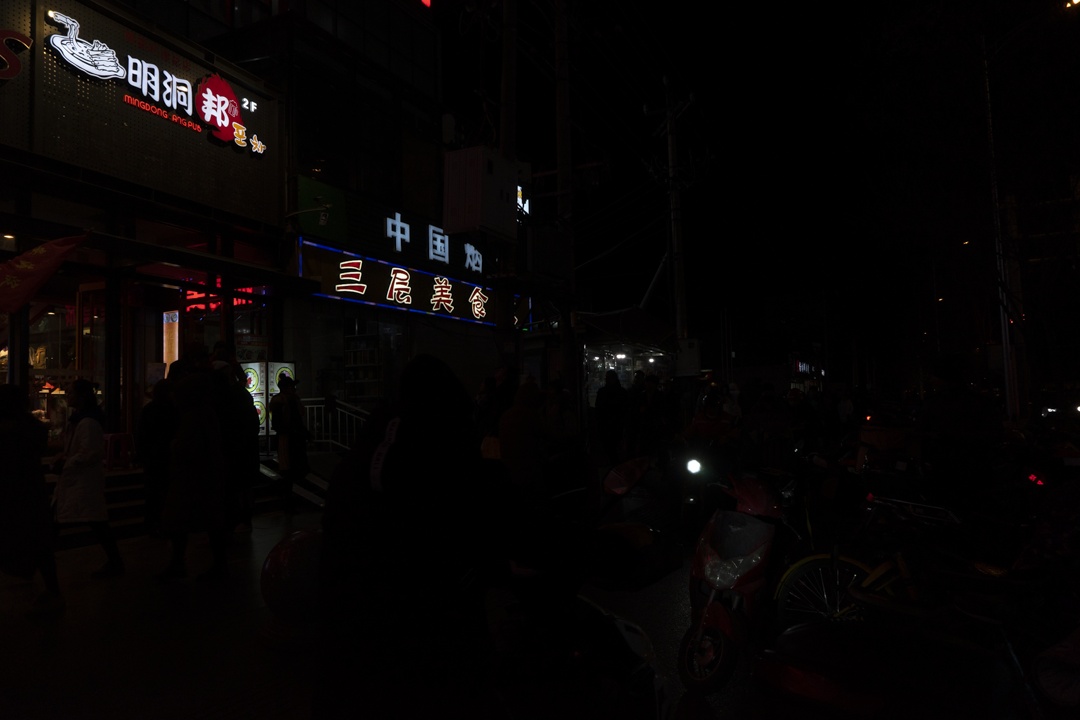}
\hspace*{\shrinkSpaceBetweenImages} &
\includegraphics[height=\heightFigDarkFaceAdaptive]{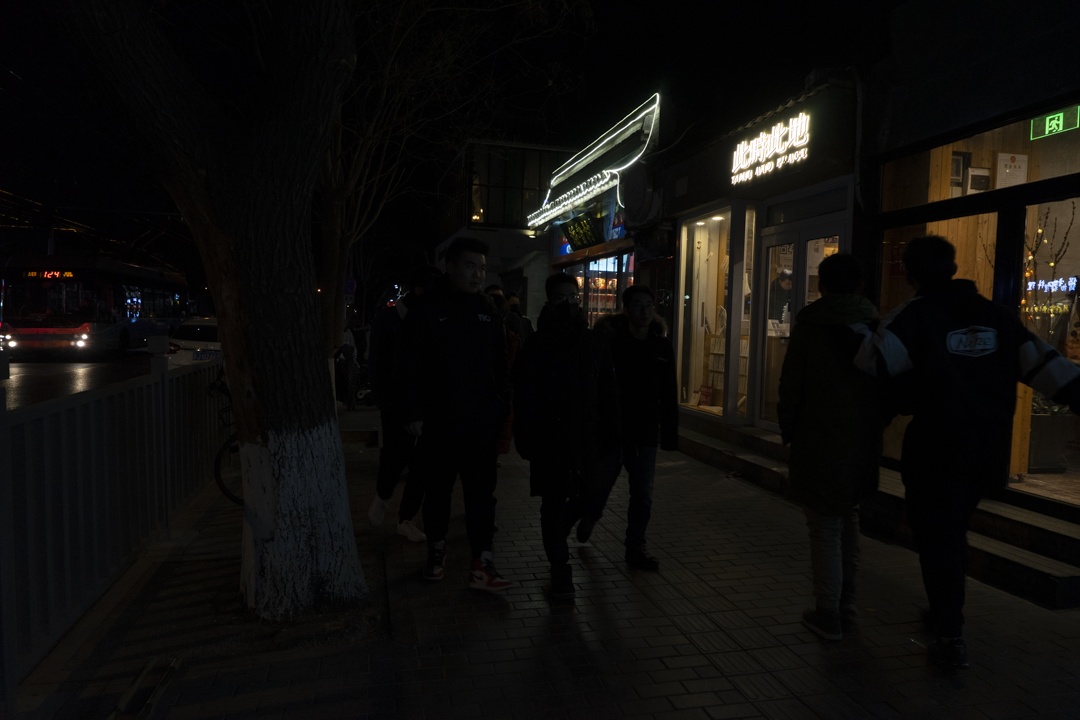}
\hspace*{\shrinkSpaceBetweenImages} &
\includegraphics[height=\heightFigDarkFaceAdaptive]{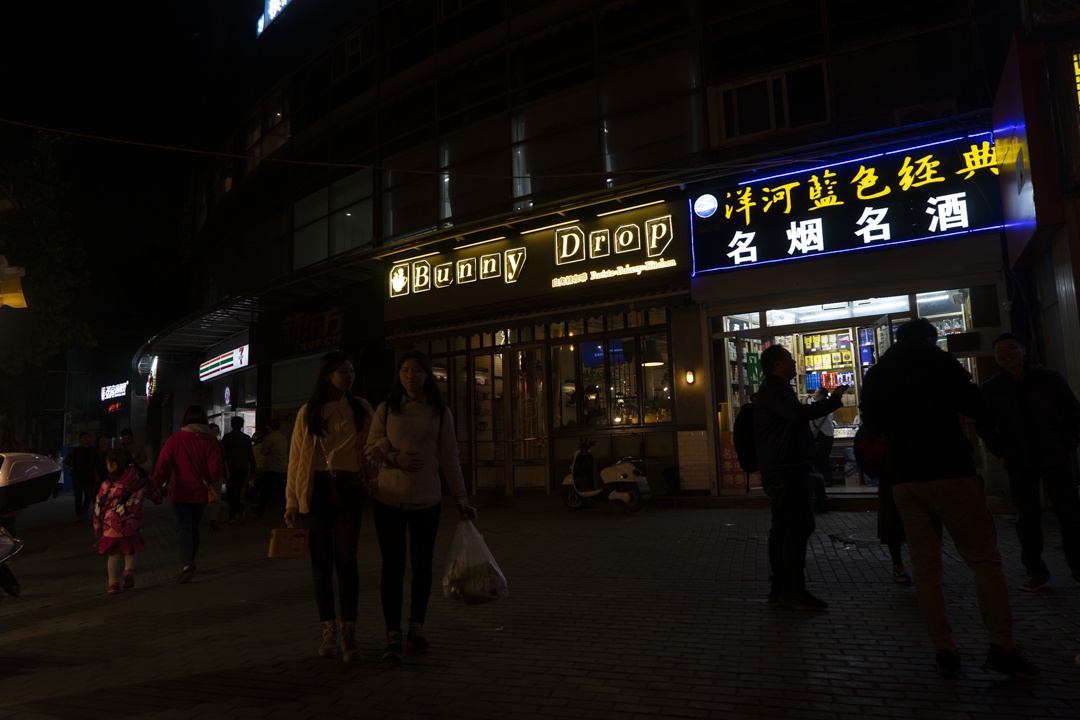}
\hspace*{\shrinkSpaceBetweenImages} &
\includegraphics[height=\heightFigDarkFaceAdaptive]{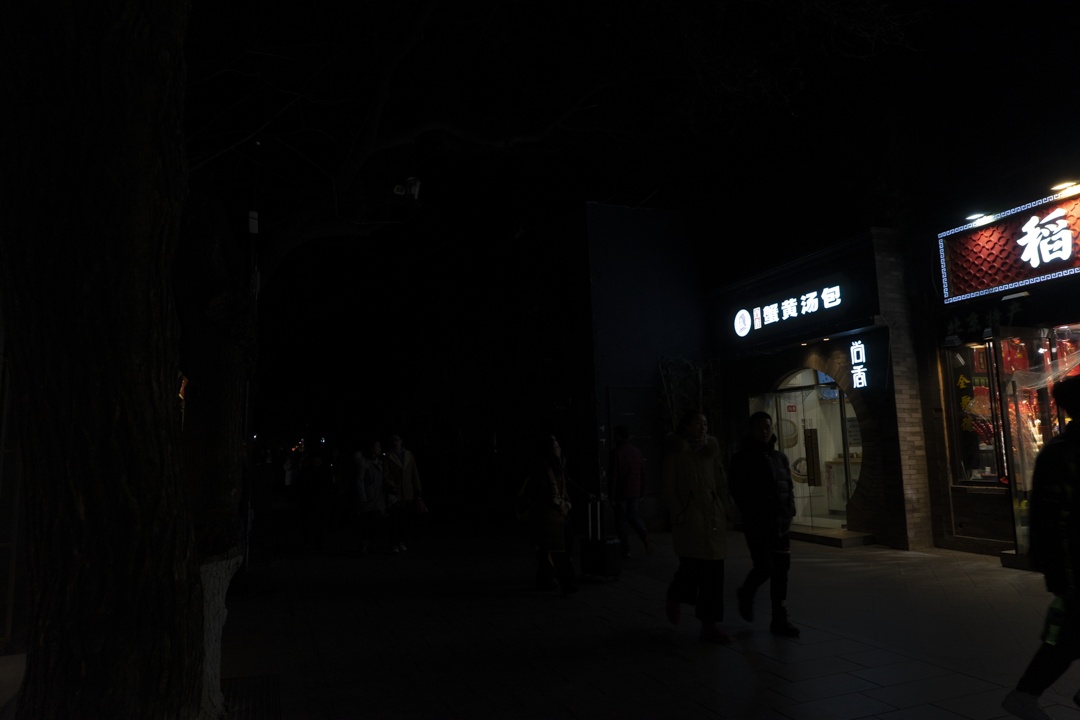}
\\
\rotatebox{90}{\scriptsize \texttt{imBeam}}
\hspace*{-0.5cm} &
\includegraphics[height=\heightFigDarkFaceAdaptive]{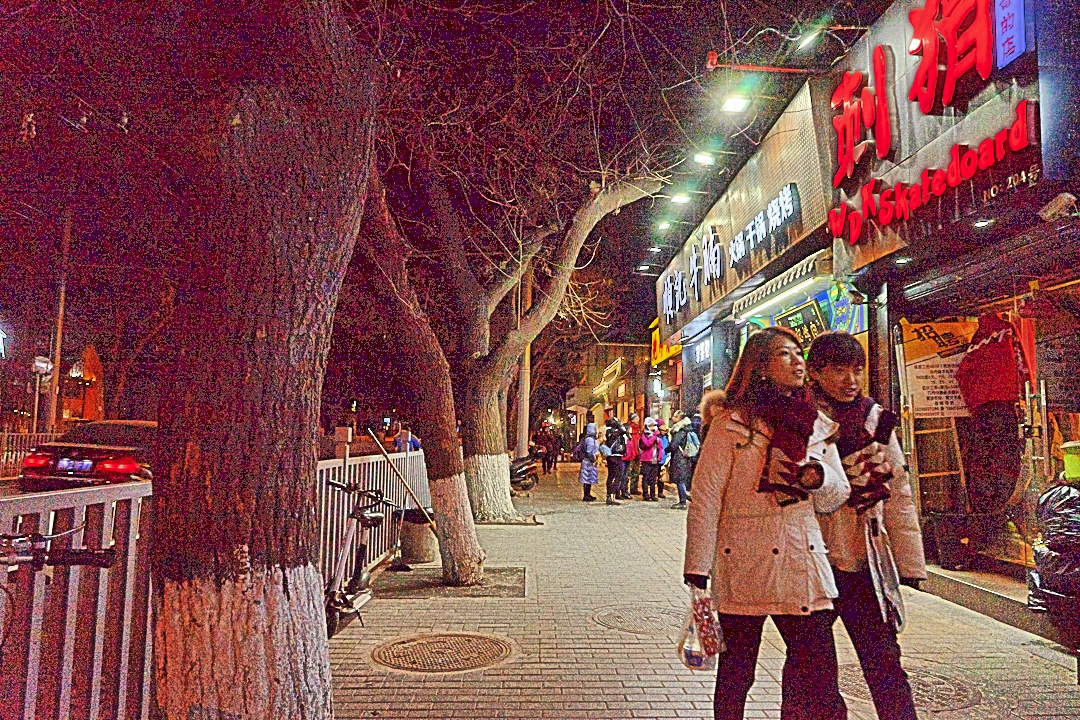}
\hspace*{\shrinkSpaceBetweenImages} &
\includegraphics[height=\heightFigDarkFaceAdaptive]{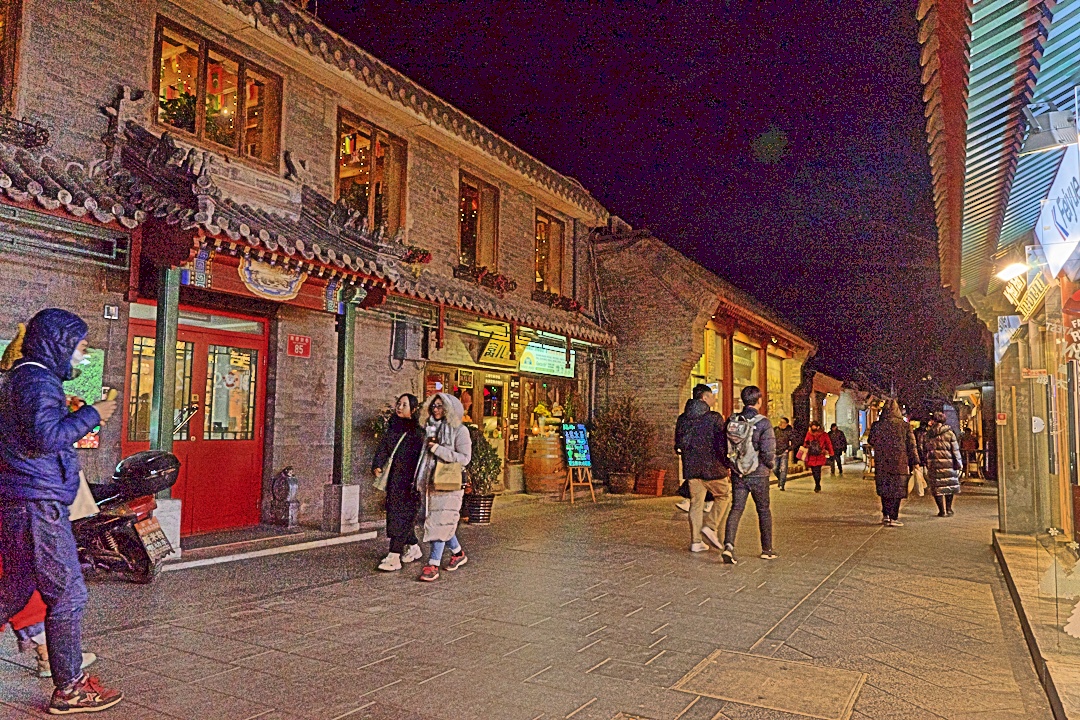}
\hspace*{\shrinkSpaceBetweenImages} &
\includegraphics[height=\heightFigDarkFaceAdaptive]{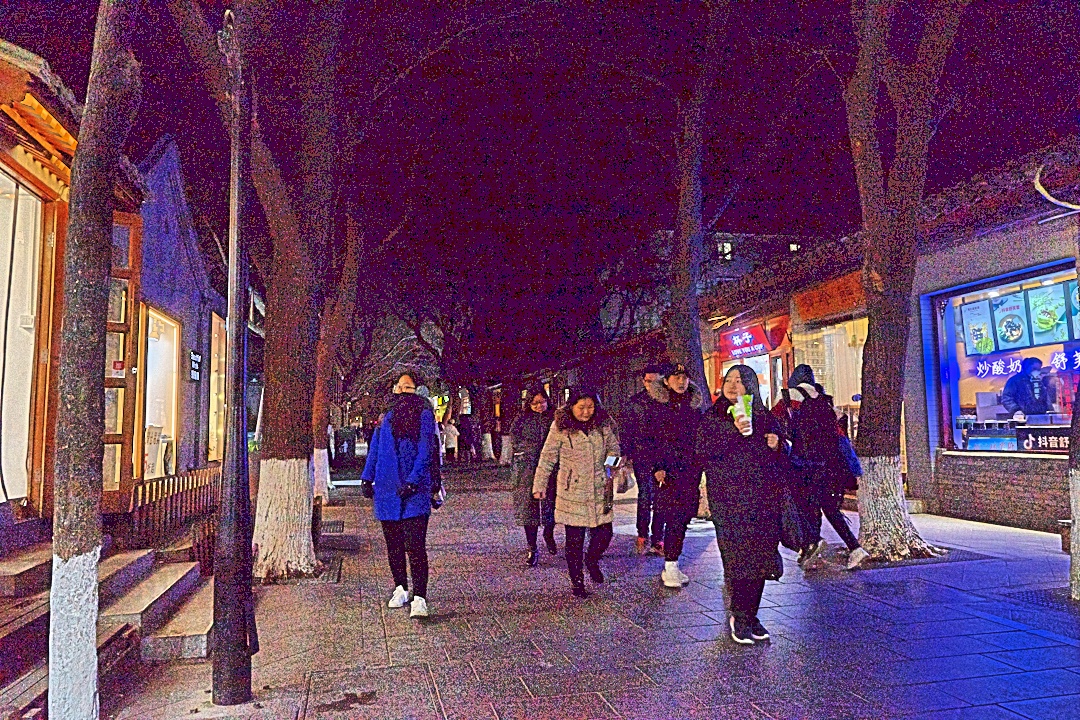}
\hspace*{\shrinkSpaceBetweenImages} &
\includegraphics[height=\heightFigDarkFaceAdaptive]{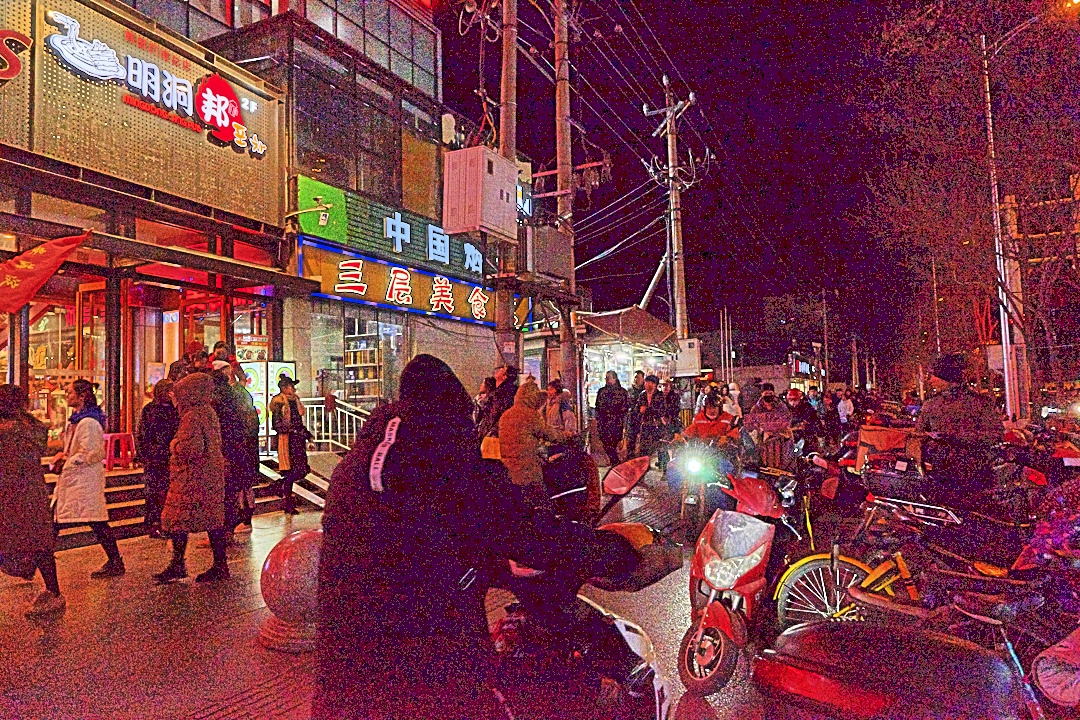}
\hspace*{\shrinkSpaceBetweenImages} &
\includegraphics[height=\heightFigDarkFaceAdaptive]{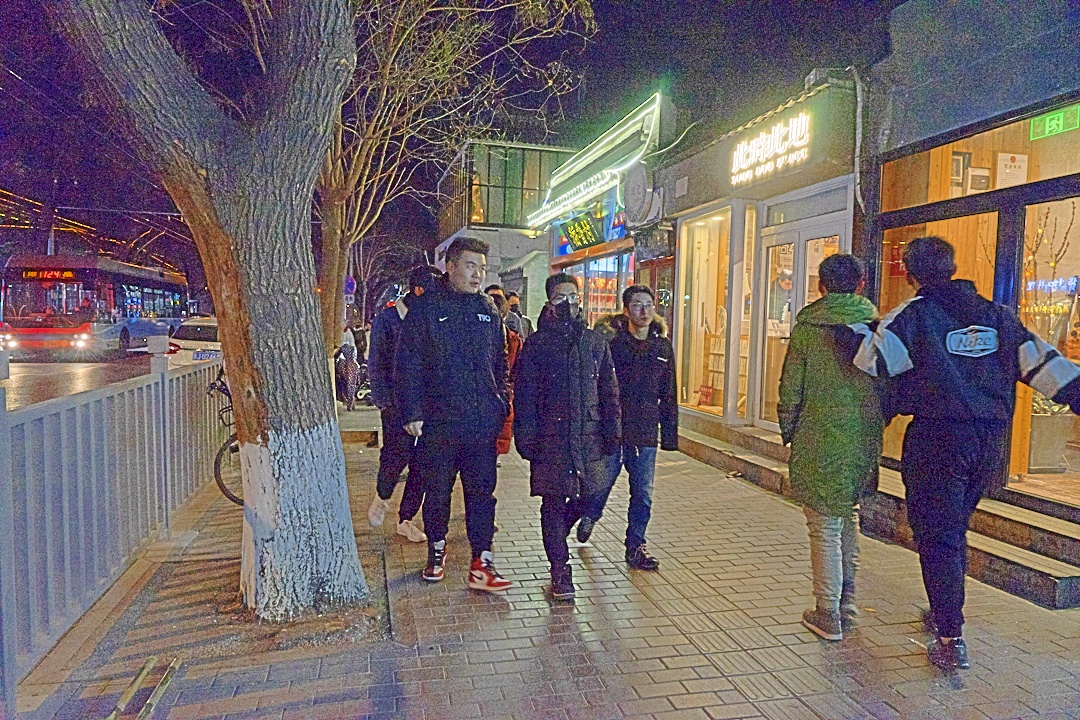}
\hspace*{\shrinkSpaceBetweenImages} &
\includegraphics[height=\heightFigDarkFaceAdaptive]{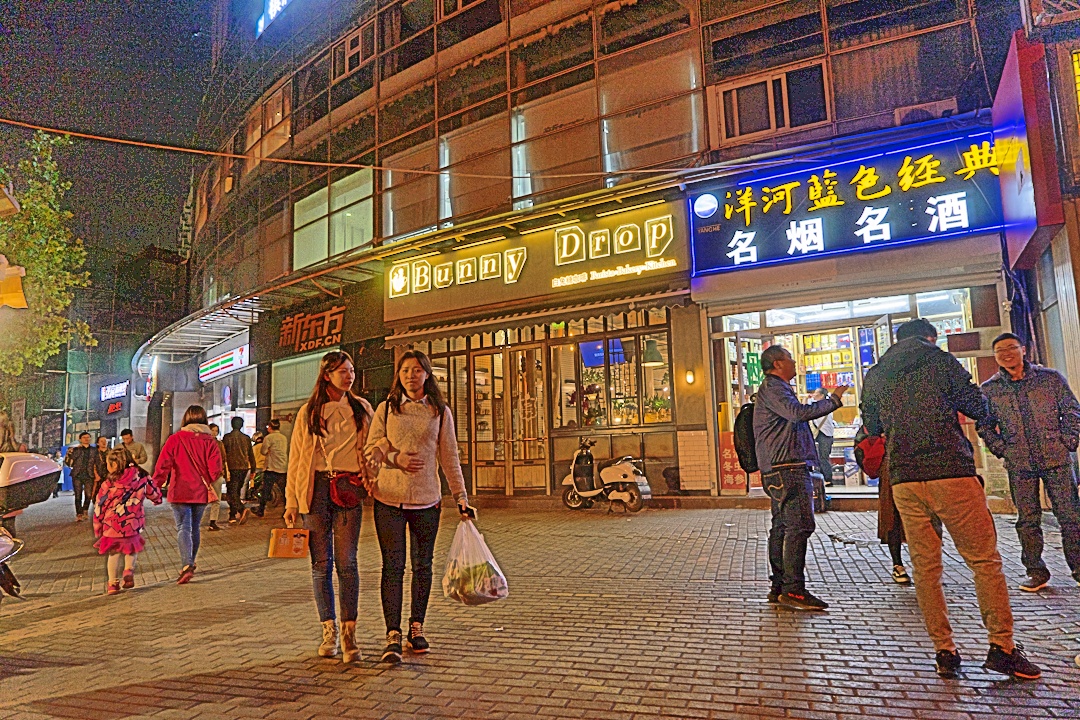}
\hspace*{\shrinkSpaceBetweenImages} &
\includegraphics[height=\heightFigDarkFaceAdaptive]{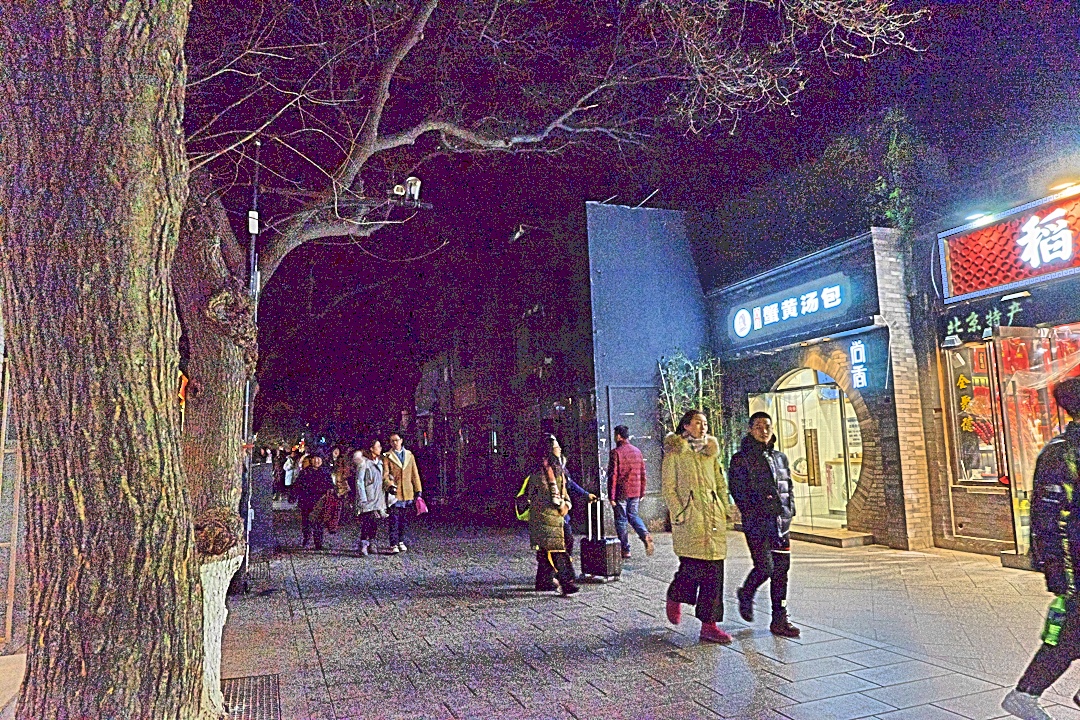}
\end{tabular}\vspace*{-0.3cm}
\caption{Images from the Dark Face collection used in various LLIE papers: Dark Face 8 \cite[fig.\,6]{CoLIE_eccv24}, Dark Face 9, \cite[fig.\,7]{LightenDiffusion_eccv24}, Dark Face 11, \cite[fig.\,8]{Zero-DCE_cvpr20}, Dark Face 101 \cite[fig.\,11]{Li-etal_pami22}, Dark Face 1353 \cite{Niu_TCI2025}, Dark Face 1462 \cite[fig.\,8]{SCI_cvpr22}, and Dark Face 5720 \cite[fig.\,6]{LMT_eccv24}) are shown in the upper row and filtered with the proposed \texttt{imBeam} in the lower row. The reader is invited to compare our results with those presented in the above-mentioned papers.}
\label{fig:usedDarkFace}
\vspace*{-0.2cm}
\end{figure*}

Fig.\,\ref{fig:LLcomparison} presents a visual comparison of the proposed approach with several state-of-the-art methods. For our experiments, we used the Dark Face dataset \cite{DarkFace_tip20_short}, a large collection of challenging images taken in diverse low-light scenarios, including many captured in extremely low-light conditions. The evaluation includes recent deep learning-based supervised methods, such as RetinexFormer \cite{Retinexformer_cvpr23} and LightenDiffusion \cite{LightenDiffusion_eccv24}, as well as zero-shot methods, including SCI \cite{SCI_cvpr22}, and the recent CoLIE \cite{CoLIE_eccv24}. The method STAR \cite{xu2020star} is also included for reference, as the most successful, non-deep learning, Retinex-based LLIE method. Although deep learning-based methods generally perform well, they struggle with images captured under extremely low-light conditions. In contrast, the proposed method produces superior results. Compared to the results obtained by STAR \cite{xu2020star}, which is possibly the most successful Retinex-based method, our approach also produces visually more satisfactory results and fewer artifacts.

To further compare the results of the proposed approach to existing methods, we provide in Fig.\,\ref{fig:usedDarkFace} results from applying \texttt{imBeam} to some images from the Dark Face dataset that appear in different publications. (The references are given in the caption of Fig.\,\ref{fig:usedDarkFace}.) The reader is invited to compare our results with those presented in the cited literature.

\subsection*{Image Clarification}
Results obtained by our image clarification approach (filtering by \imBeamAMF followed by \imBeamGF) are illustrated in Fig.\,\ref{fig:teaser}, third to sixth images from the left, to enhance hazy, sand-dust, and underwater images. We used images from the RUSH data set which was introduced in \cite{Liu-etal_pami23}.

\begin{figure*}[h!tbp]
\hspace*{-0.1cm}\centering
\begin{tabular}{ccccccc}
\includegraphics[width=\sFigHaze\linewidth]{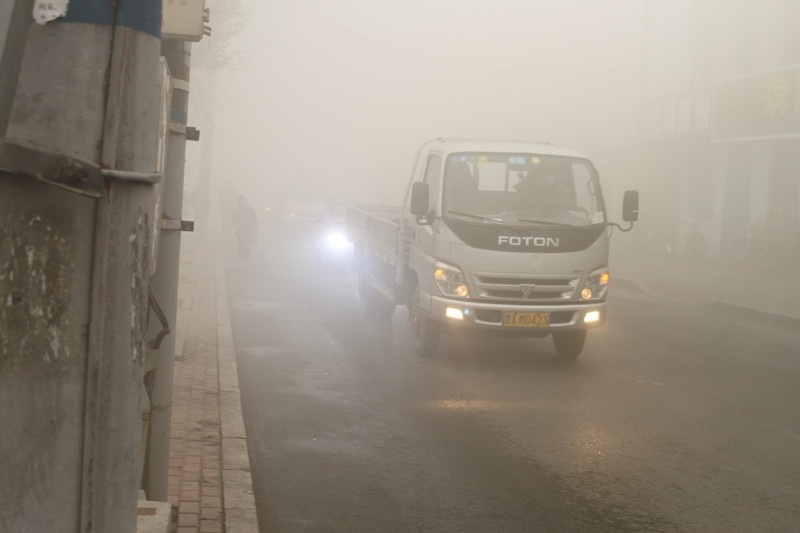}
\hspace*{\shrinkSpaceBetweenImages} &
\includegraphics[width=\sFigHaze\linewidth]{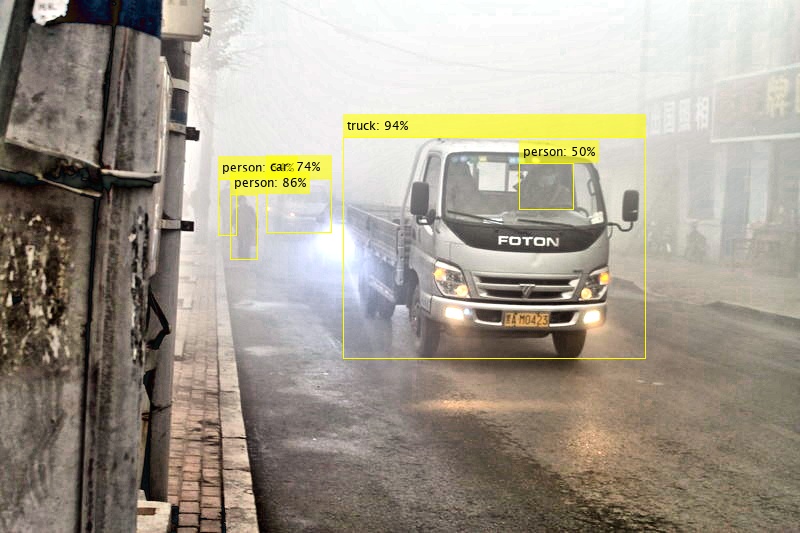}
\hspace*{\shrinkSpaceBetweenImages} &
\includegraphics[width=\sFigHaze\linewidth]{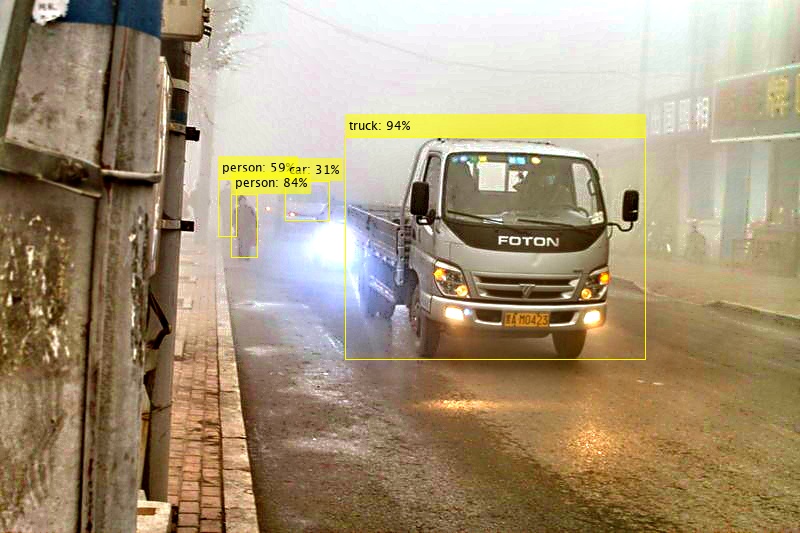}
\hspace*{\shrinkSpaceBetweenImages} &
\includegraphics[width=\sFigHaze\linewidth]{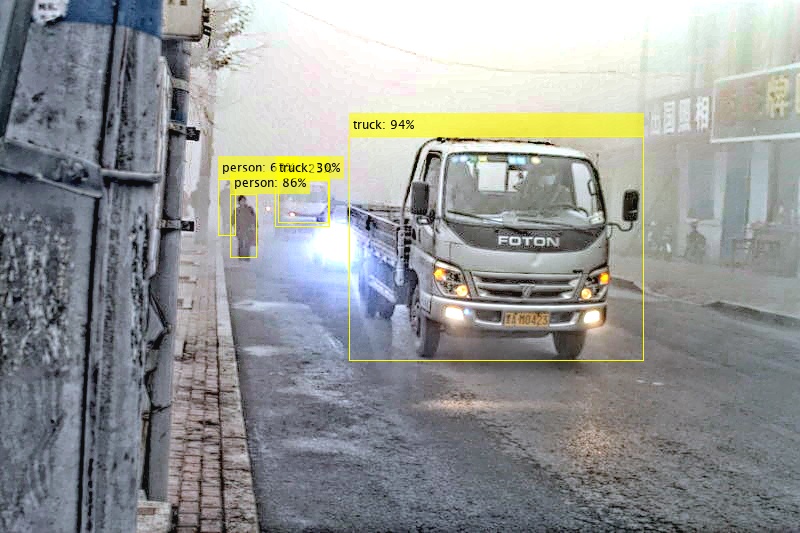}
\hspace*{\shrinkSpaceBetweenImages} &
\includegraphics[width=\sFigHaze\linewidth]{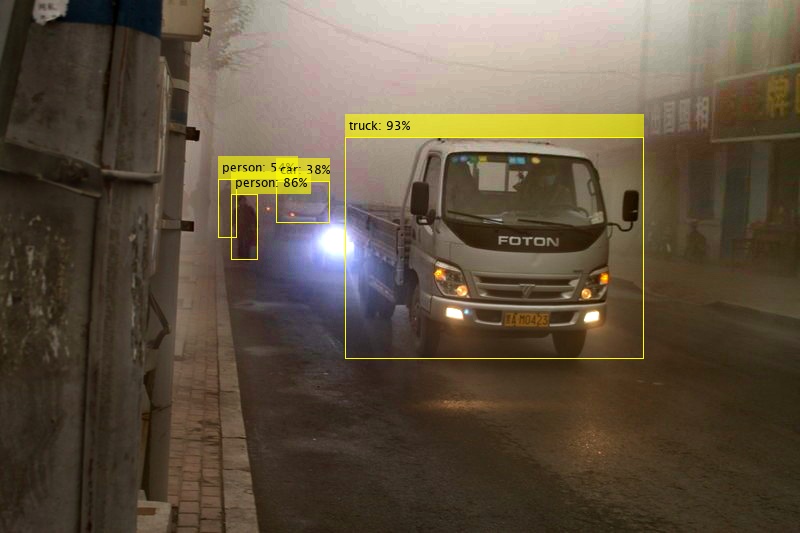}
\hspace*{\shrinkSpaceBetweenImages} &
\includegraphics[width=\sFigHaze\linewidth]{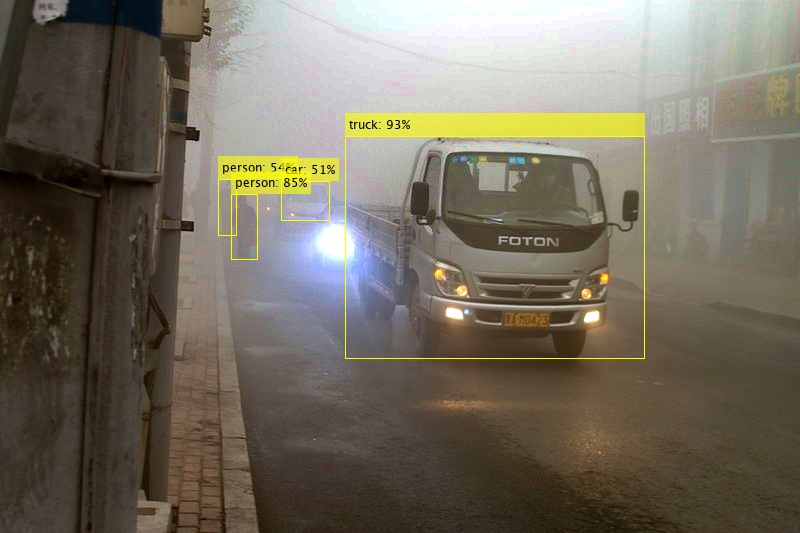}
\hspace*{\shrinkSpaceBetweenImages} &
\includegraphics[width=\sFigHaze\linewidth]{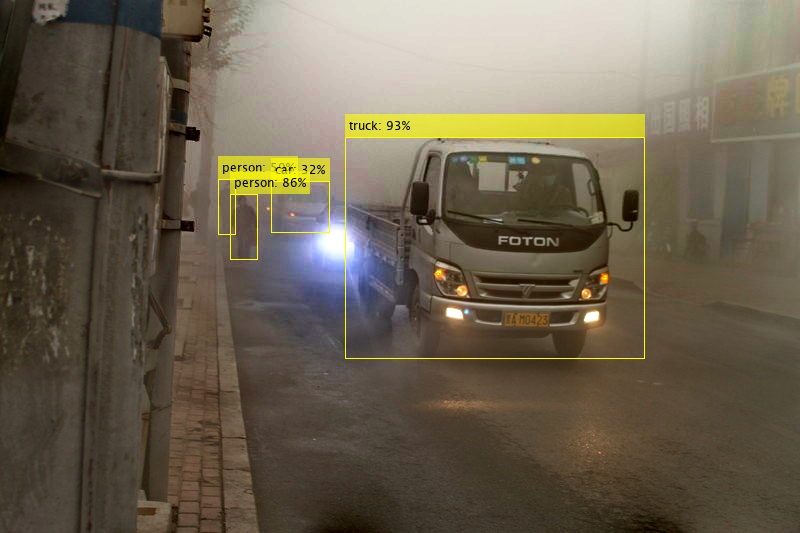
}
\\
\includegraphics[width=\sFigHaze\linewidth]{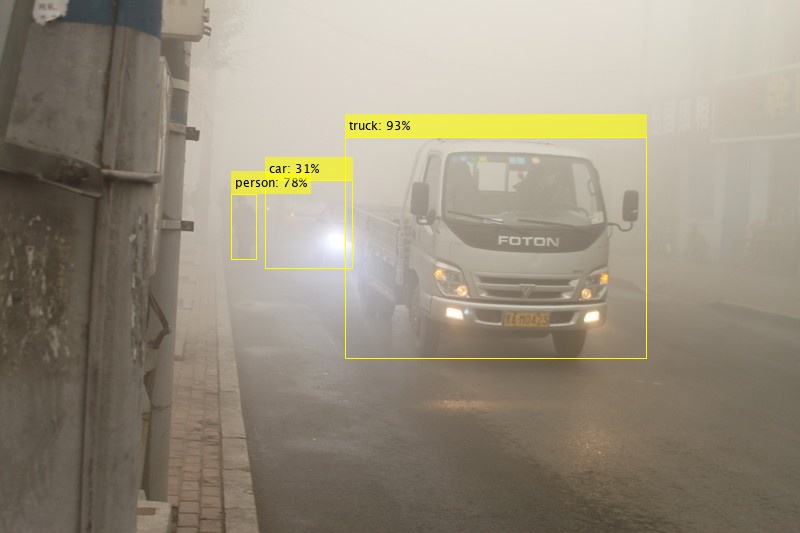}
\hspace*{\shrinkSpaceBetweenImages} &
\includegraphics[width=\sFigHaze\linewidth]{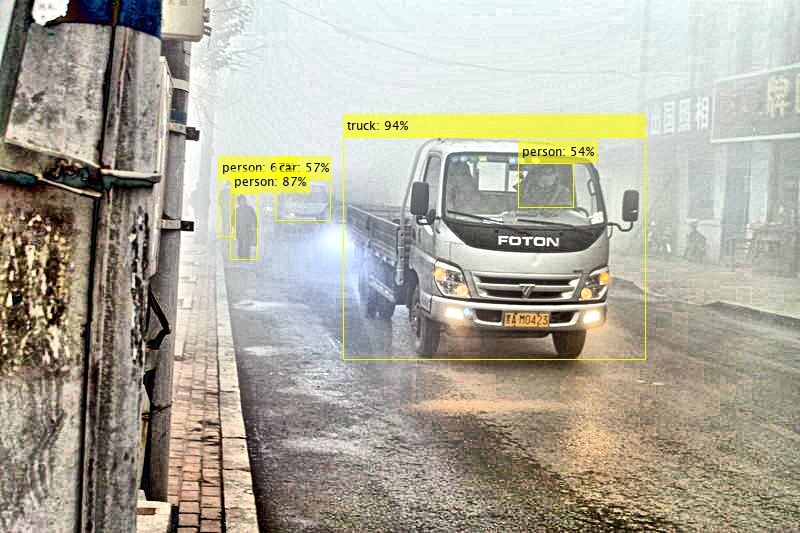}
\hspace*{\shrinkSpaceBetweenImages} &
\includegraphics[width=\sFigHaze\linewidth]{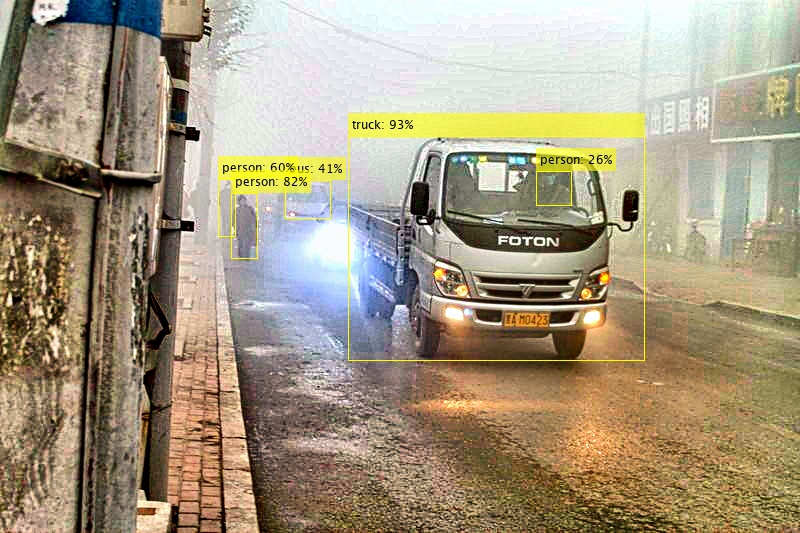}
\hspace*{\shrinkSpaceBetweenImages} &
\includegraphics[width=\sFigHaze\linewidth]{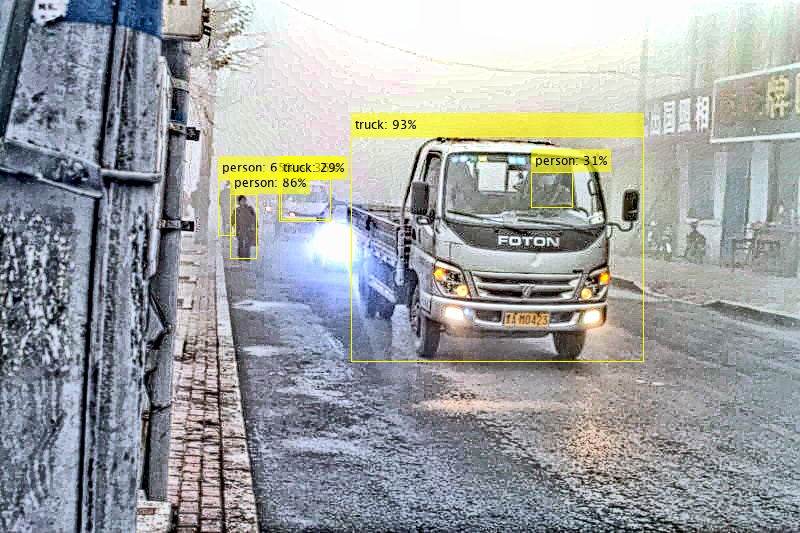}
\hspace*{\shrinkSpaceBetweenImages} &
\includegraphics[width=\sFigHaze\linewidth]{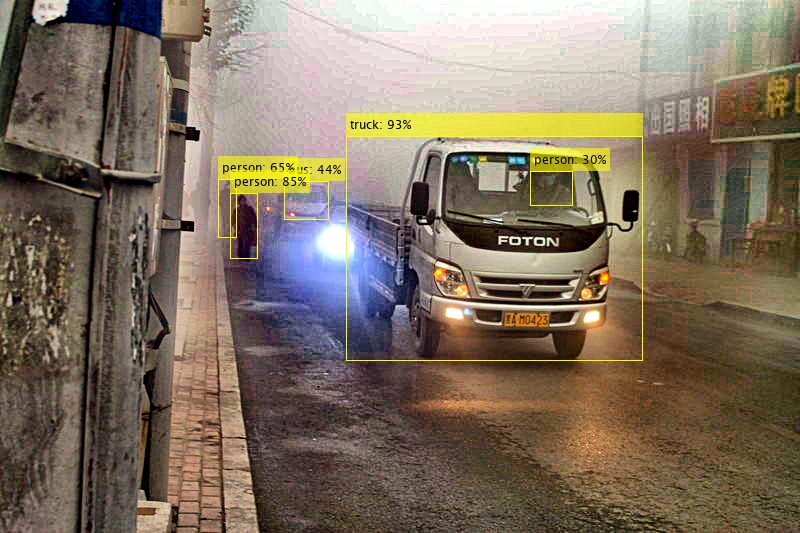}
\hspace*{\shrinkSpaceBetweenImages} &
\includegraphics[width=\sFigHaze\linewidth]{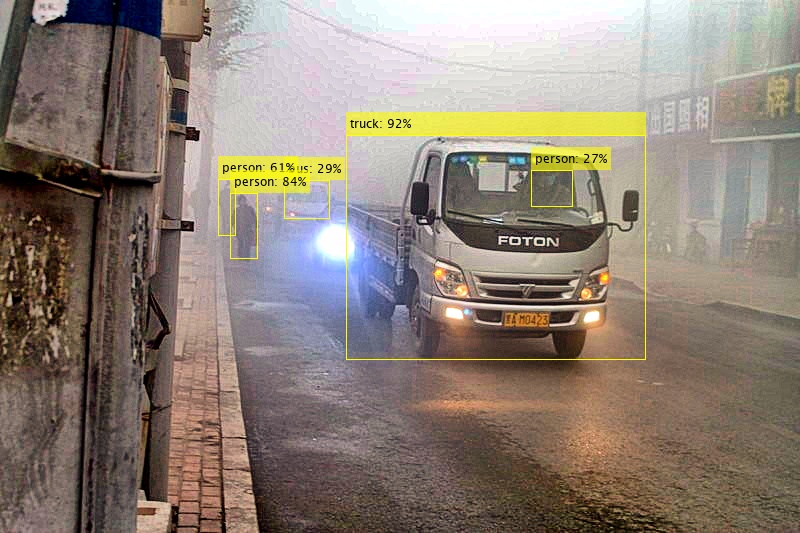}
\hspace*{\shrinkSpaceBetweenImages} &
\includegraphics[width=\sFigHaze\linewidth]{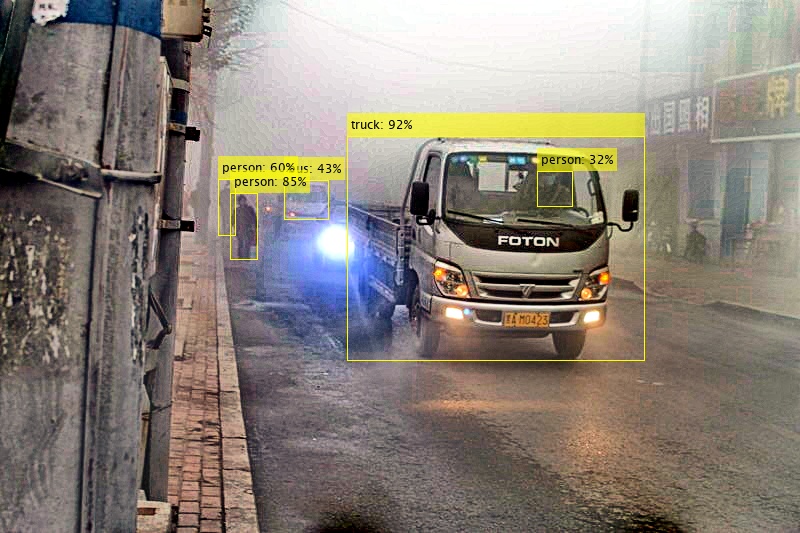}
\\
\footnotesize{hazy+YOLOX} & \footnotesize{\imBeamAMF \& +\imBeamGF} & \footnotesize{RGCP \& +\imBeamGF} & \footnotesize{$\mathrm{ROP}^+$ \& +\imBeamGF} & \footnotesize{SLP \& +\imBeamGF} & \footnotesize{RLP \& +\imBeamGF} & \footnotesize{RRO \& +\imBeamGF}
\end{tabular}\vspace*{-0.3cm}
\caption{Visual comparison of several SOTA prior-based image clarification schemes and their improved version, followed by object detection. `+\imBeamGF' indicates that the image was enhanced by \imBeamGF. Object detection was performed using YOLOX \cite{Ge2021yolox}.} \label{fig:yoloxHazy076}
\vspace*{-0.2cm}
\end{figure*}

\begin{figure*}[h!tbp]
\hspace*{-0.1cm}\centering
\begin{tabular}{ccccccc}
\includegraphics[width=\sFigHaze\linewidth]{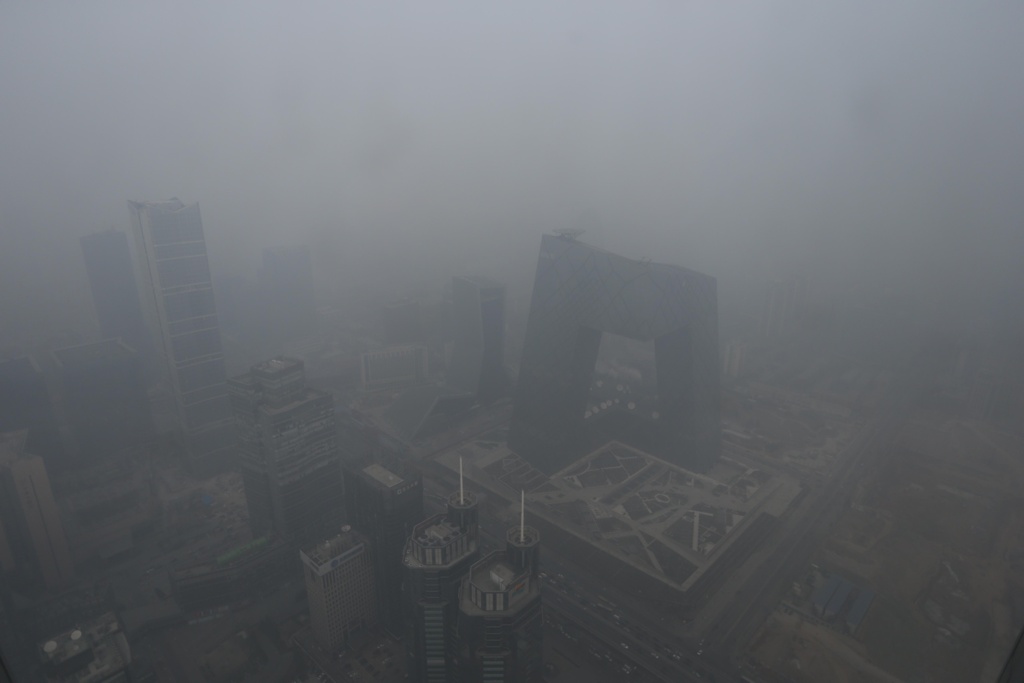}
\hspace*{\shrinkSpaceBetweenImages} &
\includegraphics[width=\sFigHaze\linewidth]{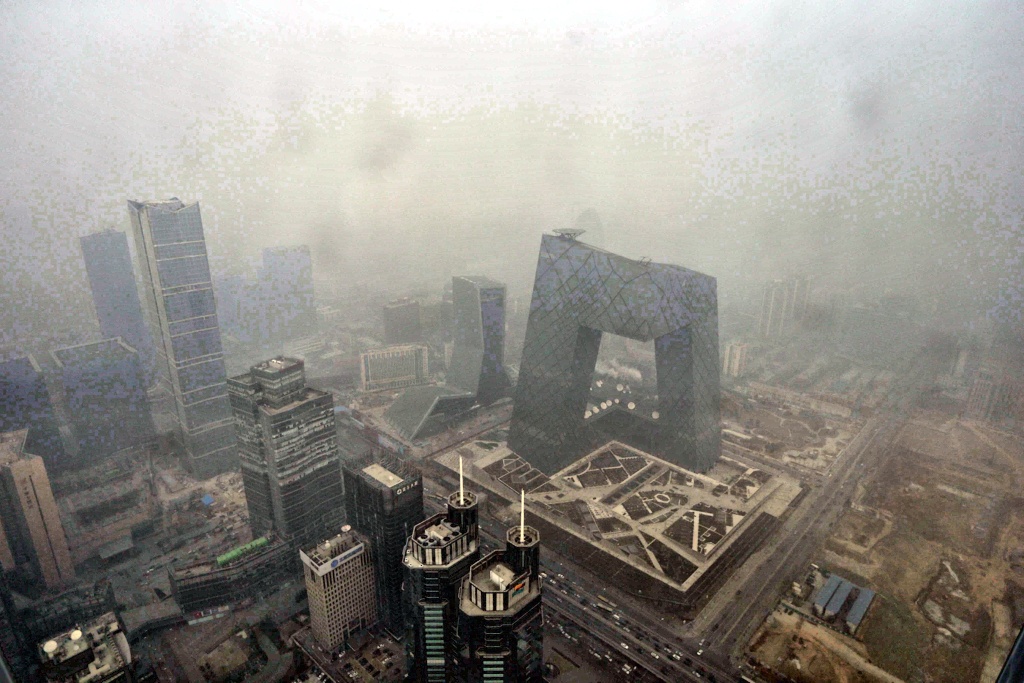}
\hspace*{\shrinkSpaceBetweenImages} &
\includegraphics[width=\sFigHaze\linewidth]{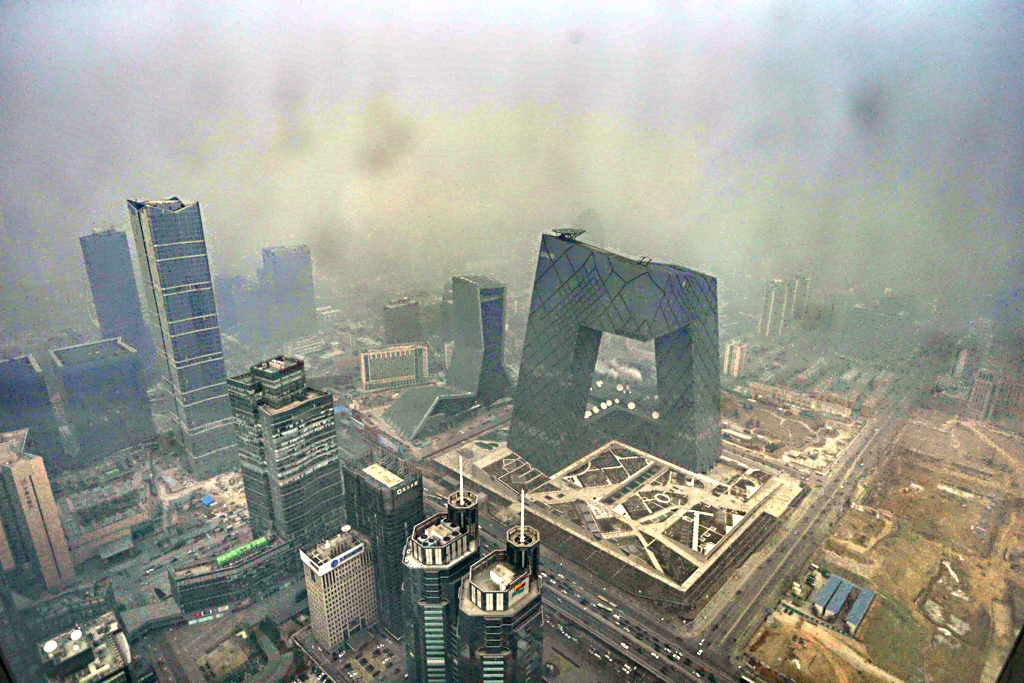}
\hspace*{\shrinkSpaceBetweenImages} &
\includegraphics[width=\sFigHaze\linewidth]{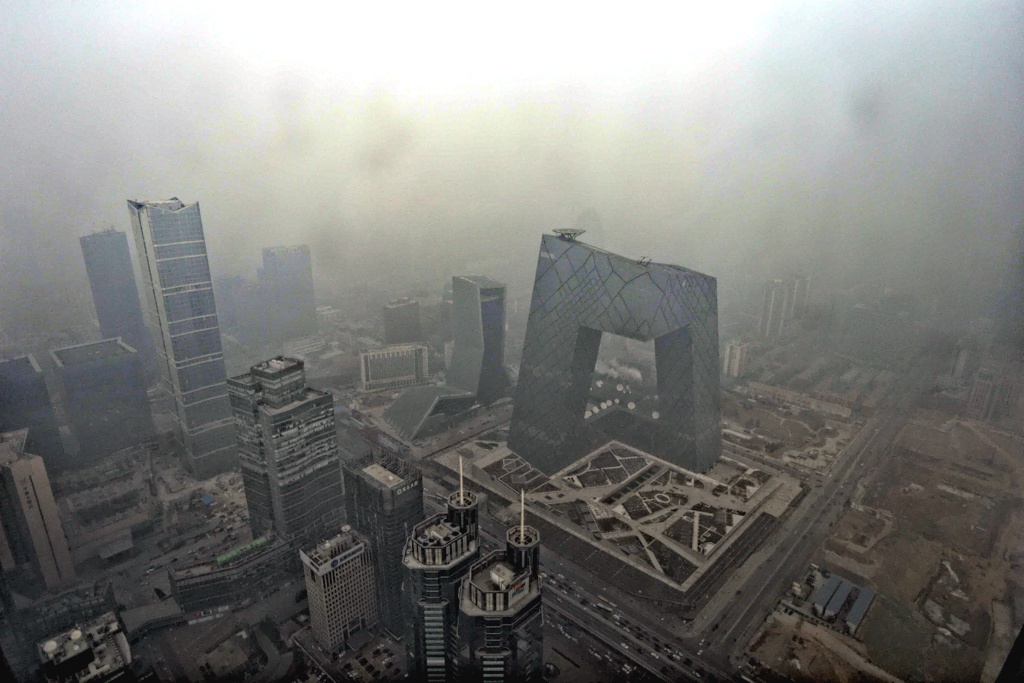}
\hspace*{\shrinkSpaceBetweenImages} &
\includegraphics[width=\sFigHaze\linewidth]{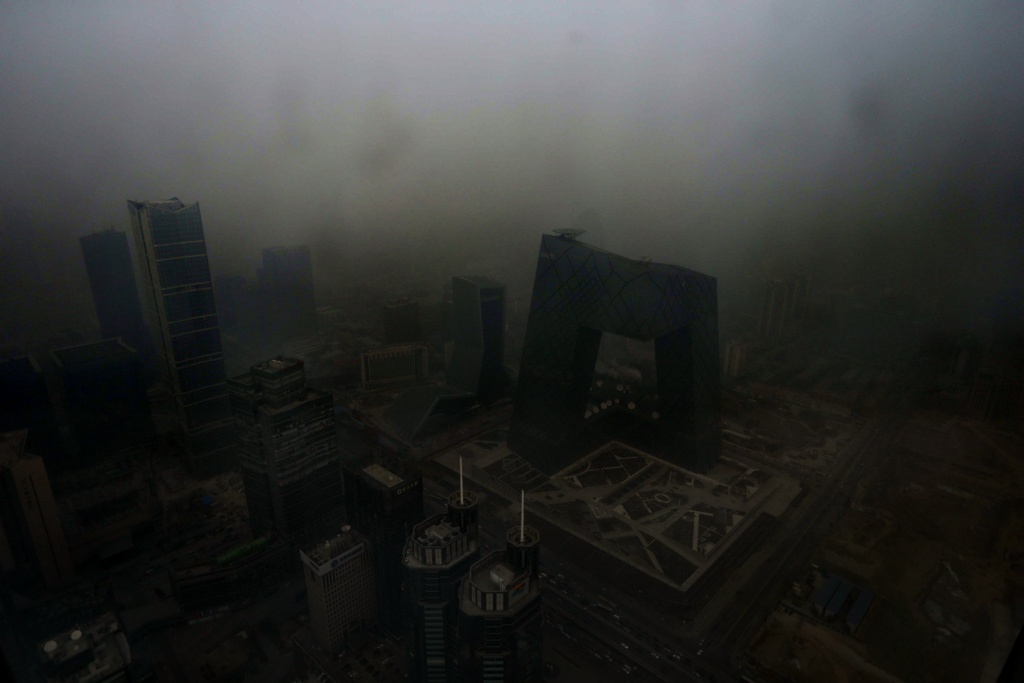}
\hspace*{\shrinkSpaceBetweenImages} &
\includegraphics[width=\sFigHaze\linewidth]{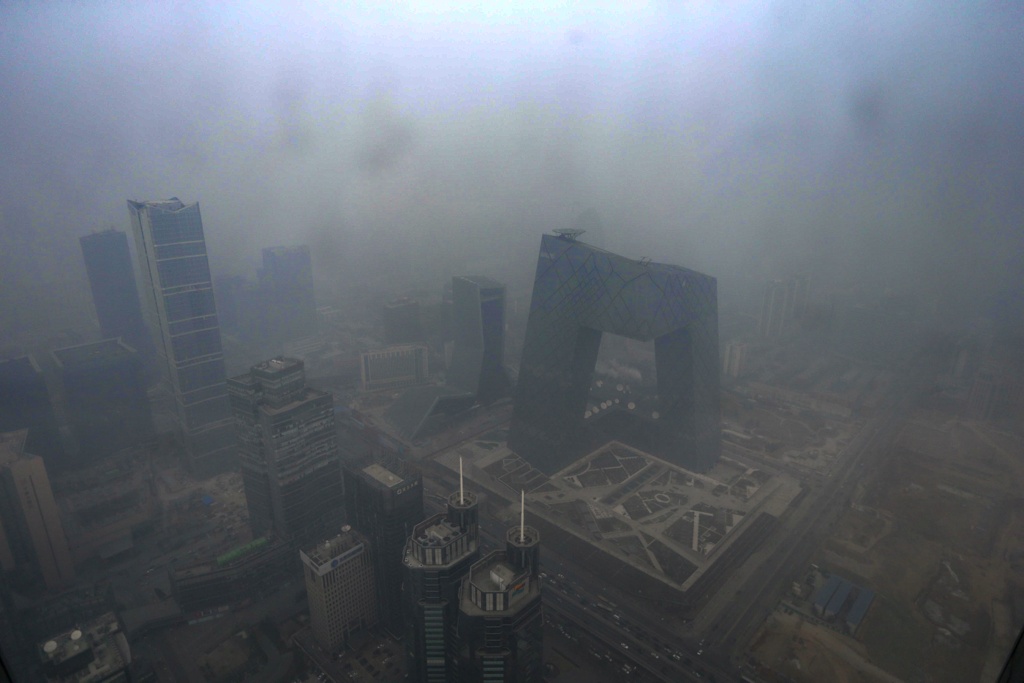}
\hspace*{\shrinkSpaceBetweenImages} &
\includegraphics[width=\sFigHaze\linewidth]{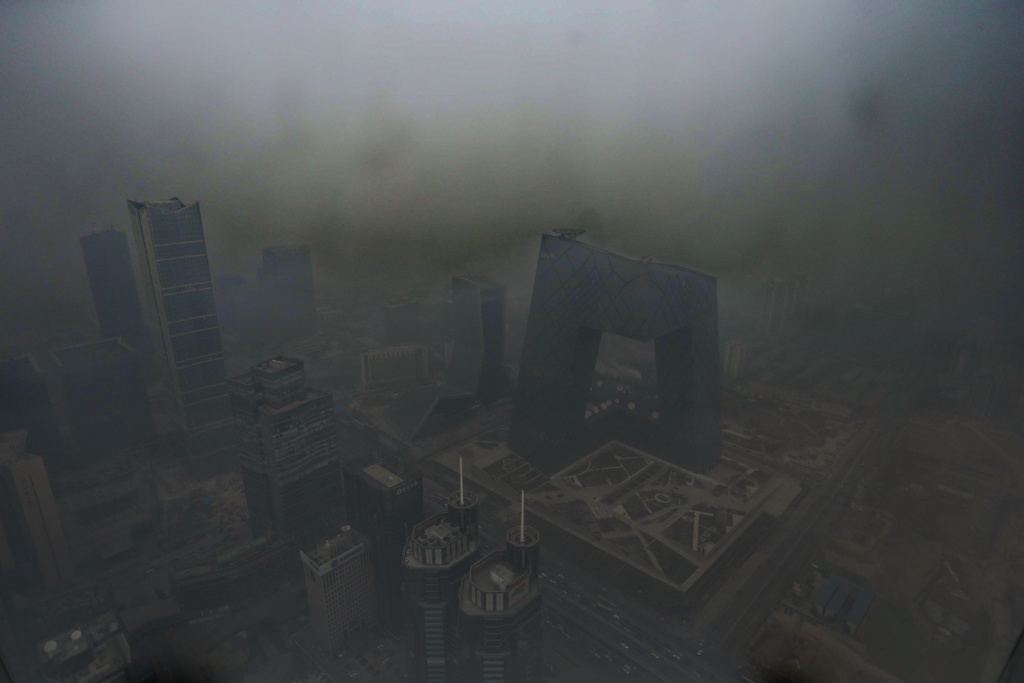}
\\
\includegraphics[width=\sFigHaze\linewidth]{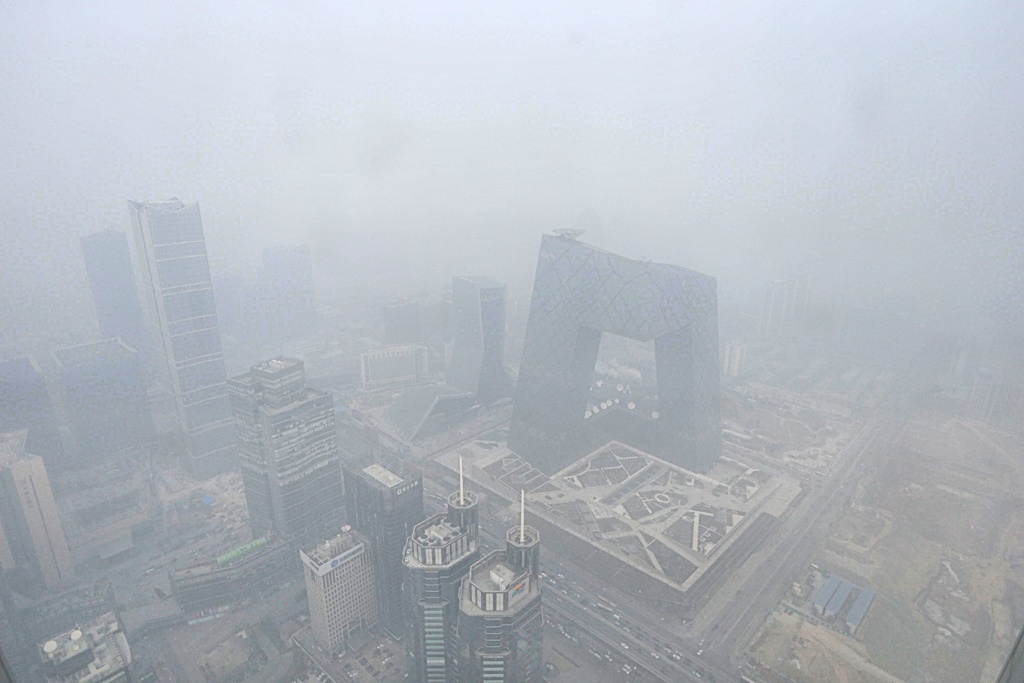}
\hspace*{\shrinkSpaceBetweenImages} &
\includegraphics[width=\sFigHaze\linewidth]{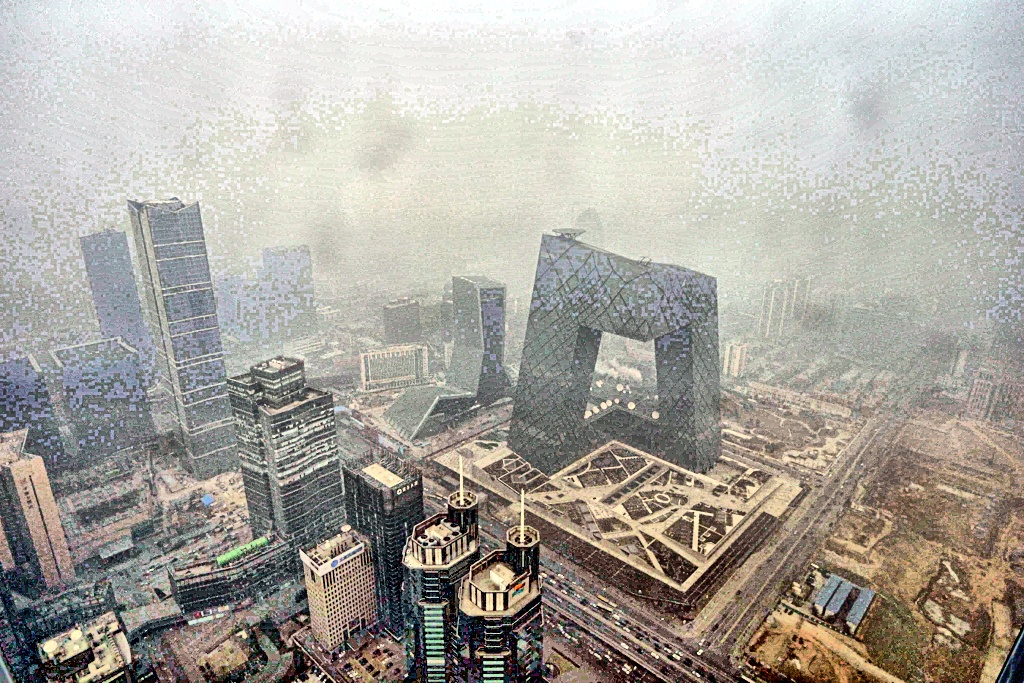}
\hspace*{\shrinkSpaceBetweenImages} &
\includegraphics[width=\sFigHaze\linewidth]{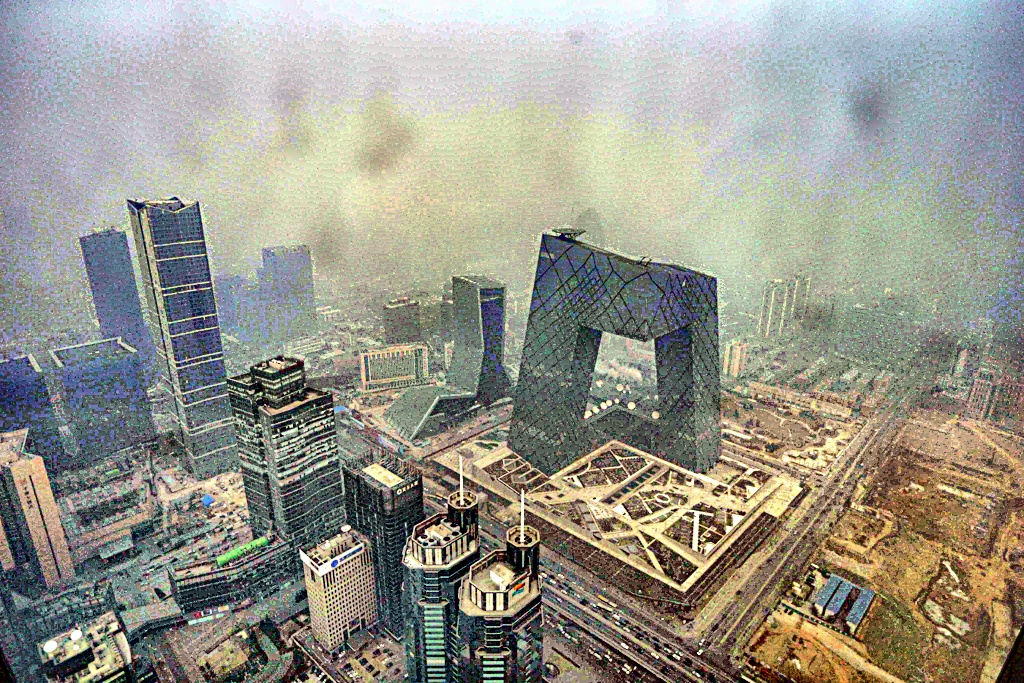}
\hspace*{\shrinkSpaceBetweenImages} &
\includegraphics[width=\sFigHaze\linewidth]{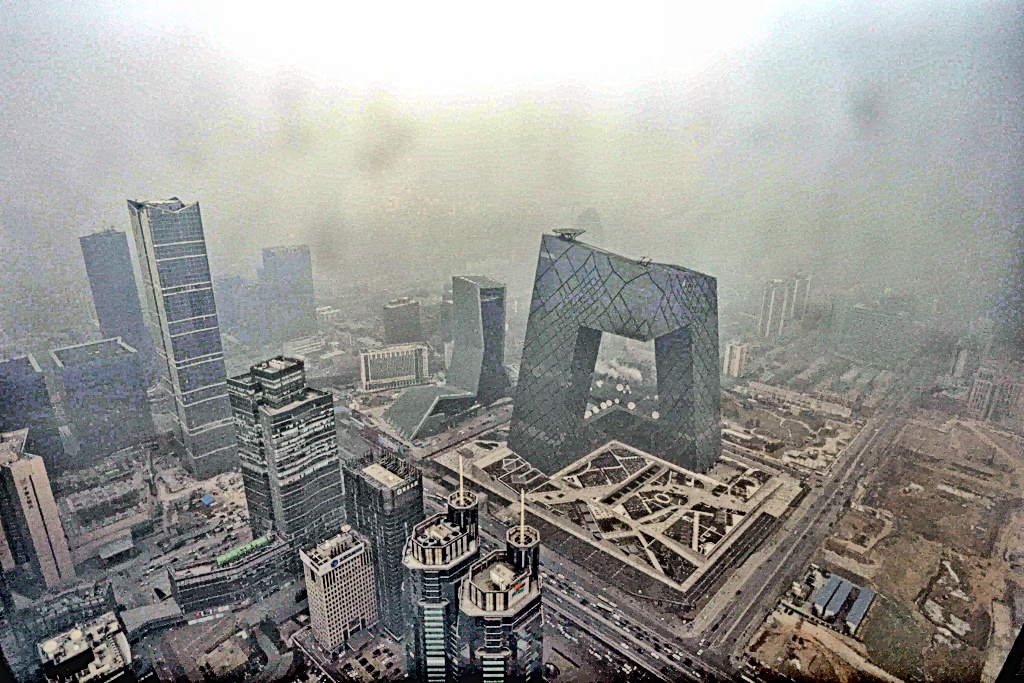}
\hspace*{\shrinkSpaceBetweenImages} &
\includegraphics[width=\sFigHaze\linewidth]{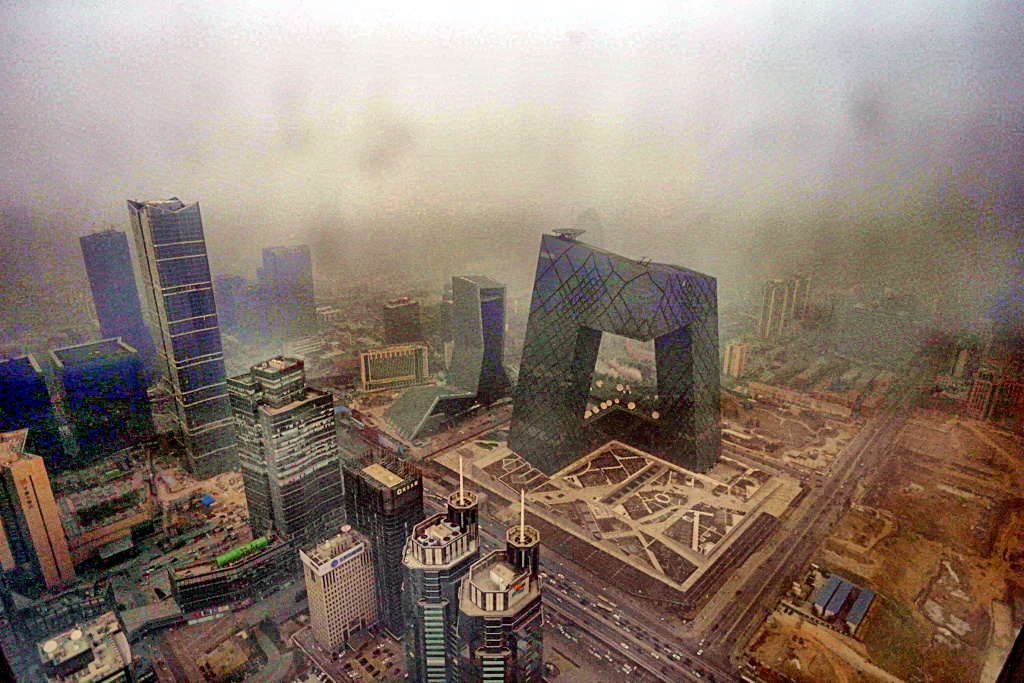}
\hspace*{\shrinkSpaceBetweenImages} &
\includegraphics[width=\sFigHaze\linewidth]{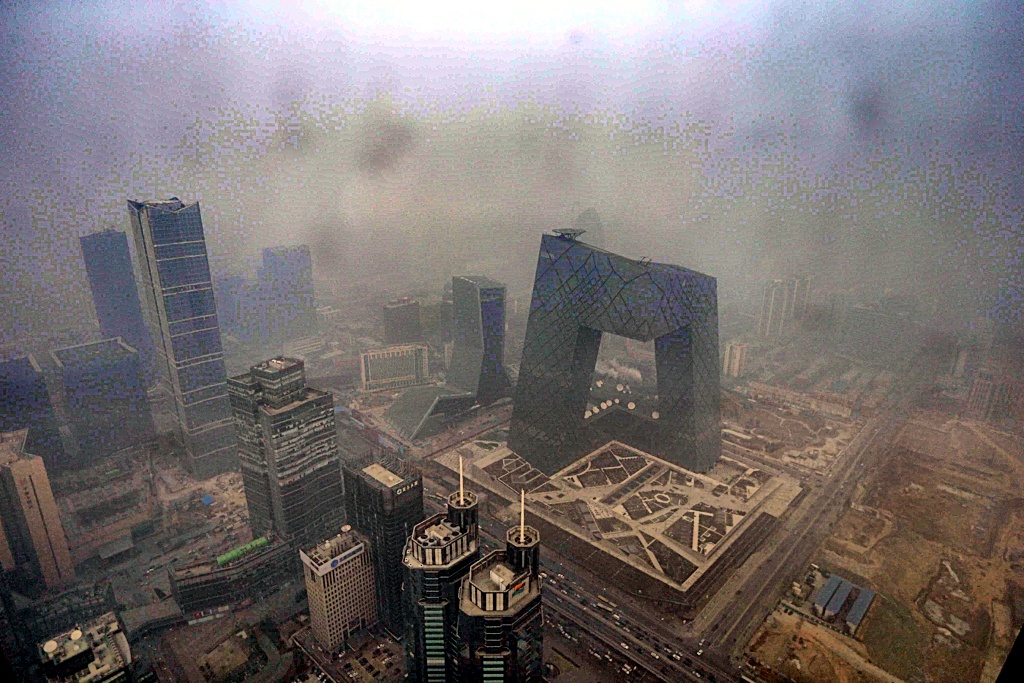}
\hspace*{\shrinkSpaceBetweenImages} &
\includegraphics[width=\sFigHaze\linewidth]{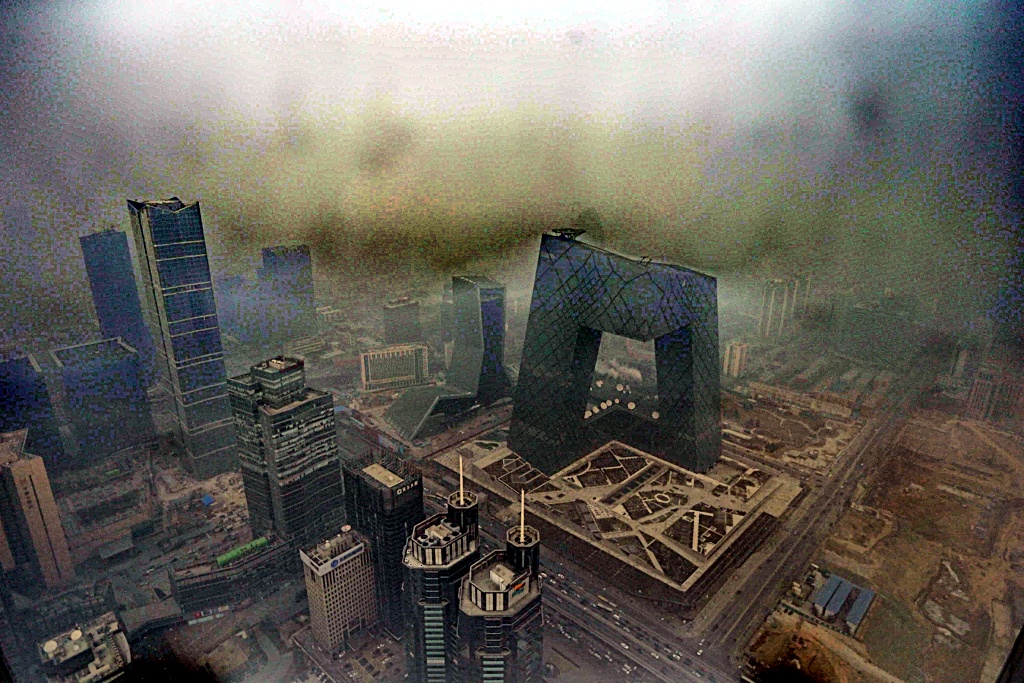}
\\
\footnotesize{hazy \& +\imBeamGF} & \footnotesize{\imBeamAMF \& +\imBeamGF}  & \footnotesize{RGCP \& +\imBeamGF} & \footnotesize{$\mathrm{ROP}^+$ \& +\imBeamGF} & \footnotesize{SLP \& +\imBeamGF} & \footnotesize{RLP \& +\imBeamGF} & \footnotesize{RRO \& +\imBeamGF}
\end{tabular}\vspace*{-0.3cm}
\caption{Visual comparison of SOTA prior-based image clarification schemes and the proposed improvement for enhancing a challenging image with heavy fog. Notice how combining \imBeamGF with other methods further improves dehazing results. This is particularly apparent for the results obtained by SLP.}\label{fig:Hazy079}
\vspace*{-0.2cm}
\end{figure*}

\begin{figure*}[h!tbp]
\hspace*{-0.2cm}\centering
\begin{tabular}{ccccccc}
\includegraphics[width=\sFigHaze\linewidth]{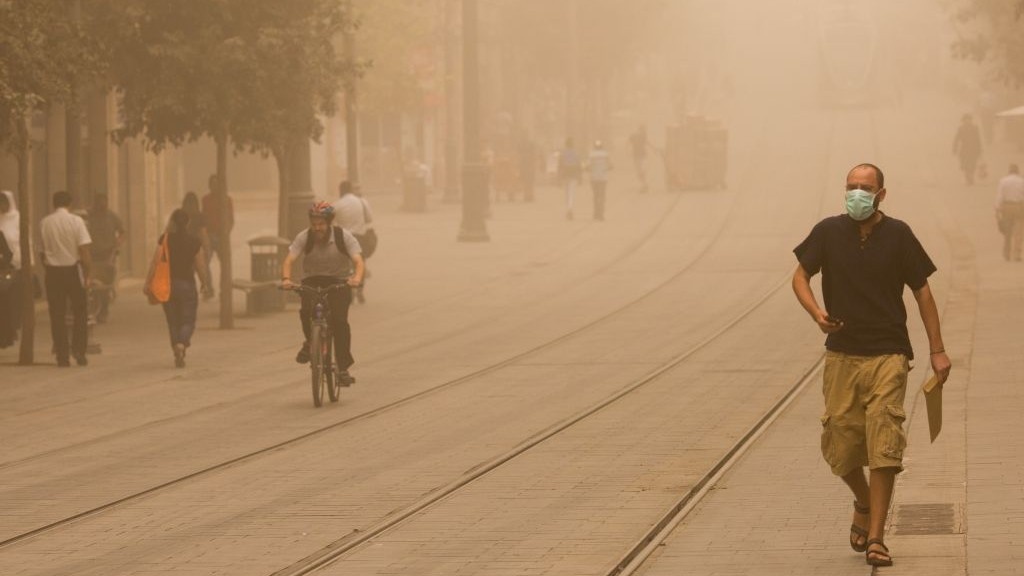}
\hspace*{\shrinkSpaceBetweenImages} &
\includegraphics[width=\sFigHaze\linewidth]{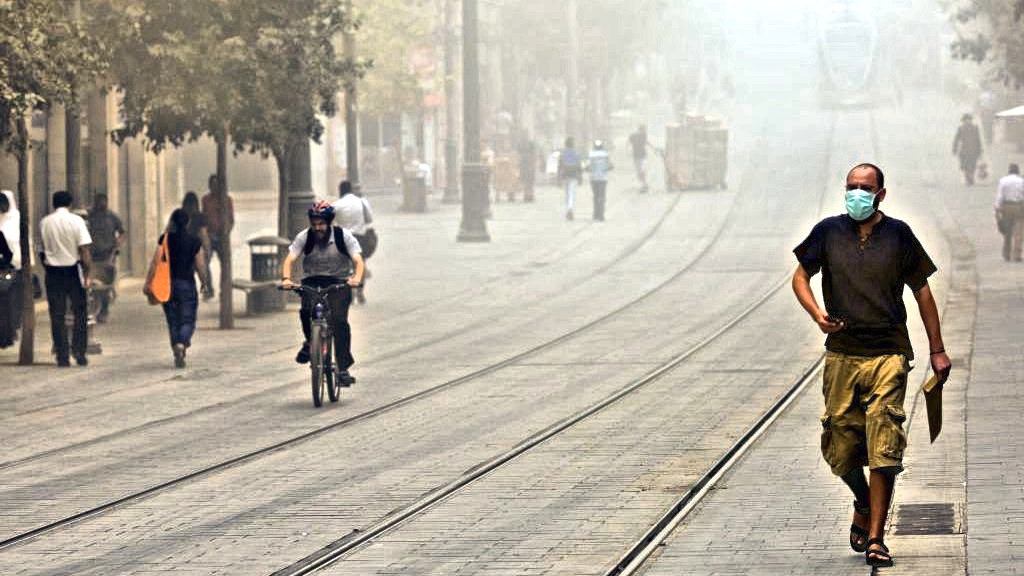}
\hspace*{\shrinkSpaceBetweenImages} &
\includegraphics[width=\sFigHaze\linewidth]{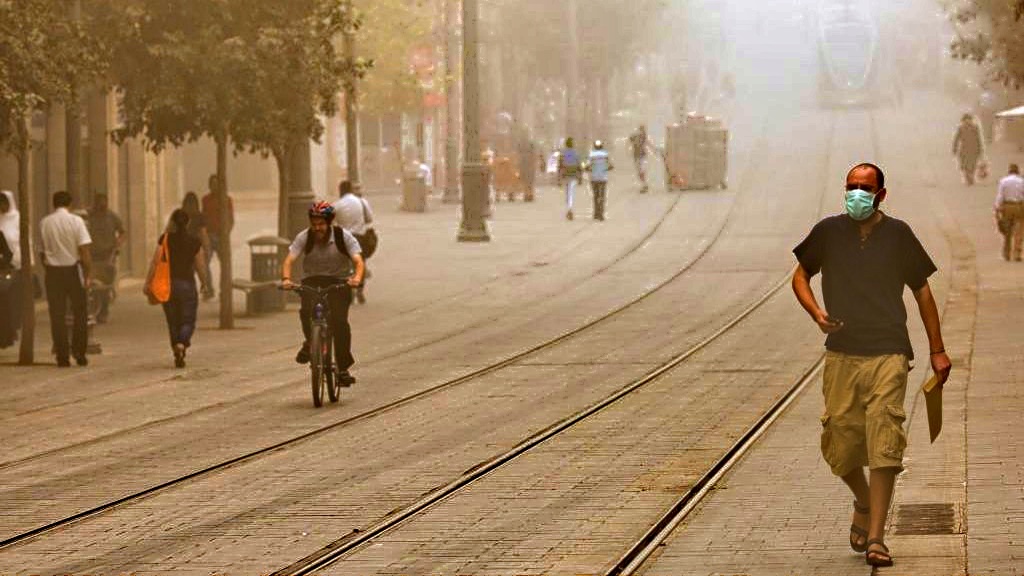}
\hspace*{\shrinkSpaceBetweenImages} &
\includegraphics[width=\sFigHaze\linewidth]{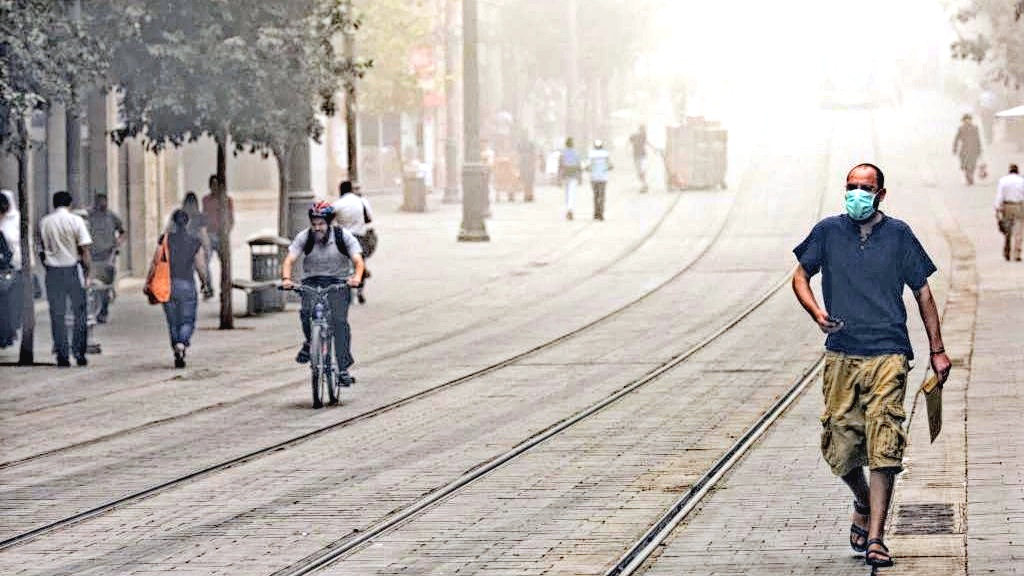}
\hspace*{\shrinkSpaceBetweenImages} &
\includegraphics[width=\sFigHaze\linewidth]{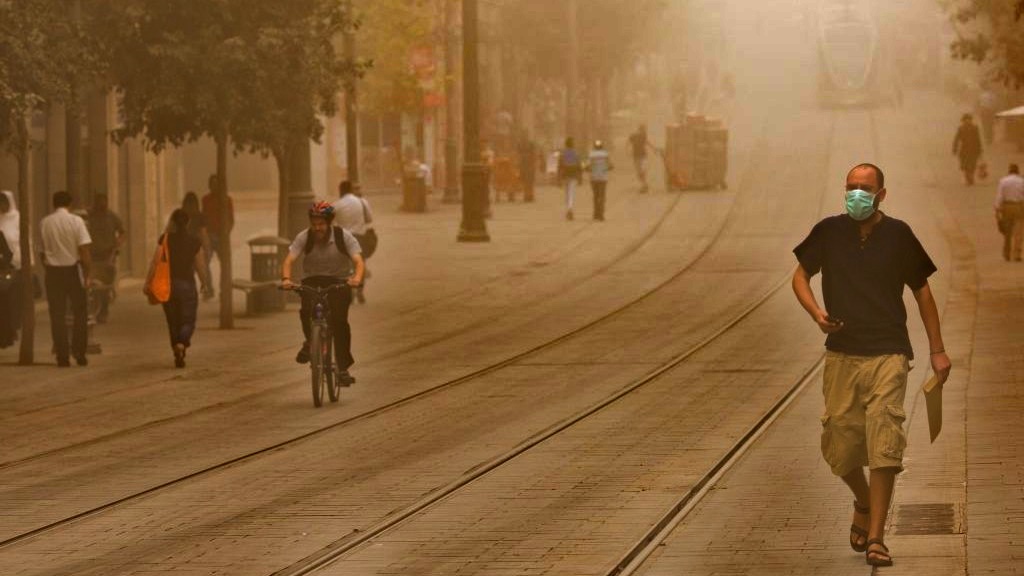}
\hspace*{\shrinkSpaceBetweenImages} &
\includegraphics[width=\sFigHaze\linewidth]{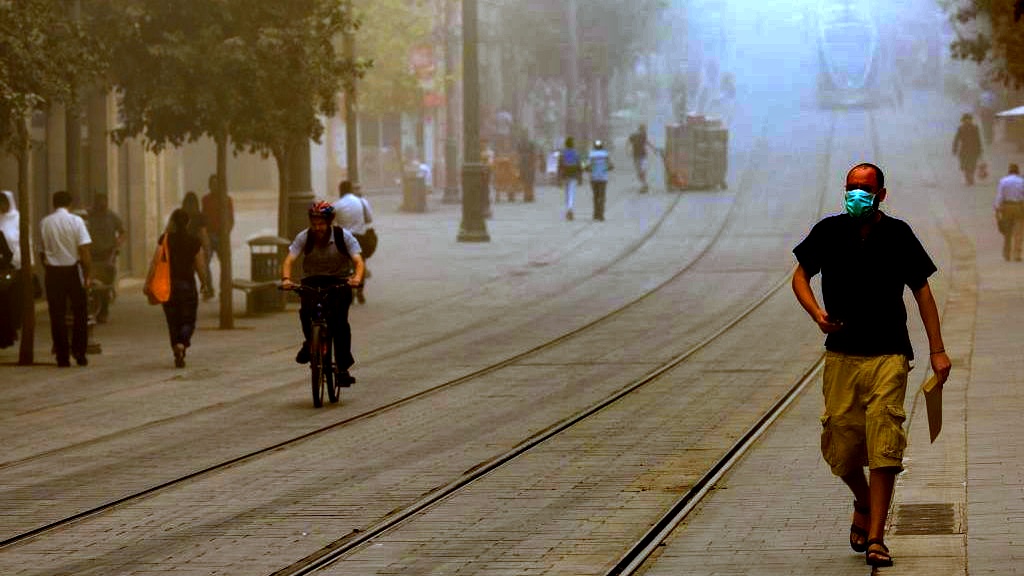}
\hspace*{\shrinkSpaceBetweenImages} &
\includegraphics[width=\sFigHaze\linewidth]{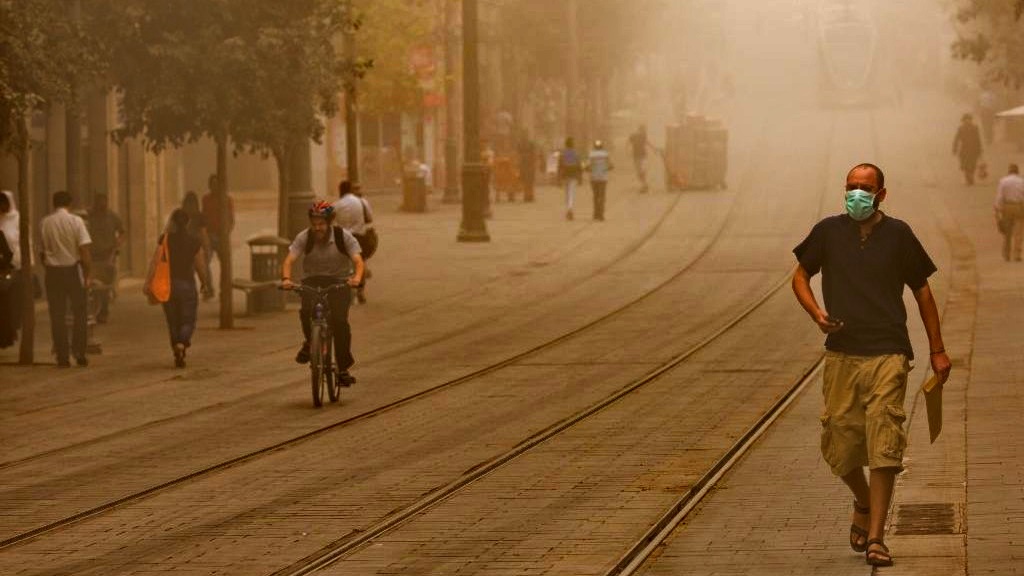}
\\
\includegraphics[width=\sFigHaze\linewidth]{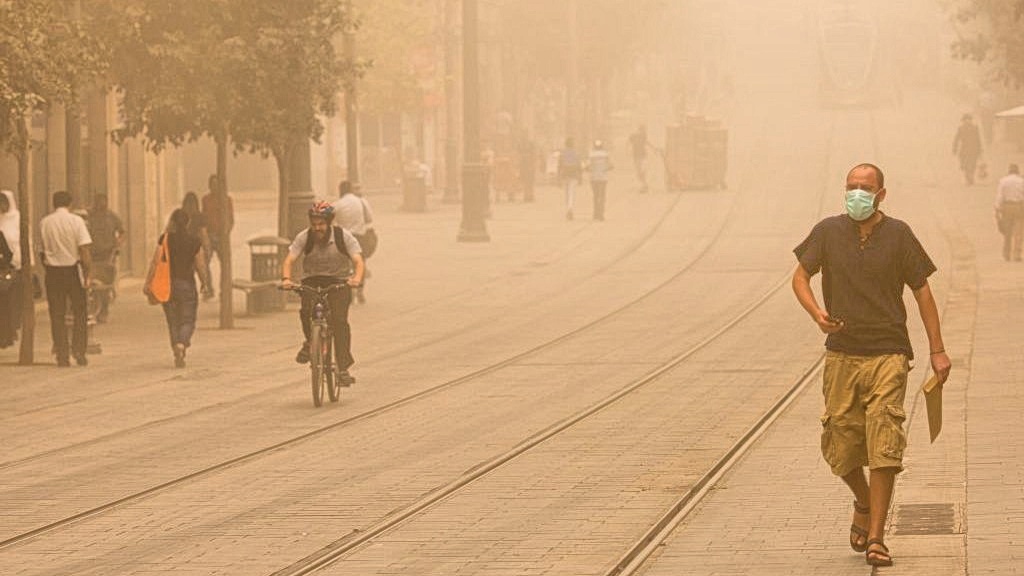}
\hspace*{\shrinkSpaceBetweenImages} &
\includegraphics[width=\sFigHaze\linewidth]{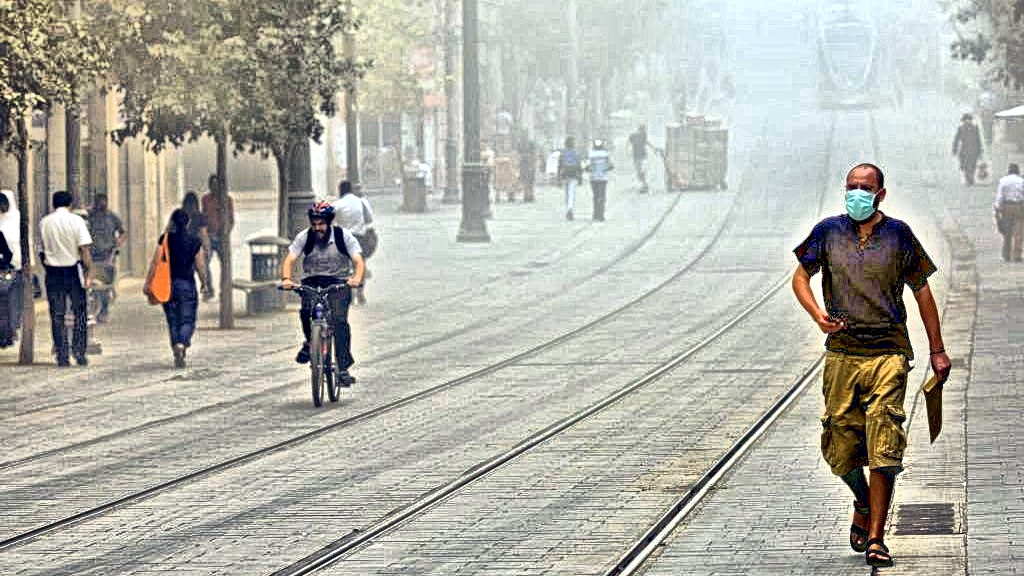}
\hspace*{\shrinkSpaceBetweenImages} &
\includegraphics[width=\sFigHaze\linewidth]{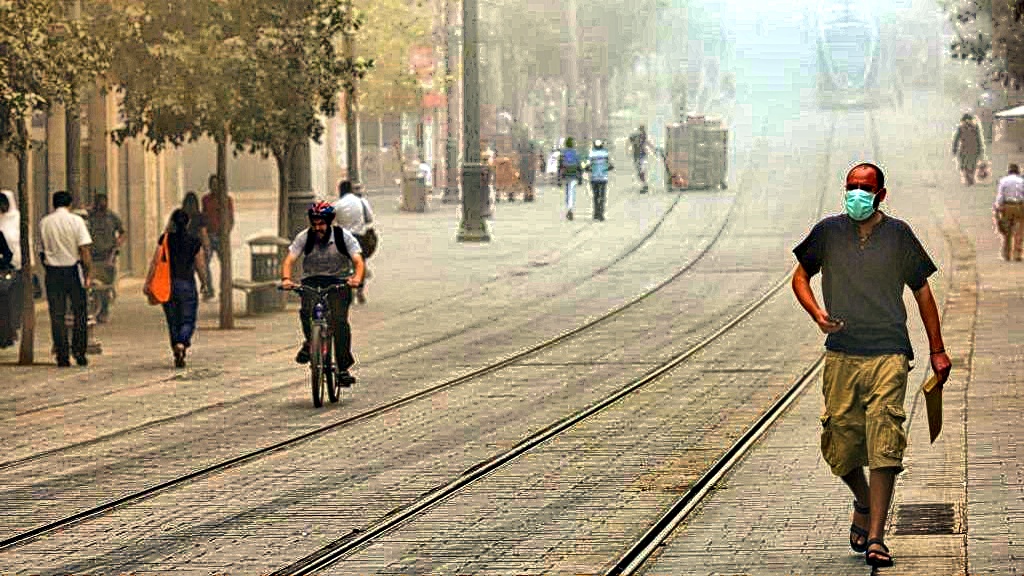}
\hspace*{\shrinkSpaceBetweenImages} &
\includegraphics[width=\sFigHaze\linewidth]{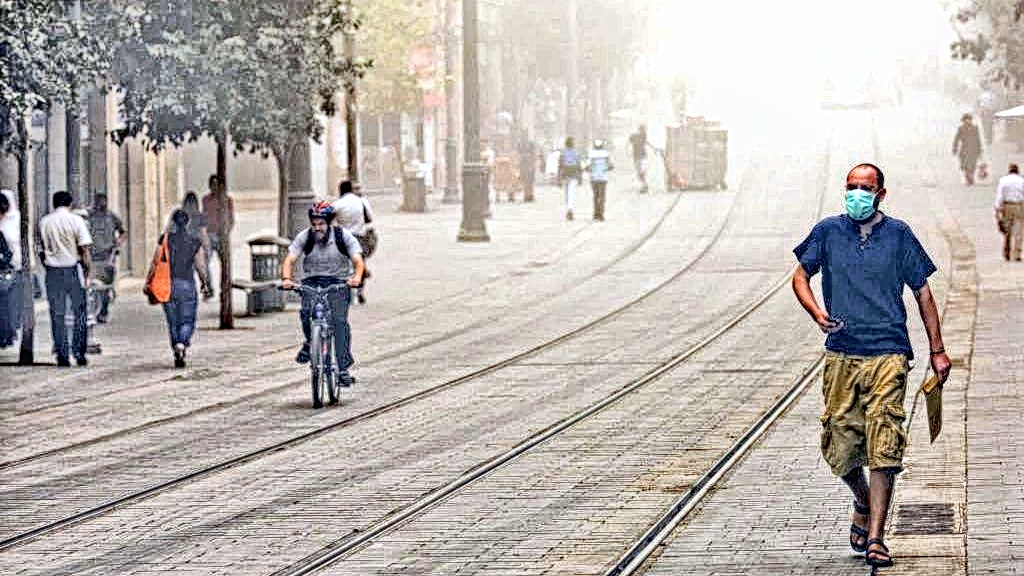}
\hspace*{\shrinkSpaceBetweenImages} &
\includegraphics[width=\sFigHaze\linewidth]{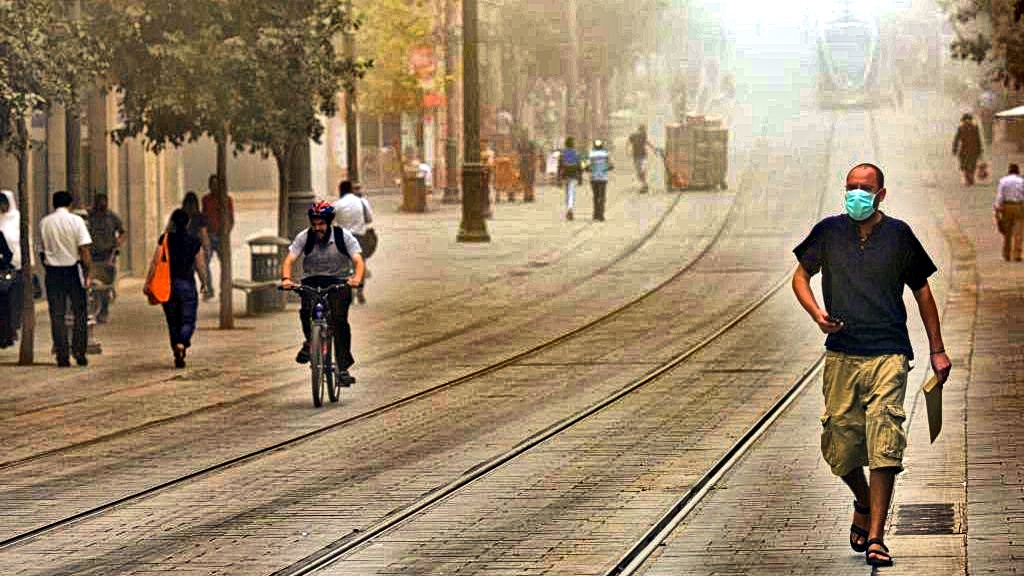}
\hspace*{\shrinkSpaceBetweenImages} &
\includegraphics[width=\sFigHaze\linewidth]{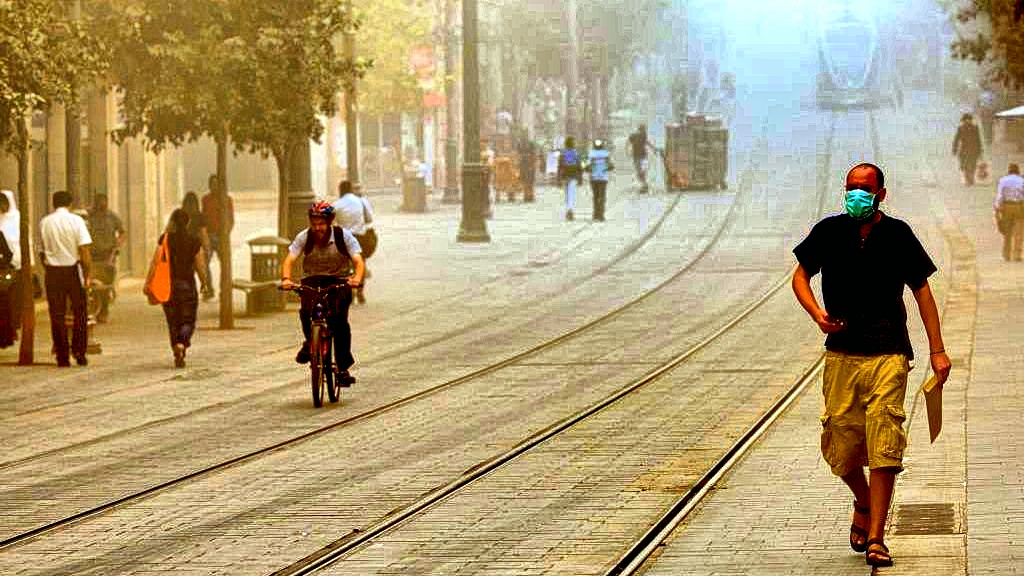}
\hspace*{\shrinkSpaceBetweenImages} &
\includegraphics[width=\sFigHaze\linewidth]{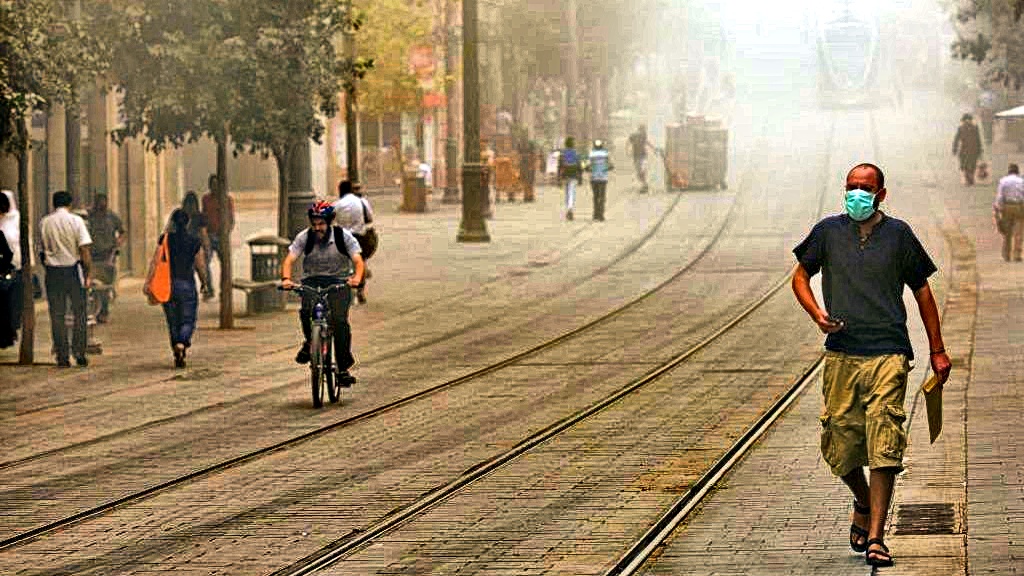}
\\
\footnotesize{sandstorm \& +\imBeamGF} & \footnotesize{\imBeamAMF \& +\imBeamGF} & \footnotesize{RGCP \& +\imBeamGF} & \footnotesize{$\mathrm{ROP}^+$ \& +\imBeamGF} & \footnotesize{SLP \& +\imBeamGF} & \footnotesize{RLP \& +\imBeamGF} & \footnotesize{RRO \& +\imBeamGF}
\end{tabular}\vspace*{-0.3cm}
\caption{Visual comparison of SOTA prior-based image clarification schemes and the proposed improved variants enhancing an image with sand-dust. RGCP combined with \imBeamGF does the best job at recovering the tram in the background.} \label{fig:Sandstorm058}
\vspace*{-0.2cm}
\end{figure*}

\begin{figure*}[h!tbp]
\hspace*{-0.3cm}\centering
\begin{tabular}{ccccccc}
\includegraphics[width=0.137\linewidth]{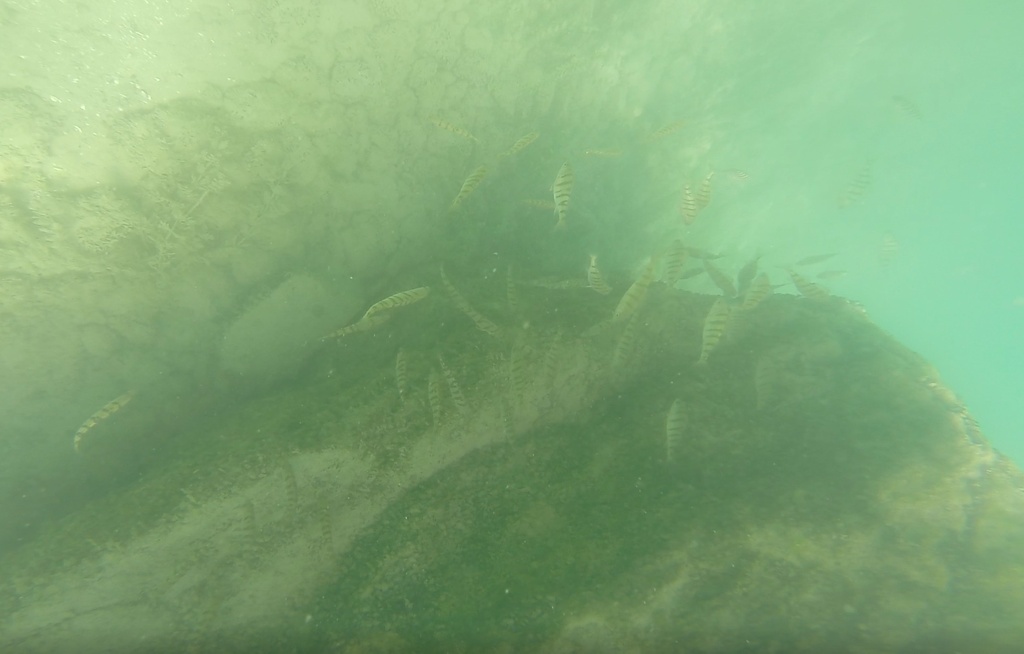}
\hspace*{\shrinkSpaceBetweenImages} &
\includegraphics[width=0.137\linewidth]{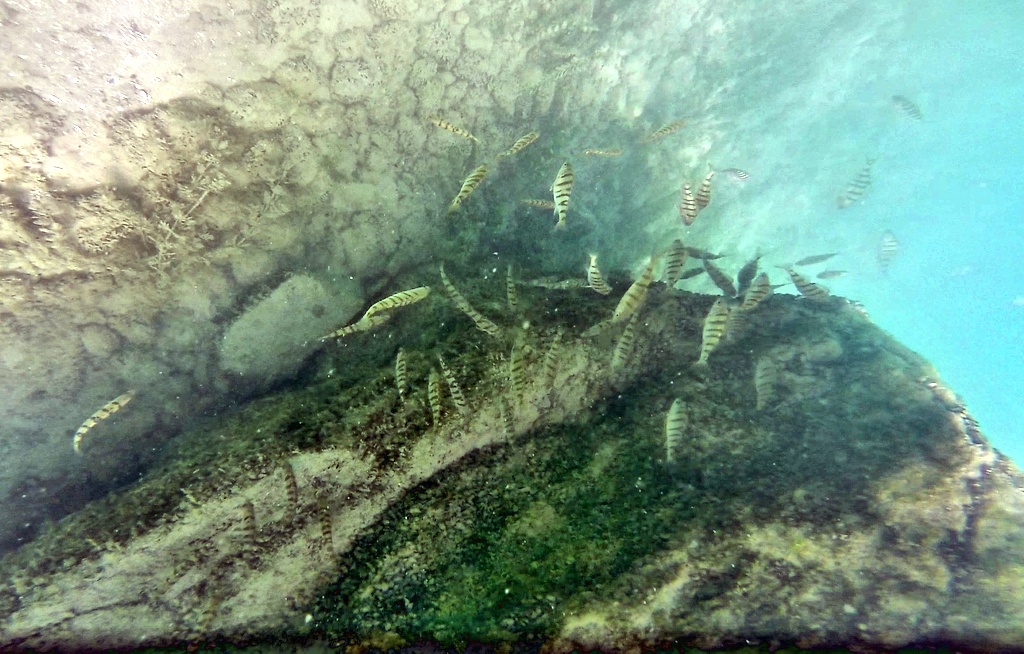}
\hspace*{\shrinkSpaceBetweenImages} &
\includegraphics[width=0.137\linewidth]{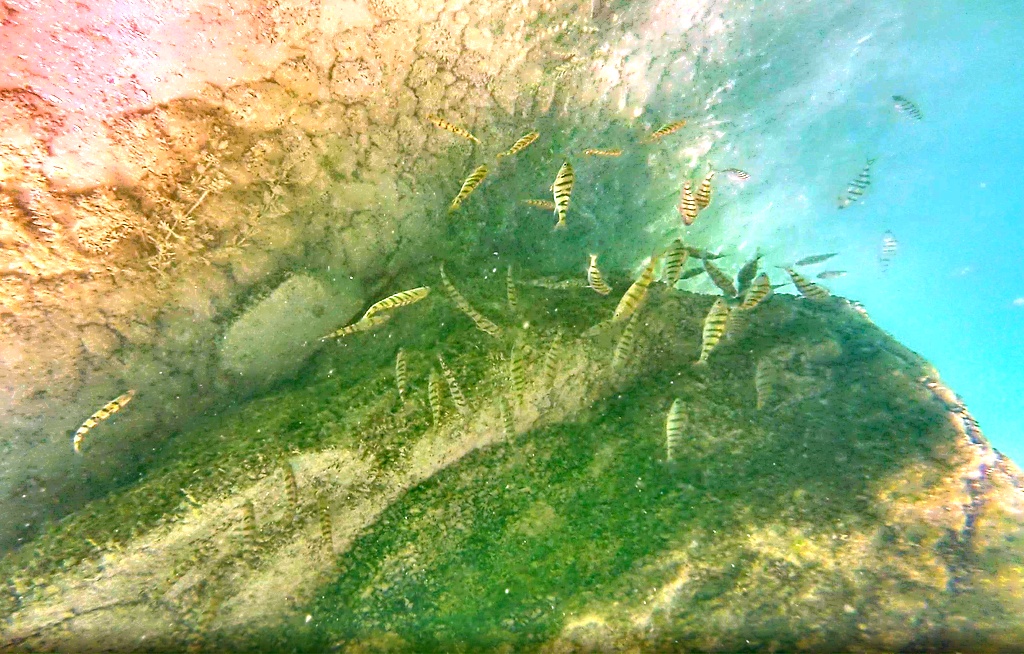}
\hspace*{\shrinkSpaceBetweenImages} &
\includegraphics[width=0.137\linewidth]{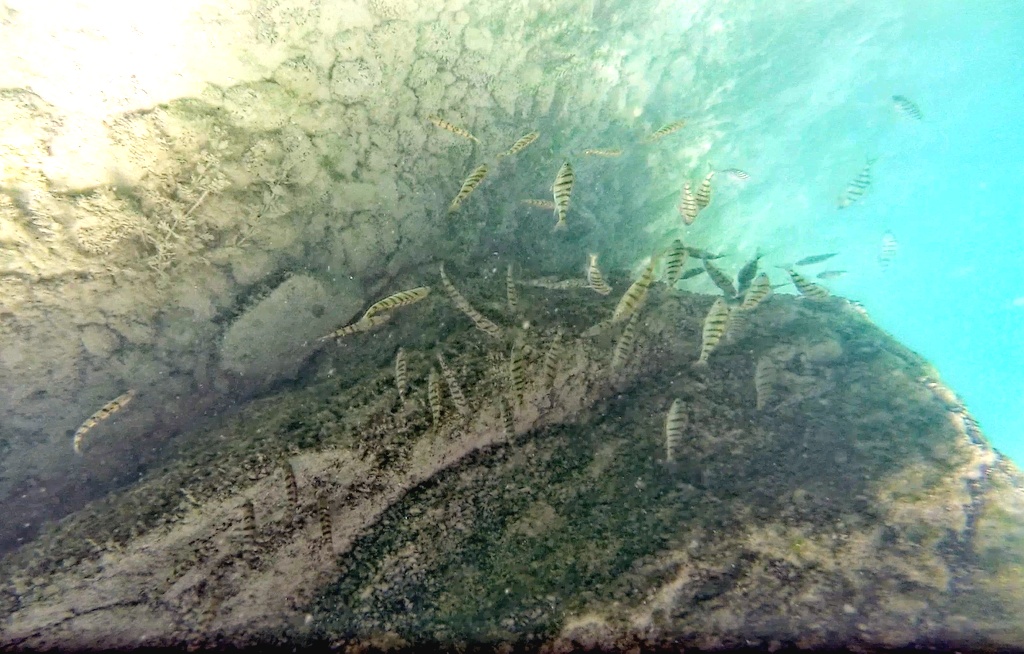}
\hspace*{\shrinkSpaceBetweenImages} &
\includegraphics[width=0.137\linewidth]{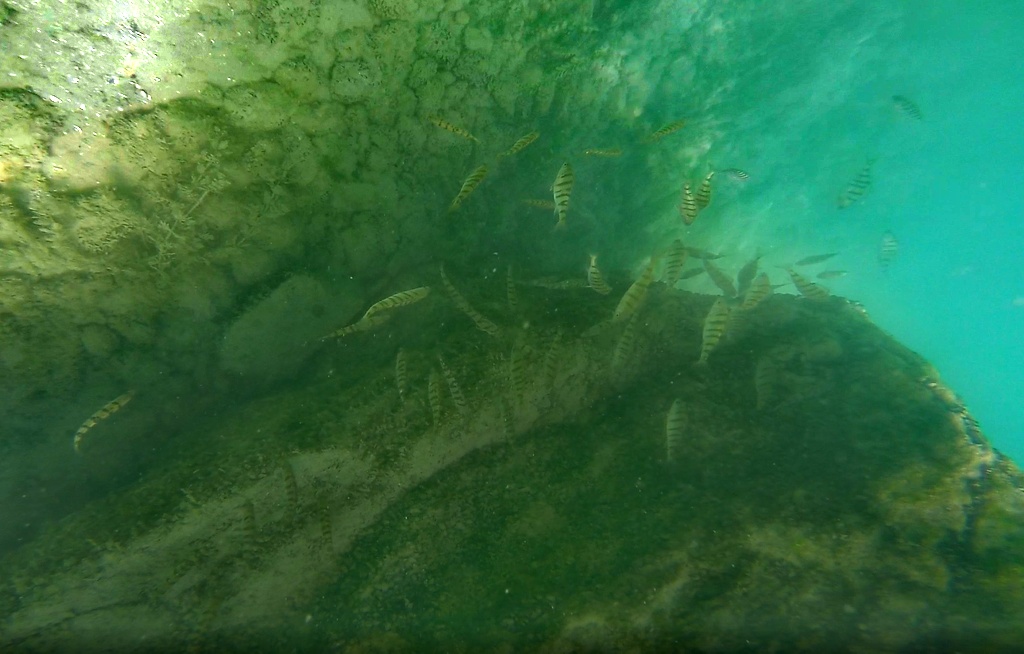}
\hspace*{\shrinkSpaceBetweenImages} &
\includegraphics[width=0.137\linewidth]{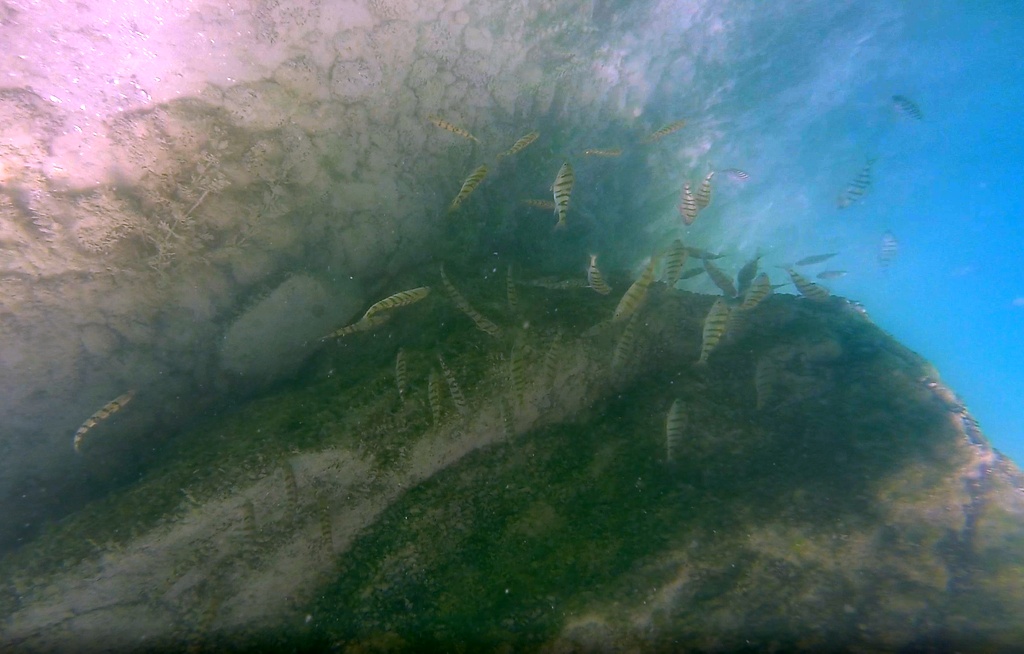}
\hspace*{\shrinkSpaceBetweenImages} &
\includegraphics[width=0.137\linewidth]{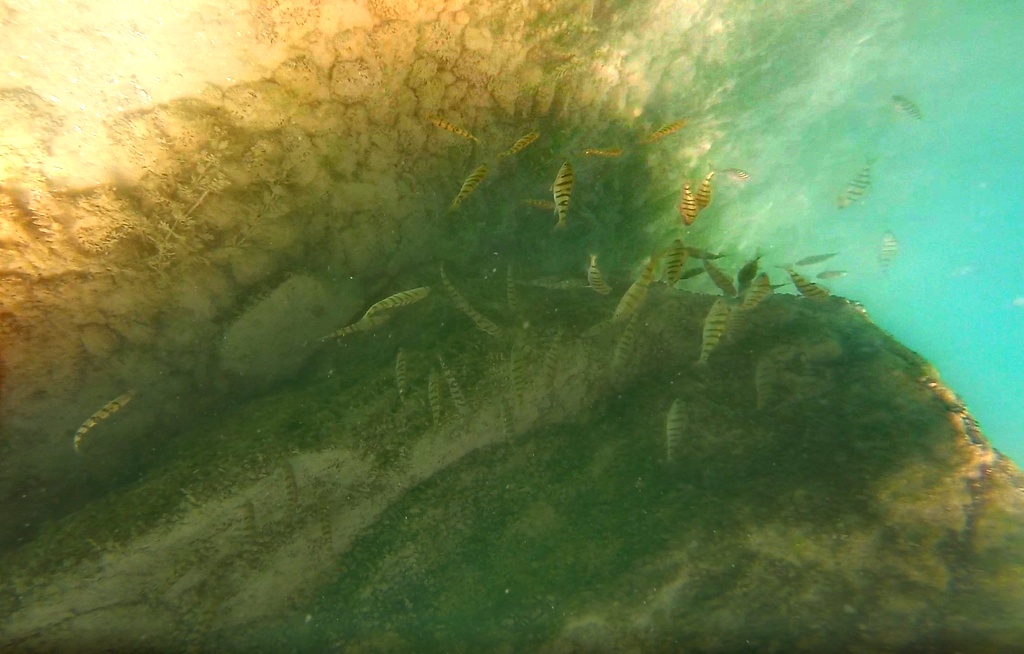}
\\
\includegraphics[width=0.137\linewidth]{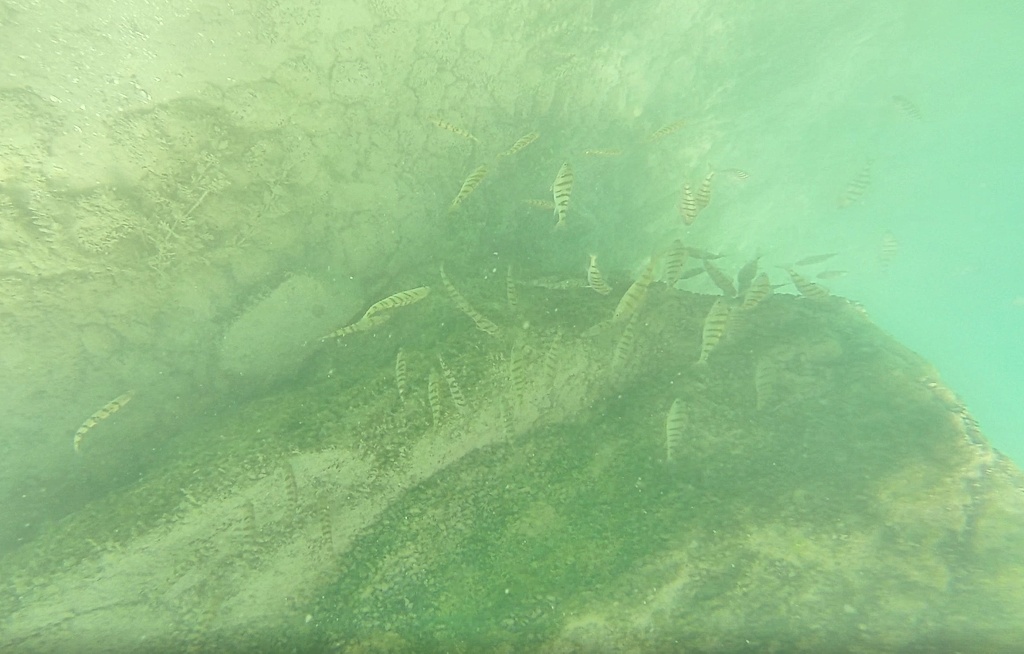}
\hspace*{\shrinkSpaceBetweenImages} &
\includegraphics[width=0.137\linewidth]{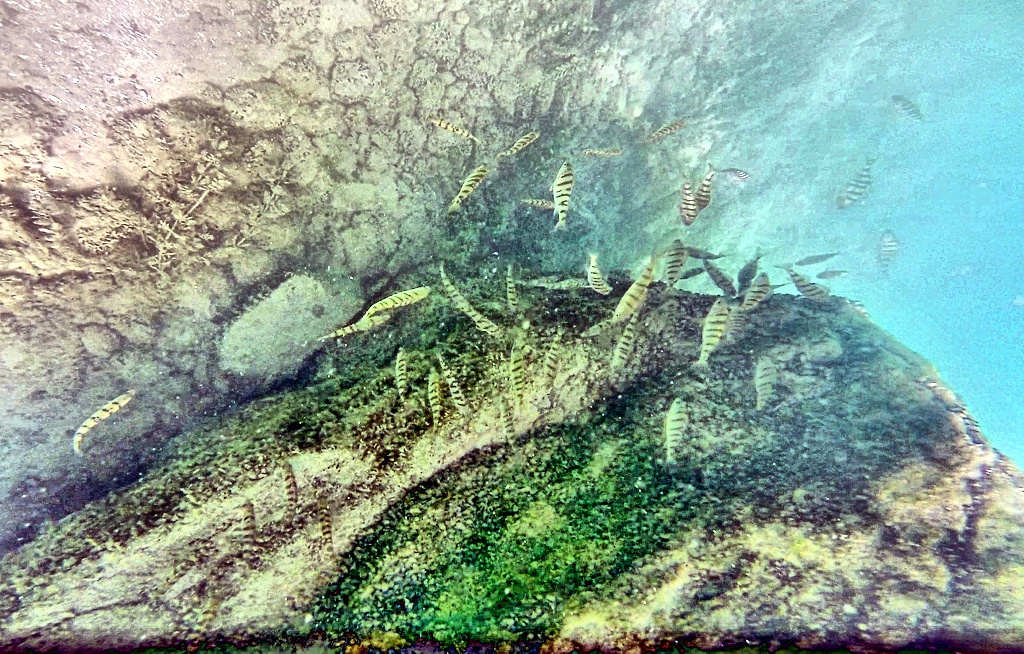}
\hspace*{\shrinkSpaceBetweenImages} &
\includegraphics[width=0.137\linewidth]{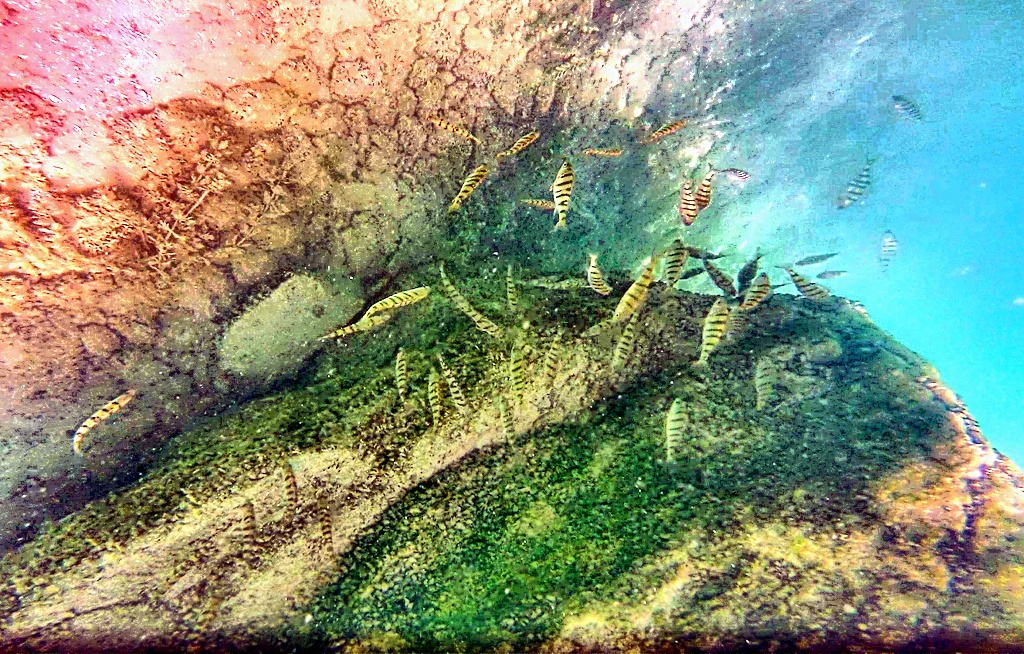}
\hspace*{\shrinkSpaceBetweenImages} &
\includegraphics[width=0.137\linewidth]{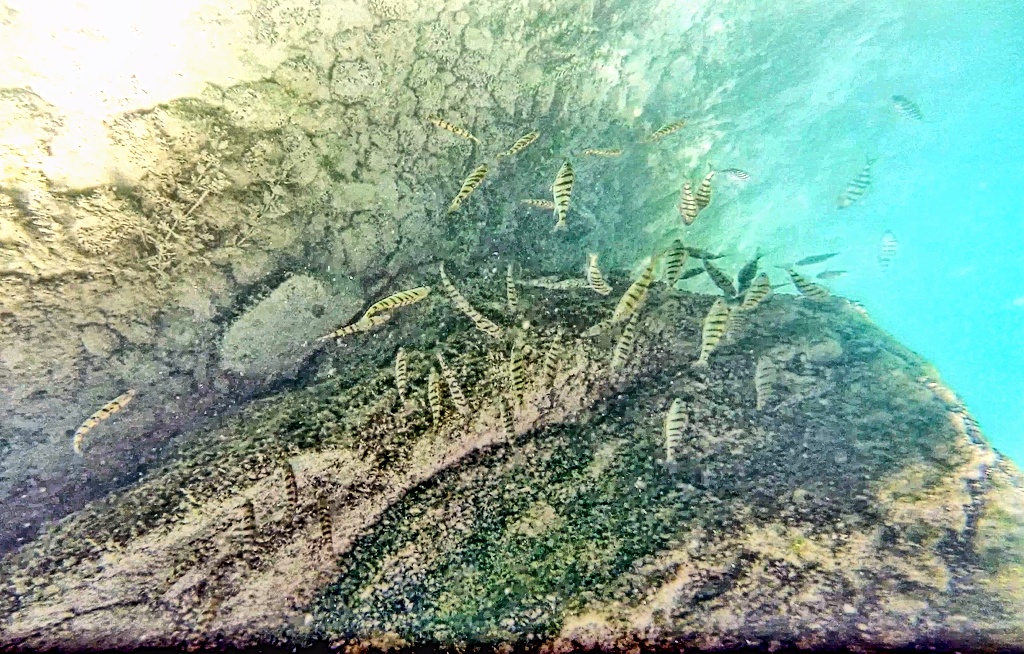}
\hspace*{\shrinkSpaceBetweenImages} &
\includegraphics[width=0.137\linewidth]{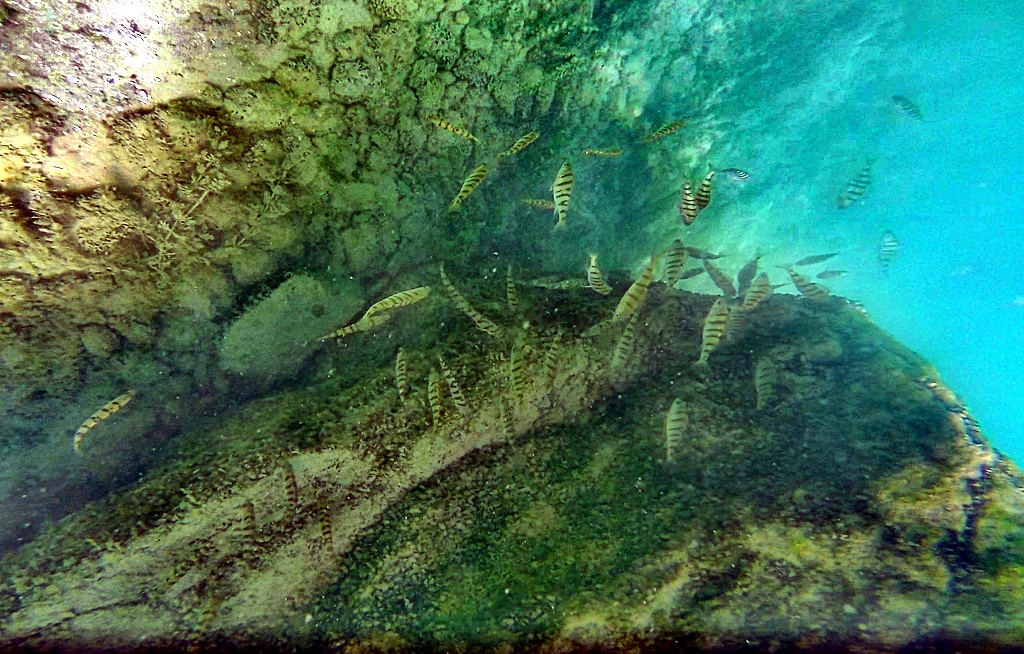}
\hspace*{\shrinkSpaceBetweenImages} &
\includegraphics[width=0.137\linewidth]{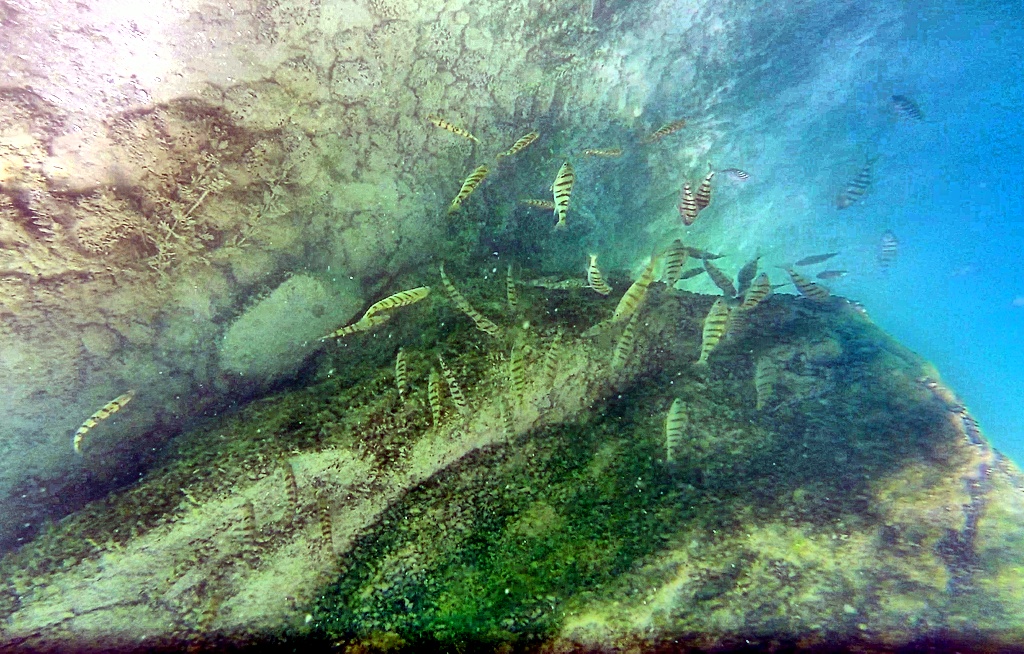}
\hspace*{\shrinkSpaceBetweenImages} &
\includegraphics[width=0.137\linewidth]{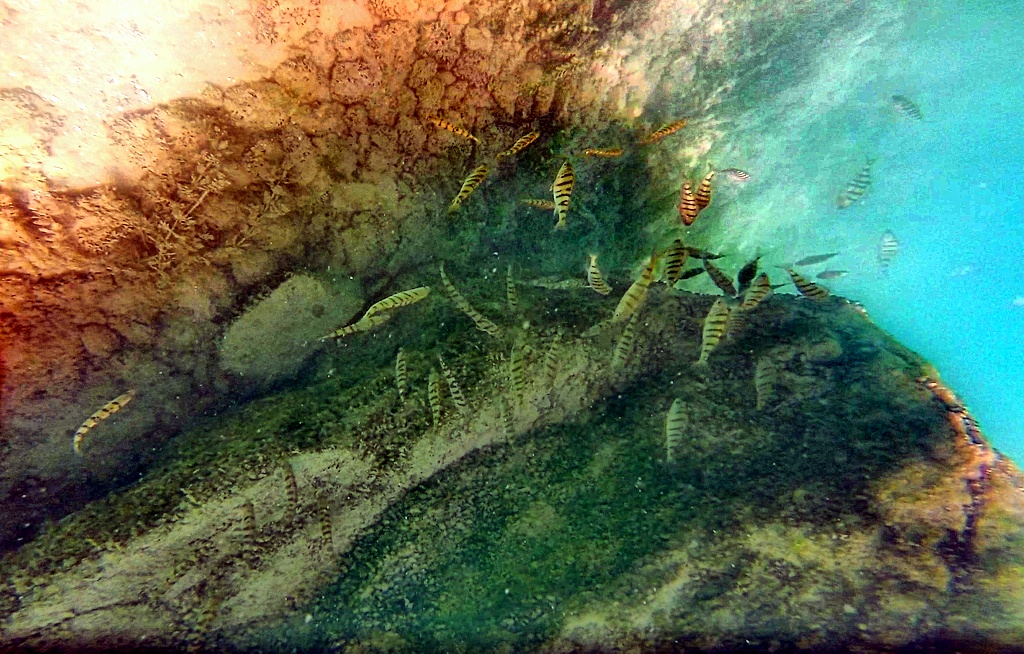}
\\
\footnotesize{underwater \& +\imBeamGF} & \footnotesize{\imBeamAMF \& +\imBeamGF} & \footnotesize{RGCP \& +\imBeamGF} & \footnotesize{$\mathrm{ROP}^+$ \& +\imBeamGF} & \footnotesize{SLP \& +\imBeamGF} & \footnotesize{RLP \& +\imBeamGF} & \footnotesize{RRO \& +\imBeamGF}
\end{tabular}\vspace*{-0.3cm}
\caption{The same methods as in Fig\,\ref{fig:yoloxHazy076}, Fig.\,\ref{fig:Hazy079} and Fig.\,\ref{fig:Sandstorm058} are compared for enhancing underwater images.} \label{fig:Underwater043}
\end{figure*}

Using \imBeamGF as an enhancing filter is not limited to its combination with \imBeamAMF as a dehazing method, but can also be combined with state-of-the-art (SOTA) dehazing methods, such as \cite{Guo-etal_tmcca23,Liu-etal_pami23,Ling-etal_tip23,Ju-etal_tip21,shin2019radiance}. The lower rows of Fig.\,\ref{fig:yoloxHazy076}, \ref{fig:Hazy079}, \ref{fig:Sandstorm058}, and \ref{fig:Underwater043} show the improved results obtained when \texttt{imGleamGF} is applied to the results obtained with RGCP \cite{Guo-etal_tmcca23}, $\mathrm{ROP}^+$ \cite{Liu-etal_pami23}, SLP \cite{Ling-etal_tip23}, RLP \cite{Ju-etal_tip21} and RRO \cite{shin2019radiance}. A comparison with the upper row shows the benefit of using \imBeamGF as an enhancement filter.

In Fig.\,\ref{fig:yoloxHazy076}, we apply an object detection module to the hazy images enhanced by these different methods. We used YOLOX \cite{Ge2021yolox} for object detection. Enhancing the images obtained by the different methods with \imBeamGF improves detection results in each case by correctly detecting the truck driver (compare upper and lower rows).

\section{Discussion and Conclusion}
We have introduced a simple and efficient method for enhancement and clarification of various image distortions (low light images, dehazing/defogging). The method is based on reverse filtering and consists of approximately inverting a filter simulating a given image distortion. Despite its simplicity, the approach shows a competitive performance to state-of-the-art LLIE, dehazing, and image clarification methods. Future research directions include applications of our approach to enhancing endoscopy images that often suffer from low light conditions \cite{Che_LighTDiff_MICCAI2024} and surgical smoke \cite{Xia_ANew_MICCAI2024}.



\bibliographystyle{ACM-Reference-Format}
\bibliography{ie}
\end{document}